\documentclass[11pt,letterpaper]{article}
\usepackage[margin=1in]{geometry}
\usepackage{times}
\usepackage{natbib}
\setcitestyle{authoryear,round,citesep={;},aysep={,},yysep={;}}
\usepackage[T1]{fontenc}
\usepackage{amsmath,amssymb,amsthm,mathtools,bm}
\usepackage{booktabs,array}
\usepackage{graphicx}
\usepackage{enumitem}
\usepackage{needspace}
\usepackage{algorithm,algpseudocode}
\usepackage{microtype}
\usepackage{aliascnt}
\usepackage[hidelinks]{hyperref}
\hypersetup{
  pdftitle={Riemannian Newton Methods with Statistical Inference},
  pdfauthor={Xiaotian Chang, Qirui Hu, Yangdi Jiang, Nan Qiao}
}
\usepackage[nameinlink,capitalise,noabbrev]{cleveref}
\makeatletter
\providecommand{\theHalgorithm}{\arabic{algorithm}}
\providecommand{\theHALG@line}{}
\renewcommand{\theHALG@line}{\theHalgorithm.\arabic{ALG@line}}
\makeatother
\allowdisplaybreaks

\newtheorem{theorem}{Theorem}[section]

\newaliascnt{proposition}{theorem}
\newtheorem{proposition}[proposition]{Proposition}
\aliascntresetthe{proposition}
\newaliascnt{lemma}{theorem}
\newtheorem{lemma}[lemma]{Lemma}
\aliascntresetthe{lemma}
\newaliascnt{corollary}{theorem}
\newtheorem{corollary}[corollary]{Corollary}
\aliascntresetthe{corollary}
\newaliascnt{assumption}{theorem}
\newtheorem{assumption}[assumption]{Assumption}
\aliascntresetthe{assumption}
\newaliascnt{remark}{theorem}

\aliascntresetthe{remark}
\newaliascnt{example}{theorem}
\newtheorem{example}[example]{Example}
\aliascntresetthe{example}
\theoremstyle{definition}
\newaliascnt{definition}{theorem}

\aliascntresetthe{definition}

\crefname{theorem}{theorem}{theorems}
\Crefname{theorem}{Theorem}{Theorems}
\crefname{proposition}{proposition}{propositions}
\Crefname{proposition}{Proposition}{Propositions}
\crefname{lemma}{lemma}{lemmas}
\Crefname{lemma}{Lemma}{Lemmas}
\crefname{corollary}{corollary}{corollaries}
\Crefname{corollary}{Corollary}{Corollaries}
\crefname{assumption}{assumption}{assumptions}
\Crefname{assumption}{Assumption}{Assumptions}
\crefname{remark}{remark}{remarks}
\Crefname{remark}{Remark}{Remarks}
\crefname{example}{example}{examples}
\Crefname{example}{Example}{Examples}
\crefname{definition}{definition}{definitions}
\Crefname{definition}{Definition}{Definitions}

\crefname{algorithm}{algorithm}{algorithms}
\Crefname{algorithm}{Algorithm}{Algorithms}

\newcommand{\M}{\mathcal M}
\newcommand{\R}{\mathbb R}
\newcommand{\E}{\mathbb E}
\newcommand{\F}{\mathcal F}
\newcommand{\T}{\mathcal T}
\newcommand{\op}{\mathrm{op}}
\newcommand{\ot}{\otimes}
\newcommand{\tr}{\operatorname{tr}}
\newcommand{\cov}{\operatorname{Cov}}
\newcommand{\diag}{\operatorname{diag}}
\newcommand{\Exp}{\operatorname{Exp}}
\newcommand{\Log}{\operatorname{Log}}
\newcommand{\norm}[1]{\left\lVert #1\right\rVert}
\newcommand{\ip}[2]{\left\langle #1,#2\right\rangle}
\newcommand{\pto}{\xrightarrow{p}}
\newcommand{\dto}{\xrightarrow{d}}
\newcommand{\ind}{\mathbf 1}
\newcommand{\Vsand}{{\mathrm{V}}_{\mathrm{sand}}}

\title{Riemannian Newton Methods with Statistical Inference}

\author{%
Xiaotian Chang\thanks{Division of Mathematical Sciences, Nanyang Technological University.}\quad
Qirui Hu\thanks{School of Statistics and Data Science, Shanghai University of Finance and Economics. Corresponding author. Email: \texttt{huqirui@mail.shufe.edu.cn}.}\quad
Yangdi Jiang\footnotemark[1]\quad
Nan Qiao\thanks{School of Mathematics, Statistics and Mechanics, Beijing University of Technology, China.}%
}
\date{August 2026}

\begin{document}
\maketitle

\begin{abstract}
We develop online statistical inference for Polyak-Ruppert averages of
inexact Newton iterates on Riemannian manifolds.  A local asymptotic
representation separates observation noise, solver randomization, and
accumulated implementation bias, and yields one-pass covariance estimators
and asymptotically valid Wald regions under local regularity and centering
conditions.  Independent fresh solves add a positive semidefinite term to
the score-sandwich covariance; averaging solver replicas reduces this term
in inverse proportion to their number.  Slowly varying predictable gains
instead cancel from the first order law.  Product torus simulations show how
Newton scaling shortens an ill-conditioned transient and how retaining
solver variation restores coverage.  Intrinsic means and principal
subspaces provide geometric applications, with a streaming data illustration.

\end{abstract}

\section{Introduction}\label{sec:introduction}
Means of directional observations, covariance matrices, and principal
subspaces are parameters with intrinsic geometry.  Riemannian
stochastic gradient methods respect that geometry, and Polyak-Ruppert
averaging converts their local iterates into regular root-$n$ estimators
\citep{Ruppert1988,PolyakJuditsky1992,KushnerYang1993,Pelletier2000,TripuraneniEtAl2018}.
In an ill-conditioned problem, however, a Newton or scaled gradient direction
can be preferable to a raw gradient.  A stochastic Newton direction solves
a linear system with an estimated Hessian and a stochastic gradient; a
randomized sketch or an iterative solver stopped after a fixed number of
steps reduces the cost of this solve
\citep{DemboEtAl1982,PilanciWainwright2017,NaDerezinskiMahoney2023}.
Randomized linear algebra supplies coordinate-free approximations
for optimization \citep{NieEtAl2026}, but its inferential effect on an
averaged estimator is not determined by its optimization accuracy alone.

The obstacle is that computational randomness can enter at two different
times.  An operator built from past observations is predictable when the
current score arrives; if its one step variation is sufficiently small,
summation by parts removes it from the first order law even when the operator
has no fixed limit \citep{AnHuoDynamic2026}.  A solver drawn with a fresh seed
at the current iteration is different: the centered product with the current
score is the martingale innovation, and its conditional mean contributes to
the root-$n$ bias, so replacing the realized product by the mean solver times
the score discards genuine sampling variation.  This temporal distinction also
separates the present argument from the augmented state analyses required for
momentum methods \citep{AnHuoAdam2026}.

Polyak-Ruppert averaging makes the distinction transparent.  On the
target tangent space, let ${{\mathrm{H}}}$ be the invertible derivative of the population
field, ${{\mathrm{S}}}$ the covariance of its centered target score $\xi$, and
${{\widetilde{\mathrm{P}}}}$ a current solver independent of $\xi$ with
invertible mean $\bar {{\mathrm{P}}}$; a star $*$ denotes the adjoint.  Under the local central limit theorem below, the
averaged estimator has covariance 
\begin{equation}\label{eq:intro-inflation}
 {{\mathrm{V}}}
 ={{\mathrm{H}}}^{-1}{{\mathrm{S}}}{({\mathrm{H}}^{*-1})}
 +{{\mathrm{H}}}^{-1}\bar {{\mathrm{P}}}^{-1}
  \E\{({{\widetilde{\mathrm{P}}}}-\bar {{\mathrm{P}}}){{\mathrm{S}}}({{\widetilde{\mathrm{P}}}}-\bar {{\mathrm{P}}})^*\}
  {(\bar {\mathrm{P}}^{*-1})}{({\mathrm{H}}^{*-1})}.
\end{equation}
The additional term is positive semidefinite, records sampling variation, and is divided by $m$ when $m$ independent solver operators are averaged.

We develop this mechanism for estimating equations on finite-dimensional
Riemannian manifolds.  A common tangent space representation, with direct
control of accumulated remainders, isolates the centered solver score product
and an optional root-$n$ local bias; it yields \eqref{eq:intro-inflation}, the exact
$1/m$ replication law, and a contrastwise variance ordering.  A local
quadratic comparison with Riemannian stochastic gradient descent (RSGD) separates this first order inferential law from
the optimization role of preconditioning: the best worst direction mean
contraction is governed by the condition number of the preconditioned drift,
which exact Newton scaling reduces to one.  The same representation identifies
the feasible estimator, implemented directions in
the fresh solver regime but raw scores after predictable cancellation.
Product centering justifies ordinary Wald regions, and a consistently estimable
local bias is removed by an explicit tangent correction.  We identify population derivatives and target scores for intrinsic
Fr\'echet means, affine-invariant covariance centers, and Grassmann
principal subspaces, and a fixed-budget rank one sketch and project solver
verifies the remaining conditions within an invariant local reporting phase.

\subsection{Related work}\label{sec:related-work}
Averaging is the classical route from a slower stochastic approximation
sequence to a root-$n$ limit
\citep{PolyakJuditsky1992,Ruppert1988,KushnerYang1993}.
For Riemannian stochastic gradients, \citet{TripuraneniEtAl2018} establish
an averaged central limit theorem under local geometric and state rate
conditions; the present analysis retains a generic $C^2$ retraction and
separates the state rate argument from the geometric and averaging
remainders.  Stochastic Newton and conditional gradient methods have
related Euclidean limits \citep{LelucPortier2023,BoyerGodichon2023}.

Randomized linear solvers introduce a second source of variation beyond
the observation stream.  The sketch and project framework of
\citet{GowerRichtarik2015} provides the rank one solver used below.
Last iterate inference and covariance estimation for sketched Newton or
sequential quadratic programming are developed by
\citet{NaMahoney2025,KuangEtAl2026,WangDuNa2026}.
For averaged iterates, \citet{DuZhuWuNa2025} prove a solver dependent CLT
and random scaling inference for sketched SQP.
Concurrent work by \citet{DuRebrovaDerezinskiNa2026} treats averaged
Newton iterations with accelerated sketch and project and identifies
covariance inflation.  Our local manifold analysis complements these
results by treating tangent space solver score products, complete
implementation error, and an implemented direction covariance estimator.

A distinct cancellation mechanism applies to sufficiently slowly varying
predictable preconditioners.  \citet{AnHuoDynamic2026} give a Euclidean
inverse gain decomposition and a sufficient stabilization threshold;
the scope of its sharpness is restricted to their stated rate class.
Our predictable branch uses the corresponding identity in a coherent
local tangent representation.  Neither predictability alone nor an
optimization convergence guarantee implies the required averaged CLT.
Model-specific online inference includes entrywise streaming PCA
\citep{KumarEtAl2025} and online multiplier bootstrap inference for
streaming $r$-PCA \citep{XuLi2026}; these concern different estimators and
studentizers from the Newton procedure studied here.

We call ${{\mathrm{H}}}^{-1}{{\mathrm{S}}}{({\mathrm{H}}^{*-1})}$ the sandwich covariance.  It is an efficiency bound
only when an information identity or a separate efficiency argument
justifies that interpretation.

The rest of the paper is organized as follows. 
{\Cref{sec:methodology} further gives the online procedure; \Cref{sec:theory} develops its local limit, covariance, and conditioning results.  \Cref{sec:simulations} examines small sample inference and solver replication.  \Cref{app:theory-details} contains proofs, sufficient conditions, and geometric applications; \Cref{app:numerical-details} gives complete numerical comparisons and the streaming principal subspace analysis.}

\section{Averaged randomized Newton inference}\label{sec:methodology}
Let $x_\star$ be a selected root of a population tangent field $G$ on a $d$-dimensional Riemannian manifold $\M$.  For optimization,
$G=\operatorname{grad}F$ for an objective $F$.  {Let }$E=T_{x_\star}\M${, let }$R${ be a }$C^2${ retraction, and write $u_k=R_{x_\star}^{-1}(x_k)$ for the iterate coordinates.}
The working neighborhood $U$ has compact closure in a larger common
retraction/inverse-retraction domain inside a normal neighborhood.
{A smooth orthonormal frame gives isometries }$\mathfrak e_x:E\to T_x\M${, with }$\mathfrak e_{x_\star}={{\mathrm{I}}}${.  Define }$\T_{x\to y}=\mathfrak e_y\mathfrak e_x^*${: it identifies tangent vectors by their frame coordinates and satisfies }$\T_{y\to z}\T_{x\to y}=\T_{x\to z}${.  It is generally different from Levi-Civita parallel transport.  A public anchor frame makes these maps computable without knowing }$x_\star${.} Stars denote adjoints, and
$(v\ot w)z=v\ip wz$.  Vector and induced operator norms are denoted by
$\norm{\cdot}$ and $\norm{\cdot}_{\op}$; $L^2$ norms take the corresponding
squared norm in expectation.

The actual update is
$x_{k+1}=R_{x_k}(-\eta_{k+1}D_{k+1})$, where
$D_{k+1}={{\mathrm{P}}}_{k+1}\widehat G_{k+1}(x_k)+e_{k+1}$, \(P_{k+1}:T_{x_k}\mathcal M\to T_{x_k}\mathcal M\), and
$\eta_k=\eta_0 k^{-\alpha}$, $\alpha\in(1/2,1)$.
The discrepancy $e_{k+1}$ collects inner solver, clipping, and
localization errors.  Current scores, operators and updated iterates
are $\F_{k+1}$-measurable, $\F_k$ contains their history, and all displayed
conditional moments exist.
The reported estimator is either the anchor average $R_a\{n^{-1}\sum_{k=1}^nR_a^{-1}(x_k)\}$ or the recursive average in \eqref{eq:riemannian-average}; both have the same first order limit under the conditions below.
\begin{algorithm}[t]
\caption{Online Newton averaging and covariance tracking}\label{alg:overview}
\begin{algorithmic}[1]
\Require $x_0${, anchor }$a${, horizon }$n${, step sizes, averaging rule, solver budgets, spectral band, and admissible safeguards.}
\State {Initialize derivative/solver means and Welford state
$(\mu,{\mathrm W})=(0,0)$.}
\For{$k=0,\ldots,n-1$}
\State {Construct }${{\mathrm{P}}}_{k+1}${ from past derivatives and fresh seeds, or use a predictable gain.}
\State {Obtain the score and derivative; apply the solver and safeguards to update }$x_{k+1}${.}
\State {Record }$D_{k+1}=-\eta_{k+1}^{-1}R_{x_k}^{-1}(x_{k+1})${; update the online average and safeguarded derivative mean.}
\If{fresh solves}
\State {Update }$\widehat {{\mathrm{P}}}^a${; set }$z=\T_{x_k\to a}D_{k+1}${.}
\Else
\State {Set }$z=\T_{x_k\to a}\widehat G_{k+1}(x_k)${.}
\EndIf
\State {Update Welford:
$\delta=z-\mu$, $\mu^+=\mu+\delta/(k+1)$,
${{\mathrm W}}^+={{\mathrm W}}+\delta\ot(z-\mu^+)$.}
\EndFor
\State {Form the fresh solver covariance in \Cref{sec:cov-estimation}, or the admissible predictable covariance in \eqref{eq:predictable-cov-estimator-algorithm}.}
\Ensure $\bar x_n$, $\T_{a\to\bar x_n}\widehat {{\mathrm{V}}}^a\T_{\bar x_n\to a}$; {report unavailable covariance if the inverse does not exist.}
\end{algorithmic}
\end{algorithm}
{\Cref{alg:main} gives the full implementation.}  Full dimensional Wald
inference also requires a positive definite covariance and a common local
chart, and a scalar interval a positive contrast variance.  Burn-in keeps the
global step index but restarts reporting accumulators with the retained count.
A step cap controls individual moves, and localization of the actual trajectory
requires a separate invariant basin or safeguard argument.  A nonsymmetric
estimating field uses a consistent Jacobian estimator and a corresponding
invertibility safeguard in place of spectral Hessian clipping.

\section{Local limit and one-pass inference}\label{sec:theory}
\subsection{Conditions and influence representation}
The expansion rests on a local linear score, control of
the implemented solver and its accumulated errors, and a martingale limit
for the resulting directions.  We summarize them here;
{their full statements are in \Cref{sec:assumptions}, primitive criteria in \Cref{app:primitive-conditions} (\Cref{prop:sample-split,prop:remainder-rates,prop:geometry-primitive}), and proofs in \Cref{app:all-proofs}.}

\paragraph{Local score and solver.}
The reported phase starts at deterministic $x_0\in U$, remains in $U$ almost surely,
and its reported average lies in the common chart with probability tending
to one.  In the reference tangent space,
$\T_{x\to x_\star}G(x)={{\mathrm{H}}}{R_{x_\star}^{-1}(x)}+r_G(x)$, where ${{\mathrm{H}}}$ is invertible,
$r_G(x)=o(\norm{{R_{x_\star}^{-1}(x)}})$, and
$\norm{{{\mathrm{H}}}{R_{x_\star}^{-1}(x)}+r_G(x)}\le C\norm{{R_{x_\star}^{-1}(x)}}$ on $U$.
The score decomposes as
$\T_{x_k\to x_\star}\widehat G_{k+1}(x_k) ={{\mathrm{H}}}u_k+r_G(x_k)+\xi_{k+1}+q_{k+1}$, where \(\xi_{k+1}\) is the target score and \(q_{k+1}\) is the centered state dependent score increment  with $\E(\xi_{k+1}\mid\F_k)=\E(q_{k+1}\mid\F_k)=0$.
Put ${{\widetilde{\mathrm{P}}}}_{k+1}=\T_{x_k\to x_\star}{{\mathrm{P}}}_{k+1}\T_{x_\star\to x_k}$
and ${{\mathrm{\Pi}}}_k=\E({{\widetilde{\mathrm{P}}}}_{k+1}\mid\F_k)$.
Define $b^\xi_{k+1}=\E({{\widetilde{\mathrm{P}}}}_{k+1}\xi_{k+1}\mid\F_k)$,
$b^q_{k+1}=\E({{\widetilde{\mathrm{P}}}}_{k+1}q_{k+1}\mid\F_k)$, and centered products
$\zeta_{k+1}={{\widetilde{\mathrm{P}}}}_{k+1}\xi_{k+1}-b^\xi_{k+1}$,
$\chi_{k+1}={{\widetilde{\mathrm{P}}}}_{k+1}q_{k+1}-b^q_{k+1}$.
Assume a deterministic invertible $\bar {{\mathrm{P}}}$ with positive stable (all eigenvalues have positive real parts)
${{\mathrm{A}}}=\bar {{\mathrm{P}}} {{\mathrm{H}}}$, a uniform conditional second moment bound for ${{\widetilde{\mathrm{P}}}}_{k+1}$,
and $\E(\norm{\chi_{k+1}}^2\mid\F_k)\le C\norm{u_k}^2$.

\paragraph{Accumulated errors.}
Decompose $\T_{x_k\to x_\star}e_{k+1}=a_{k+1}+\nu_{k+1}$ into its
conditional mean and centered part, and put
$\beta_{k+1}=b^\xi_{k+1}+b^q_{k+1}+a_{k+1}$.  The actual coordinate recursion is
$u_{k+1}=u_k-\eta_{k+1}\T_{x_k\to x_\star}D_{k+1}
+\eta_{k+1}(c_{k+1}+d_{k+1})$,
 where \(c_{k+1}\) and \(d_{k+1}\) are centered and accumulated geometric remainders, $\E(c_{k+1}\mid\F_k)=0$.
For deterministic $b\in E$, require
\begin{align*}
 n^{-1/2}\sum_{k<n}({{\mathrm{\Pi}}}_k-\bar {{\mathrm{P}}}){{\mathrm{H}}}u_k&\pto0,
 &n^{-1/2}\sum_{k<n}{{\mathrm{\Pi}}}_kr_G(x_k)&\pto0,\\
 n^{-1/2}\sum_{k<n}\beta_{k+1}&\pto b,
 &n^{-1/2}\sum_{k<n}d_{k+1}&\pto0,\\
 n^{-1}\sum_{k<n}\E(\norm{\nu_{k+1}}^2+\norm{c_{k+1}}^2\mid\F_k)&\pto0.
\end{align*}
Here and below $\sum_{k<n}$ runs from $k=0$ to $n-1$.
\paragraph{State rate and limiting noise.}
Finally, $\E\norm{u_k}^2\le C\eta_k$,
$\sqrt n\{R_{x_\star}^{-1}(\bar x_n)-n^{-1}\sum_{k=1}^nu_k\}\pto0$,
and for deterministic positive semidefinite ${{\mathrm{Q}}}$ and every $\epsilon>0$,
\[
 n^{-1}\sum_{k<n}\E(\zeta_{k+1}\ot\zeta_{k+1}\mid\F_k)\pto {{\mathrm{Q}}},\qquad
 n^{-1}\sum_{k<n}\E\{\norm{\zeta_{k+1}}^2
 \ind_{\{\norm{\zeta_{k+1}}>\epsilon\sqrt n\}}\mid\F_k\}\pto0.
\]
The appendix supplies a Lyapunov state rate criterion (\Cref{prop:state-rate}) and $C^2$ averaging
arguments that avoid a fourth moment state recursion, and shows that restarting the average after a proportional burn-in leaves the limit unchanged (\Cref{prop:tail-average}).
Conditional product centering, rather than separate
unbiasedness, determines whether $b=0$ (\Cref{cor:centered}).

\begin{theorem}\label{thm:main}
Suppose \Cref{ass:local-field,ass:score,ass:fresh-solver,ass:bias-inexact,ass:geometry,ass:averaging,ass:state-clt} hold.  Then
\begin{equation}\label{eq:main-representation}
 \sqrt n\,R_{x_\star}^{-1}(\bar x_n)
 =-{{\mathrm{A}}}^{-1}\frac1{\sqrt n}\sum_{k=1}^n\zeta_k
  -{{\mathrm{A}}}^{-1}b+o_p(1),
 \qquad {{\mathrm{A}}}=\bar {{\mathrm{P}}} {{\mathrm{H}}}.
\end{equation}
Consequently,
\begin{equation}\label{eq:main-clt-new}
 \sqrt n\,R_{x_\star}^{-1}(\bar x_n)
 \dto N_E(-{{\mathrm{A}}}^{-1}b,{{\mathrm{V}}}),
 \qquad
 {{\mathrm{V}}}={{\mathrm{A}}}^{-1}{{\mathrm{Q}}}{({\mathrm{A}}^{*-1})}.
\end{equation}
Here $N_E$ denotes a Gaussian law on $E$.  For every $C^1$ map
$\psi:\M\to\R^p$ with differential $D\psi_{x_\star}:E\to\R^p$,
\begin{equation}\label{eq:delta-method-new}
 \sqrt n\{\psi(\bar x_n)-\psi(x_\star)\}
 \dto N_p\!\left(
  -D\psi_{x_\star}{{\mathrm{A}}}^{-1}b,
  D\psi_{x_\star}{{\mathrm{V}}}D\psi_{x_\star}^*
 \right).
\end{equation}
\end{theorem}
The mean drift is ${{\mathrm{A}}}$, whereas fluctuations retain the realized solver score
product.  The shift $-{{\mathrm{A}}}^{-1}b$ persists under covariance estimation.

Dividing the coordinate recursion by its step size and summing separates
the average state from a summation by parts boundary.  The state bound and
$\alpha<1$ make that boundary negligible at the root-$n$ scale.  The
centered solver acting on the current state and the state dependent score
increment also vanish after averaging, whereas the target score product
$\zeta_k$ remains.  Its conditional variance and Lindeberg conditions give
the Gaussian limit.  Finally, the $C^2$ chart and averaging bounds transfer
the linear expansion back to the manifold.  {\Cref{app:main-proof} gives the proof.}

\subsection{Covariance estimation and Wald regions}\label{sec:cov-estimation}
In anchor coordinates, let
${{\mathrm{H}}}_a=\T_{x_\star\to a}{{\mathrm{H}}}\T_{a\to x_\star}$,
$\bar {{\mathrm{P}}}_a=\T_{x_\star\to a}\bar {{\mathrm{P}}}\T_{a\to x_\star}$,
${{\mathrm{Q}}}_a=\T_{x_\star\to a}{{\mathrm{Q}}}\T_{a\to x_\star}$ and
${{\mathrm{V}}}_a=(\bar {{\mathrm{P}}}_a{{\mathrm{H}}}_a)^{-1}{{\mathrm{Q}}}_a{((\bar {\mathrm{P}}_a{\mathrm{H}}_a)^{*-1})}$.
Define $Y_{k+1}=\T_{x_k\to a}D_{k+1}$, $\bar Y_n=n^{-1}\sum_{k<n}Y_{k+1}$,
\[
\widehat {{\mathrm{Q}}}_n^a
=n^{-1}\sum_{k<n}(Y_{k+1}-\bar Y_n)
\ot(Y_{k+1}-\bar Y_n)
={{\mathrm W}}_n/n, 
 \qquad \widehat {{\mathrm{A}}}_n^a=\widehat {{\mathrm{P}}}_n^a\widehat {{\mathrm{H}}}_n^a.
\]
Set $\widehat {{\mathrm{V}}}_n^a=(\widehat {{\mathrm{A}}}_n^a)^{-1}\widehat {{\mathrm{Q}}}_n^a{((\widehat {\mathrm{A}}_n^a)^{*-1})}$
when invertible, extend it by zero otherwise for convergence statements,
and transport it to obtain
$\widehat {{\mathrm{V}}}_n^{\bar x}=\T_{a\to\bar x_n}\widehat {{\mathrm{V}}}_n^a\T_{\bar x_n\to a}$.
\begin{assumption}
\label{ass:cov-estimation}
The online derivative and mean solver estimators $\widehat {{\mathrm{H}}}_n^a$ and $\widehat {{\mathrm{P}}}_n^a$ satisfy
\begin{equation}\label{eq:H-P-consistency}
 \widehat {{\mathrm{H}}}_n^a\pto {{\mathrm{H}}}_a,\qquad
 \widehat {{\mathrm{P}}}_n^a\pto\bar {{\mathrm{P}}}_a,
\end{equation}
and the total conditional bias $\beta_{k+1}=b_{k+1}^{\xi}+b_{k+1}^{q}+a_{k+1}$ obeys
$n^{-1}\sum_{k<n}\norm{\beta_{k+1}}^2\pto0$.
\end{assumption}
The squared bias requirement is distinct from the signed root-$n$ condition:
alternating errors can cancel in the estimator and remain in its covariance.
\begin{theorem}\label{thm:cov-estimator}
Suppose the assumptions of \Cref{thm:main} hold and
\Cref{ass:cov-estimation} is satisfied.  Then
\begin{equation}\label{eq:Qhat-consistency}
 \widehat {{\mathrm{Q}}}_n^a\pto {{\mathrm{Q}}}_a,
 \qquad
 \widehat {{\mathrm{V}}}_n^a\pto {{\mathrm{V}}}_a,
 \qquad
 \widehat {{\mathrm{V}}}_n^{\bar x}
 -\T_{x_\star\to\bar x_n}{{\mathrm{V}}}\T_{\bar x_n\to x_\star}\pto0.
\end{equation}
If ${{\mathrm{Q}}}$ is positive definite, $b=0$, and $x_\star$ and $\bar x_n$ lie in a common
inverse-retraction neighborhood with probability tending to one, fix
$\gamma\in(0,1)$ and let $\chi^2_{d,1-\gamma}$ be the $(1-\gamma)$ quantile
of the chi-squared law with $d$ degrees of freedom.  When the required
inverse exists and $\widehat {{\mathrm{V}}}_n^{\bar x}$ is positive definite,
define the uncorrected Wald region
\begin{equation*}
 \mathcal C_{n,1-\gamma}
 =\left\{x\in U: R_{\bar x_n}^{-1}(x)\text{ is defined and }
 n\,\ip{R_{\bar x_n}^{-1}(x)}
 {(\widehat {{\mathrm{V}}}_n^{\bar x})^{-1}R_{\bar x_n}^{-1}(x)}_{\bar x_n}
 \le \chi^2_{d,1-\gamma}
 \right\}
\end{equation*}
and otherwise put $\mathcal C_{n,1-\gamma}=\varnothing$.  Availability
has probability tending to one, and
\begin{equation}\label{eq:wald-coverage}
 \mathbb P\{x_\star\in\mathcal C_{n,1-\gamma}\}\longrightarrow1-\gamma.
\end{equation}
Under the same positive definiteness and neighborhood conditions, if instead
$b\ne0$, the uncorrected statistic at $x_\star$ converges to a
noncentral chi-squared law with noncentrality
\begin{equation}\label{eq:uncorrected-noncentrality}
 \lambda_b=\ip{{{\mathrm{A}}}^{-1}b}{{{\mathrm{V}}}^{-1}{{\mathrm{A}}}^{-1}b},
\end{equation}
and therefore does not in general have nominal coverage.

Bias-corrected inference is feasible when the signed accumulated bias can be
estimated.  Put $b_a=\T_{x_\star\to a}b$, and suppose an anchor space
estimator satisfies $\widehat b_n^a\pto b_a$. 
Since the limiting shift is $-{\mathrm A}^{-1}b$, define
\begin{align}
\bar x_n^{\mathrm{bc}}
&=R_{\bar x_n}\!\left[
n^{-1/2}\T_{a\to\bar x_n}
(\widehat{\mathrm A}_n^a)^{-1}\widehat b_n^a
\right], \label{eq:bias-corrected-average}\\
\widehat{\mathrm V}_n^{\mathrm{bc}}
&=\T_{\bar x_n\to\bar x_n^{\mathrm{bc}}}
\widehat{\mathrm V}_n^{\bar x}
\T_{\bar x_n^{\mathrm{bc}}\to\bar x_n}.
\notag
\end{align}
Then $\sqrt nR_{x_\star}^{-1}(\bar x_n^{\mathrm{bc}})\dto N_E(0,{{\mathrm{V}}})$,
and the Wald region obtained by replacing $\bar x_n$ and
$\widehat {{\mathrm{V}}}_n^{\bar x}$ in $\mathcal C_{n,1-\gamma}$ by
$\bar x_n^{\mathrm{bc}}$ and $\widehat {{\mathrm{V}}}_n^{\mathrm{bc}}$ has asymptotic
coverage $1-\gamma$.

For a scalar $C^1$ functional $\psi$ with
$D\psi_{x_\star}{{\mathrm{V}}}D\psi_{x_\star}^*>0$, a consistent derivative estimate
gives the usual two-sided interval when it is centered at $\bar x_n$ under
$b=0$, or at $\bar x_n^{\mathrm{bc}}$ under the displayed consistent bias
correction.
\end{theorem}
{\Cref{app:recorded-bias}} develops a computable bias estimator from recorded
implementation discrepancies, with its additional centering conditions.

\subsection{Fresh solves and covariance inflation}
For independent current solver randomization, the covariance separates into
the score sandwich $\Vsand={{\mathrm{H}}}^{-1}{{\mathrm{S}}}{({\mathrm{H}}^{*-1})}$ and a solver contribution.
\begin{theorem}\label{thm:inflation}
Suppose that, at $x_\star$, a solver operator ${{\widetilde{\mathrm{P}}}}$ is independent of a centered score $\xi$, with
\[
 \E{{\widetilde{\mathrm{P}}}}=\bar {{\mathrm{P}}},
 \qquad
 \E(\xi\ot\xi)={{\mathrm{S}}},
 \qquad
 \E\norm{{{\widetilde{\mathrm{P}}}}}_{\op}^2<\infty,
 \qquad
 {{\mathrm{H}}}\ \text{and }\bar {{\mathrm{P}}}\ \text{invertible}.
\]
Define the target product covariance and the solver variance operator by
\begin{equation}\label{eq:R-sketch}
 {{\mathrm{Q}}}=\E\{({{\widetilde{\mathrm{P}}}}\xi)\ot({{\widetilde{\mathrm{P}}}}\xi)\},
 \qquad
 {{\mathrm{R}}}
 =\E\{({{\widetilde{\mathrm{P}}}}-\bar {{\mathrm{P}}}){{\mathrm{S}}}({{\widetilde{\mathrm{P}}}}-\bar {{\mathrm{P}}})^*\}.
\end{equation}
Then
\begin{equation*}
 {{\mathrm{Q}}}=\bar {{\mathrm{P}}} {{\mathrm{S}}}\bar {{\mathrm{P}}}^*+{{\mathrm{R}}},
 \qquad {{\mathrm{R}}}\succeq0,
\end{equation*}
and
\begin{equation}\label{eq:V-inflation}
 {{\mathrm{V}}}=\Vsand+
 {{\mathrm{H}}}^{-1}\bar {{\mathrm{P}}}^{-1}{{\mathrm{R}}}{(\bar {\mathrm{P}}^{*-1})}{({\mathrm{H}}^{*-1})}.
\end{equation}
The second term is positive semidefinite and is zero if and only if
\[
 ({{\widetilde{\mathrm{P}}}}-\bar {{\mathrm{P}}}){{\mathrm{S}}}^{1/2}=0
 \quad\text{almost surely}.
\]
\end{theorem}
Averaging $m$ conditionally independent replicas with a common conditional
law and mean divides the extra term by $m$, which orders the limiting covariances in the Loewner sense and contrastwise (\Cref{cor:multiple-sketch,cor:loewner-order}).
Sandwich recovery requires the solver error applied to the target score to
vanish in conditional mean square; probability convergence of the operator
alone is insufficient (\Cref{cor:sandwich,cor:inner-accuracy}).

{For a general fixed-budget randomized sketch, \Cref{prop:fixed-sketch} states the solver conditions under which \Cref{thm:main,thm:inflation} apply.  For rank one sketch and project, \Cref{prop:rank-one-sap} verifies these conditions from a common spectral band, a spanning sketch law, and the local score and invariant basin assumptions.  Its coordinate sketch specialization underlies \Cref{sec:simulations}.}

\subsection{Conditioning and predictable gains}

Newton scaling can improve the transient even when averaging ultimately
recovers the same sandwich covariance as RSGD; a local
quadratic comparison identifies the relevant spectral ratio.
\begin{proposition}
\label{prop:rsgd-contraction}
Let ${{\mathrm{H}}}\succ0$ be self-adjoint and consider the noiseless linearized recursion on $E$ with a fixed self-adjoint preconditioner ${{\mathrm{P}}}\succ0$, $u_{k+1}=({{\mathrm{I}}}-\eta {{\mathrm{P}}}{{\mathrm{H}}})u_k$. Let $\lambda_{\min}$ and $\lambda_{\max}$ be the smallest and largest eigenvalues of ${{\mathrm{P}}}^{1/2}{{\mathrm{H}}}{{\mathrm{P}}}^{1/2}$, and define $\kappa_{{\mathrm{P}}} =\frac{\lambda_{\max}}{\lambda_{\min}}$. In the ${{\mathrm{P}}}^{-1}$ energy norm (\Cref{app:contraction-norm}),
\[
\inf_{\eta>0}
\norm{{{\mathrm{I}}}-\eta {{\mathrm{P}}}{{\mathrm{H}}}}
_{\op}
=
\frac{\kappa_{{\mathrm{P}}}-1}
{\kappa_{{\mathrm{P}}}+1},
\qquad
\eta_\star
=
\frac{2}{\lambda_{\min}+\lambda_{\max}}.
\]
For RSGD, ${{\mathrm{P}}}={{\mathrm{I}}}$ and
$\kappa_{{\mathrm{P}}}=\kappa({{\mathrm{H}}})$; for exact Newton conditioning,
${{\mathrm{P}}}={{\mathrm{H}}}^{-1}$ and
$\kappa_{{\mathrm{P}}}=1$, so the unit step removes the linearized
deterministic error.  More generally, a smaller effective condition number
yields a smaller best worst direction contraction factor in the
corresponding energy norm.
\end{proposition}
For a fresh solver with mean ${{\mathrm{P}}}$, this contraction governs the linearized
mean drift; its realized variation contributes the additional covariance
above.  A predictable gain has a different inferential effect: multiplication
by its inverse removes the gain from the leading score term, provided the
resulting summation by parts boundary is negligible.

For a predictable invertible ${{\mathrm{C}}}_k$, consider
$u_{k+1}=u_k-\eta_{k+1}\{{{\mathrm{C}}}_k({{\mathrm{H}}}u_k+\xi_{k+1})+\varepsilon_{k+1}\}$,
where $\varepsilon_{k+1}$ contains geometric and other remainders.
\begin{theorem}\label{thm:predictable}
Assume that ${{\mathrm{H}}}$ is invertible.  Let ${{\mathrm{M}}}_k={{\mathrm{C}}}_k^{-1}$ and
${{\mathrm{\Gamma}}}_k={{\mathrm{M}}}_k/\eta_{k+1}$.  Suppose
\begin{align}
 \frac1{\sqrt n}\left\{
  \norm{{{\mathrm{\Gamma}}}_0u_0}+\norm{{{\mathrm{\Gamma}}}_{n-1}u_n}
  +\sum_{k=1}^{n-1}\norm{({{\mathrm{\Gamma}}}_k-{{\mathrm{\Gamma}}}_{k-1})u_k}
 \right\}&\pto0,
 \label{eq:predictable-abel-condition}\\
 \frac1{\sqrt n}\sum_{k=0}^{n-1}{{\mathrm{M}}}_k\varepsilon_{k+1}&\pto0,
 \label{eq:predictable-rem-condition}
\end{align}
assume also that the reported online average is root-$n$ equivalent to the arithmetic average of $u_k$ and that
$n^{-1/2}(u_n-u_0)\pto0$.  If
$n^{-1/2}\sum_{k=0}^{n-1}\xi_{k+1}\dto N_E(0,{{\mathrm{S}}})$, then
\begin{equation}\label{eq:predictable-clt}
 \sqrt n\,R_{x_\star}^{-1}(\bar x_n)
 =-{{\mathrm{H}}}^{-1}\frac1{\sqrt n}\sum_{k=0}^{n-1}\xi_{k+1}+o_p(1)
 \dto N_E(0,\Vsand).
\end{equation}
No deterministic limit of ${{\mathrm{C}}}_k$ is required.
\end{theorem}
The inverse gain criterion holds, for example, when
$\sup_k\norm{{{\mathrm{C}}}_k^{-1}}_{L^2(\op)}<\infty$,
$\E\norm{u_k}^2=O(k^{-\alpha})$, and \(\|C_k^{-1}-C_{k-1}^{-1}\|_{L^2(\mathrm{op})}=O(k^{-\beta})\) with $\beta>(1+\alpha)/2$ (\Cref{cor:stabilization});
the weighted remainder condition is still required.
Under the raw score approximation and covariance conditions in
\Cref{cor:predictable-covariance}, the raw score Welford covariance
$\widehat {{\mathrm{S}}}_n^a$ yields the estimator
$(\widehat {{\mathrm{H}}}_n^a)^{-1}\widehat {{\mathrm{S}}}_n^a{((\widehat {\mathrm{H}}_n^a)^{*-1})}$ even if
${{\mathrm{C}}}_k$ has no limit.  Exact predictable solves of a spectrally clipped
streaming Hessian satisfy the increment criterion under
\Cref{cor:predictable-hessian}, with the averaging rate supplied by \Cref{prop:hessian-rate}.

When $\M=\R^d$, the retraction is addition and all geometric remainders vanish, so the representation reduces to the Euclidean averaged law: a deterministic or predictable preconditioner returns the classical sandwich covariance of \citet{PolyakJuditsky1992}, and only current solver randomization adds the replica-averaged inflation term of \eqref{eq:V-inflation} (\Cref{sec:special-cases,cor:loewner-order}).  \Cref{app:model-details} identifies the derivative and target score covariance for intrinsic Fr\'echet means, the affine-invariant covariance center, and streaming principal subspaces on the Grassmann manifold (\Cref{cor:frechet,cor:sphere-frechet,cor:spd-frechet,cor:grassmann-pca}), and \Cref{app:wdbc-details} applies the Grassmann case to a streaming principal subspace analysis of the Wisconsin Diagnostic Breast Cancer data.
\section{Numerical experiments}\label{sec:simulations}
{We compare RSGD, exact Newton, and our randomized Newton inference method (proposed), implemented by coordinate sketch and project to separate the effect of conditioning on initialization bias from the estimation and coverage effects of solver randomization.  A product torus model keeps the iterates in an invariant angular chart and gives exact linear benchmarks for both effects.}

\subsection{Model and implementation}
On $\mathbb S^1\times\mathbb S^1$ with the product metric, consider
\[
 f(\theta,Y)=\frac12\sum_{j=1}^2 h_j
       \operatorname{ang}(\theta_j-Y_j)^2,\qquad (h_1,h_2)=(1,\kappa),
\]
where $\operatorname{ang}$ is the shortest signed angular difference.
Generate $Y_j=\theta_{\star j}+\varepsilon_j$ with
$\theta_\star=(0.20,-0.15)$ and independent
$\varepsilon_j\sim\operatorname{Unif}[-\sqrt3\sigma,\sqrt3\sigma]$,
$\sigma=0.20$.  The known loss weights give ${{\mathrm{H}}}=\diag(1,\kappa)$;
{the iteration updates use the observations, not the target or noise covariance.}
The initial angles are $(-0.35,0.30)$ and the public anchor is $(0,0)$. {We cross the condition number }$\kappa\in\{1,30,300\}${ with the number of observations }$n\in\{256,1024,4096\}${, using $m\in\{1,4,16\}$ independent coordinate sketches per observation.}

RSGD uses ${{\mathrm{P}}}={{\mathrm{I}}}$ and exact Newton uses ${{\mathrm{P}}}={{\mathrm{H}}}^{-1}$.  For {proposed}, draw $m$
independent uniform coordinate sketches $s_\ell\in\{e_1,e_2\}$ per observation
and use
\[
 {{\mathrm{P}}}^{(m)}=\frac2m\sum_{\ell=1}^m
       \frac{s_\ell s_\ell^*}{s_\ell^*{{\mathrm{H}}}s_\ell},
 \qquad \E {{\mathrm{P}}}^{(m)}={{\mathrm{H}}}^{-1}.
\]
The scalar factor two matches the mean Newton drift.  {The limiting covariance is }${{\mathrm{V}}}_m=(1+1/m)\sigma^2{{\mathrm{I}}}${ for {proposed} and }${{\mathrm{V}}}=\sigma^2{{\mathrm{I}}}${ for RSGD and exact Newton.}

The schedule is $\eta_k=\min\{c(k+10)^{-0.65},\eta_{\max}\}$, with caps
$1/\kappa$, $1$, and $1/2$ for RSGD, exact Newton, and {proposed}, respectively.
{For each method and configuration, select $c\in\{0.2,0.4,0.8,1.6,3.2,6.4\}$ by minimum mean squared geodesic error over 48 pilot replications per candidate, then freeze it before independent validation and the 500-replication evaluation.  This simulation tuning uses the known target; the subsequent updates use only the observations and specified loss weights.}
Methods share observation streams; solver randomness is independent.
Discard the first $n/8$ iterates, retaining the global step index, and restart
the average and reporting accumulators.

{In replication }$r${, the reported estimate }$\widehat\theta_r${ is the fixed anchor average of the }$7n/8${ iterates retained after burn-in.}
{Proposed} uses the full observed
directions and empirical mean solver for covariance estimation; RSGD and
exact Newton use the score sandwich.  Joint $95\%$ Wald regions use the
retained count $7n/8$ and the $\chi^2_2$ quantile.  Estimation error is
$\operatorname{RMSE}_g=\{500^{-1}\sum_{r=1}^{500}
d^2(\widehat\theta_r,\theta_\star)\}^{1/2}$.
The Monte Carlo standard error of coverage near $95\%$ is about one
percentage point.  {\Cref{app:torus-design,app:timing} give the remaining design details, all selected constants, checkpoint errors, and timings.}

\subsection{Conditioning and small sample inference}

\begin{table}[!ht]
\centering
\caption{Product torus comparison across spectral ratios and sample sizes.  Coverage
is for joint $95\%$ Wald regions; geodesic RMSE is multiplied by $10^3$.
{Proposed} uses $m=4$ independent coordinate solves.  Every entry uses 500 paired
replications; solver budgets and checkpoint errors are in the appendix.}
\label{tab:torus-main}
\begin{tabular}{rrrrrrrr}
\toprule
& & \multicolumn{2}{c}{RSGD} & \multicolumn{2}{c}{{Proposed}, $m=4$} & \multicolumn{2}{c}{Exact Newton}\\
\cmidrule(lr){3-4}\cmidrule(lr){5-6}\cmidrule(lr){7-8}
$\kappa$ & $n$ & Cov. (\%) & RMSE & Cov. (\%) & RMSE & Cov. (\%) & RMSE\\
\midrule
1 & 256 & 95.8 & 18.99 & 95.2 & 21.00 & 95.8 & 18.99\\
1 & 1024 & 95.2 & 9.84 & 94.6 & 11.11 & 95.2 & 9.84\\
1 & 4096 & 96.2 & 4.63 & 94.4 & 5.38 & 96.2 & 4.63\\
\addlinespace
30 & 256 & 75.4 & 29.60 & 96.4 & 21.60 & 96.6 & 18.85\\
30 & 1024 & 96.0 & 9.56 & 95.0 & 10.77 & 94.8 & 9.56\\
30 & 4096 & 95.4 & 4.67 & 93.2 & 5.53 & 95.8 & 4.66\\
\addlinespace
300 & 256 & 0.0 & 347.57 & 95.4 & 20.68 & 95.8 & 18.88\\
300 & 1024 & 0.0 & 114.03 & 94.8 & 10.86 & 95.4 & 9.65\\
300 & 4096 & 41.2 & 9.55 & 96.2 & 5.25 & 95.4 & 4.70\\
\bottomrule\end{tabular}\end{table}

{At }$\kappa=1${, RSGD and exact Newton coincide, while {proposed} adds randomization error (\Cref{tab:torus-main}).  At }$\kappa=30${, Newton scaling improves the shortest horizon comparison, but RSGD approaches the exact Newton error as }$n${ increases.  At }$\kappa=300${, substantial differences persist: with }$n=256${, {proposed} with }$m=4${ attains }$95.4\%${ coverage, whereas none of the RSGD regions covers the target.  Exact Newton is the deterministic preconditioning benchmark and has the smallest errors; {Proposed} is its randomized counterpart.  The improvement over RSGD reflects conditioning under the stated step schedules, while randomization carries a separate statistical cost.}

{The diagonal exact solve is inexpensive in this design, and {Proposed} takes longer per path (\Cref{tab:torus-time}; \Cref{app:timing} reports the timing values and the aggregated coordinate implementation), so the gain over RSGD is in performance at a fixed observation budget.}

The linear chart makes the source of this transient explicit.  For RSGD,
the mean error in coordinate $j$ at step $k$ is
\[
 \E(\theta_{kj}-\theta_{\star j})
 =(\theta_{0j}-\theta_{\star j})
   \prod_{t=1}^k(1-\eta_t h_j).
\]
Exact Newton and mean rescaled {proposed} replace $h_j$ by one in this expression,
using their respective step schedules.  {The caps make every coordinate update a convex combination of its preceding value and the observation.  Since both lie in }$(-0.6,0.6)^2${, all paths stay in this angular chart, where the recursion and transport are exactly linear.} The RSGD cap $1/\kappa$ protects
the stiff coordinate but slows the removal of initial error in the
other coordinate, and tail averaging reduces this error only as the iterates
themselves approach the target.  Poor coverage can therefore persist
despite a decreasing RMSE: the remaining offset is still large relative to
the sampling standard error.  Newton scaling directly reduces that offset,
while the direction covariance accounts for {proposed}'s additional variation.

\paragraph{Error along the averaging window.}
To distinguish a transient from the eventual sampling error,
\Cref{tab:torus-trajectory} follows the $n=4096$ runs through three
checkpoints.  Each method keeps the step constant chosen for this horizon;
the intermediate averages are not separately tuned.  In the well-conditioned
case, error roughly halves when the retained sample count quadruples,
as the root-$n$ approximation suggests.  The Newton methods show the same
pattern at $\kappa=300$.  RSGD instead has a much larger early error and
continues to shed its initialization effect throughout the window.

{At }$\kappa=30${, the RSGD checkpoint errors are already close to the exact Newton benchmark, between the well-conditioned and persistent transient regimes.}

\begin{table}[!ht]
\centering
\caption{Geodesic RMSE along the $n=4096$ averaging window, multiplied by
$10^3$.  The checkpoints contain 896, 1792 and 3584 retained observations
after a burn-in of 512.  Each row follows the same 500 paths with a fixed
step constant; {Proposed} uses $m=4$.}
\label{tab:torus-trajectory}
\begin{tabular}{rlrrr}
\toprule
$\kappa$ & Method & 25\% & 50\% & 100\%\\
\midrule
1 & RSGD & 9.35 & 6.57 & 4.63\\
 & Exact Newton & 9.35 & 6.57 & 4.63\\
 & {Proposed}, $m=4$ & 10.86 & 7.57 & 5.38\\
\addlinespace
30 & RSGD & 9.20 & 6.67 & 4.67\\
 & Exact Newton & 8.80 & 6.51 & 4.66\\
 & {Proposed}, $m=4$ & 11.01 & 7.99 & 5.53\\
\addlinespace
300 & RSGD & 32.81 & 17.77 & 9.55\\
 & Exact Newton & 9.40 & 6.60 & 4.70\\
 & {Proposed}, $m=4$ & 9.81 & 7.25 & 5.25\\
\bottomrule
\end{tabular}
\end{table}

\subsection{Covariance correction and solver replication}
\begin{table}[!ht]
\centering
\caption{Solver covariance correction and replication.  The two coverage
columns use the same {proposed} estimator with $m=1$: direction covariance versus
the score only counterfactual that omits solver variation.  Geodesic RMSEs
for $m \in \{1,4,16\}$ use 500 replications and are multiplied by $10^3$.}
\label{tab:solver-budget}
\begin{tabular}{rrrrrrr}
\toprule
& & \multicolumn{2}{c}{Coverage (\%), $m=1$} & \multicolumn{3}{c}{RMSE $\times10^3$}\\
\cmidrule(lr){3-4}\cmidrule(lr){5-7}
$\kappa$ & $n$ & Direction & Score only & $m=1$ & $m=4$ & $m=16$\\
\midrule
1 & 256 & 95.8 & 76.2 & 27.01 & 21.00 & 19.57\\
1 & 1024 & 93.0 & 75.2 & 14.01 & 11.11 & 10.16\\
1 & 4096 & 95.6 & 80.0 & 6.61 & 5.38 & 4.85\\
\addlinespace
30 & 256 & 95.4 & 78.4 & 26.94 & 21.60 & 19.21\\
30 & 1024 & 95.4 & 78.6 & 13.40 & 10.77 & 9.86\\
30 & 4096 & 95.4 & 77.8 & 6.62 & 5.53 & 4.81\\
\addlinespace
300 & 256 & 94.4 & 79.0 & 27.19 & 20.68 & 19.29\\
300 & 1024 & 94.0 & 76.6 & 13.92 & 10.86 & 9.94\\
300 & 4096 & 93.4 & 74.0 & 6.88 & 5.25 & 4.80\\
\bottomrule
\end{tabular}
\end{table}

{For {proposed} with }$m=1${ in this product torus design, retaining solver variation substantially improves the reported coverage (\Cref{tab:solver-budget}): direction based coverage ranges from }$93.0\%${ to }$95.8\%${, against }$74.0\%$--$80.0\%${ for the same estimates using only score covariance, which isolates the inferential effect of the omitted solver term.}  Increasing $m$ reduces RMSE throughout
the grid, consistent with the $1/m$ covariance contribution.  Corrected
{coverage is generally close to nominal across the budgets but ranges from }$93.0\%${ to }$98.4\%${; the }$98.4\%${ coverage at $(\kappa,n,m)=(30,256,16)$ is a finite-sample departure rather than evidence of exact calibration (\Cref{tab:torus-all,tab:torus-all-kappa30,tab:torus-all-kappa300} give the complete comparison).}

The score only coverage loss also has an explicit benchmark: the limiting
covariance is $(1+1/m)\sigma^2{{\mathrm{I}}}$, so studentizing by the score sandwich
$\sigma^2{{\mathrm{I}}}$ gives the limiting Wald statistic $(1+1/m)\chi^2_2$ and coverage
\[
 \Pr\{(1+1/m)\chi^2_2\le\chi^2_{2,0.95}\}
 =1-0.05^{\,m/(m+1)},
\]
that is $77.6\%$, $90.9\%$, and $94.0\%$ for $m \in \{1,4,16\}$.
The observed score only results follow this progression: replication
reduces the omitted variation, but additional observations alone do not
remove it at a fixed solver budget, whereas direction based studentization
targets the covariance at that budget.

\section{Concluding remarks}\label{sec:conclusion}
Our proposed framework enables valid inference for averaged randomized Newton methods without requiring solver randomness to vanish. Implemented direction covariance captures fresh solver variation, while replication reduces this cost. In ill-conditioned settings, the method retains Newton-type conditioning benefits and improves finite-sample estimation and coverage over RSGD.

Slowly varying predictable gains instead permit score based inference. Simulations support these mechanisms; dependent streams and residual adaptive solvers remain for future work.


\clearpage

\section*{AI use statement}
Generative AI tools were used to revise wording and improve the clarity and readability of the manuscript; to identify potentially relevant literature, references, and search directions, and to assist with research execution, including drafting, debugging, and refining code. For mathematical proofs, AI was used primarily to review and check the authors’ arguments for potential errors, inconsistencies, or gaps. AI-assisted manuscript revisions, literature-search suggestions, feedback on research ideas and proofs, and generated or modified code were reviewed and verified by the authors before incorporation into the final work. The authors independently assessed the accuracy and appropriateness of these outputs and retain full responsibility for the methodology, proofs, implementation, analysis, and conclusions presented in the paper.

\section*{Reproducibility Statement}
The code and an accompanying \texttt{README} file are submitted with the
manuscript. We provide a dedicated reproduction script for each figure and
table, allowing the reported results to be reproduced individually. The
\texttt{README} provides instructions for running the code.

\clearpage
\appendix

\section{Numerical and application details}\label{app:numerical-details}

\subsection{Coordinate sketch and project implementation}\label{app:torus-design}
For the model in \Cref{sec:simulations}, let
$B\sim\operatorname{Bin}(m,1/2)$ count the first-coordinate sketches.
Then
\[
 {{\mathrm{P}}}^{(m)}{{\mathrm{H}}}=\diag\{2B/m,2(m-B)/m\}.
\]
Sampling this count directly is distributionally equivalent to summing
$m$ independent sketches.  Its first two moments give mean drift ${{\mathrm{I}}}$ and
limiting covariance $(1+1/m)\sigma^2{{\mathrm{I}}}$, whereas exact Newton and RSGD have
limiting covariance $\sigma^2{{\mathrm{I}}}$.

The step caps make each coordinate update a convex combination of its
previous value and the current observation.  The initial point and all
observations lie in $(-0.6,0.6)^2$, so this rectangle is invariant.
Angular differences never meet the cut locus, transport is the identity,
and fixed anchor averaging is arithmetic averaging in these coordinates.
Thus the experiment realizes the local linear model exactly on the torus,
separating conditioning from curvature.

After burn-in, all reporting accumulators restart while the step index is
unchanged.  {Proposed} uses $\widehat {{\mathrm{A}}}=\widehat {{\mathrm{P}}} {{\mathrm{H}}}$ and the centered empirical
covariance of implemented directions; RSGD and exact Newton use the score
sandwich.  Methods share observations within each replication, with
independent observation and solver random generators.
The following tables give every selected step constant, both covariance
calculations, and errors at three fractions of the retained window.

{After selecting step constants using 48 pilot replications per
candidate, 96 independent validation replications per configuration checked
that outputs were finite and paths stayed in the invariant rectangle.
This stage did not alter the selected constants.  The subsequent 500
evaluation replications used independent seeds.}

{For {proposed} with }$m=4${ minus RSGD, the paired mean squared-geodesic-error difference has a }$95\%${ Monte Carlo interval }$[-4.74,-3.45]\times10^{-4}${ at $(\kappa,n)=(30,256)$ and } $[-0.121,-0.120]$ 
{ at $(\kappa,n)=(300,256)$.  Both comparisons use the same 500 paired replications as the reported RMSEs.}

\begin{table}[p]\centering
\caption{Complete product torus results at $\kappa=1$.
Cov. is implemented direction coverage for {proposed} and score-sandwich coverage
for RSGD and exact Newton.  Score is the score only counterfactual for the
same estimate.  All RMSEs are multiplied by $10^3$; fractions refer to
the retained averaging window.  The constant $c$ was selected before
evaluation.  Each row has 500 replications.}
\label{tab:torus-all}
\begin{tabular}{rlrrrrrr}
\toprule
& & & \multicolumn{2}{c}{Coverage (\%)} & \multicolumn{3}{c}{RMSE $\times10^3$}\\
\cmidrule(lr){4-5}\cmidrule(lr){6-8}
$n$ & Method & $c$ & Cov. & Score & 25\% & 50\% & 100\%\\
\midrule
256 & RSGD & 3.2 & 95.8 & 95.8 & 36.35 & 26.29 & 18.99\\
256 & Exact Newton & 3.2 & 95.8 & 95.8 & 36.35 & 26.29 & 18.99\\
256 & {Proposed}, $m=1$ & 0.8 & 95.8 & 76.2 & 46.82 & 36.69 & 27.01\\
256 & {Proposed}, $m=4$ & 1.6 & 95.2 & 91.6 & 39.29 & 29.49 & 21.00\\
256 & {Proposed}, $m=16$ & 1.6 & 95.8 & 94.4 & 36.85 & 26.79 & 19.57\\
\addlinespace
1024 & RSGD & 6.4 & 95.2 & 95.2 & 19.38 & 14.04 & 9.84\\
1024 & Exact Newton & 6.4 & 95.2 & 95.2 & 19.38 & 14.04 & 9.84\\
1024 & {Proposed}, $m=1$ & 0.4 & 93.0 & 75.2 & 24.47 & 18.99 & 14.01\\
1024 & {Proposed}, $m=4$ & 6.4 & 94.6 & 89.6 & 21.50 & 15.82 & 11.11\\
1024 & {Proposed}, $m=16$ & 6.4 & 93.6 & 92.4 & 19.92 & 14.33 & 10.16\\
\addlinespace
4096 & RSGD & 6.4 & 96.2 & 96.2 & 9.35 & 6.57 & 4.63\\
4096 & Exact Newton & 6.4 & 96.2 & 96.2 & 9.35 & 6.57 & 4.63\\
4096 & {Proposed}, $m=1$ & 6.4 & 95.6 & 80.0 & 13.58 & 9.28 & 6.61\\
4096 & {Proposed}, $m=4$ & 6.4 & 94.4 & 90.2 & 10.86 & 7.57 & 5.38\\
4096 & {Proposed}, $m=16$ & 6.4 & 95.0 & 94.0 & 9.59 & 6.77 & 4.85\\
\bottomrule\end{tabular}\end{table}
\begin{table}[p]\centering
\caption{Complete product torus results at $\kappa=30$.
Cov. is implemented direction coverage for {proposed} and score-sandwich coverage
for RSGD and exact Newton.  Score is the score only counterfactual for the
same estimate.  All RMSEs are multiplied by $10^3$; fractions refer to
the retained averaging window.  The constant $c$ was selected before
evaluation.  Each row has 500 replications.}
\label{tab:torus-all-kappa30}
\begin{tabular}{rlrrrrrr}
\toprule
& & & \multicolumn{2}{c}{Coverage (\%)} & \multicolumn{3}{c}{RMSE $\times10^3$}\\
\cmidrule(lr){4-5}\cmidrule(lr){6-8}
$n$ & Method & $c$ & Cov. & Score & 25\% & 50\% & 100\%\\
\midrule
256 & RSGD & 1.6 & 75.4 & 75.4 & 87.75 & 52.02 & 29.60\\
256 & Exact Newton & 6.4 & 96.6 & 96.6 & 36.18 & 26.87 & 18.85\\
256 & {Proposed}, $m=1$ & 0.8 & 95.4 & 78.4 & 46.75 & 36.03 & 26.94\\
256 & {Proposed}, $m=4$ & 6.4 & 96.4 & 93.6 & 42.95 & 31.40 & 21.60\\
256 & {Proposed}, $m=16$ & 6.4 & 98.4 & 97.4 & 37.41 & 27.80 & 19.21\\
\addlinespace
1024 & RSGD & 0.4 & 96.0 & 96.0 & 19.43 & 13.78 & 9.56\\
1024 & Exact Newton & 0.4 & 94.8 & 94.8 & 18.16 & 13.65 & 9.56\\
1024 & {Proposed}, $m=1$ & 1.6 & 95.4 & 78.6 & 26.23 & 18.64 & 13.40\\
1024 & {Proposed}, $m=4$ & 0.4 & 95.0 & 91.6 & 20.12 & 15.03 & 10.77\\
1024 & {Proposed}, $m=16$ & 0.4 & 94.2 & 93.6 & 18.64 & 14.08 & 9.86\\
\addlinespace
4096 & RSGD & 0.4 & 95.4 & 95.4 & 9.20 & 6.67 & 4.67\\
4096 & Exact Newton & 0.4 & 95.8 & 95.8 & 8.80 & 6.51 & 4.66\\
4096 & {Proposed}, $m=1$ & 0.4 & 95.4 & 77.8 & 12.24 & 9.13 & 6.62\\
4096 & {Proposed}, $m=4$ & 0.2 & 93.2 & 89.2 & 11.01 & 7.99 & 5.53\\
4096 & {Proposed}, $m=16$ & 3.2 & 95.2 & 93.8 & 9.91 & 7.01 & 4.81\\
\bottomrule\end{tabular}\end{table}
\begin{table}[p]\centering
\caption{Complete product torus results at $\kappa=300$.
Cov. is implemented direction coverage for {proposed} and score-sandwich coverage
for RSGD and exact Newton.  Score is the score only counterfactual for the
same estimate.  All RMSEs are multiplied by $10^3$; fractions refer to
the retained averaging window.  The constant $c$ was selected before
evaluation.  Each row has 500 replications.}
\label{tab:torus-all-kappa300}
\begin{tabular}{rlrrrrrr}
\toprule
& & & \multicolumn{2}{c}{Coverage (\%)} & \multicolumn{3}{c}{RMSE $\times10^3$}\\
\cmidrule(lr){4-5}\cmidrule(lr){6-8}
$n$ & Method & $c$ & Cov. & Score & 25\% & 50\% & 100\%\\
\midrule
256 & RSGD & 0.2 & 0.0 & 0.0 & 450.73 & 412.01 & 347.57\\
256 & Exact Newton & 3.2 & 95.8 & 95.8 & 35.90 & 25.89 & 18.88\\
256 & {Proposed}, $m=1$ & 0.8 & 94.4 & 79.0 & 49.39 & 37.28 & 27.19\\
256 & {Proposed}, $m=4$ & 0.8 & 95.4 & 92.6 & 36.57 & 27.67 & 20.68\\
256 & {Proposed}, $m=16$ & 3.2 & 96.6 & 95.4 & 37.68 & 26.91 & 19.29\\
\addlinespace
1024 & RSGD & 0.4 & 0.0 & 0.0 & 252.63 & 186.01 & 114.03\\
1024 & Exact Newton & 6.4 & 95.4 & 95.4 & 18.53 & 13.49 & 9.65\\
1024 & {Proposed}, $m=1$ & 1.6 & 94.0 & 76.6 & 26.39 & 18.86 & 13.92\\
1024 & {Proposed}, $m=4$ & 0.4 & 94.8 & 90.6 & 19.26 & 14.48 & 10.86\\
1024 & {Proposed}, $m=16$ & 3.2 & 95.0 & 94.8 & 19.01 & 13.73 & 9.94\\
\addlinespace
4096 & RSGD & 0.8 & 41.2 & 41.2 & 32.81 & 17.77 & 9.55\\
4096 & Exact Newton & 6.4 & 95.4 & 95.4 & 9.40 & 6.60 & 4.70\\
4096 & {Proposed}, $m=1$ & 0.2 & 93.4 & 74.0 & 13.13 & 9.57 & 6.88\\
4096 & {Proposed}, $m=4$ & 0.4 & 96.2 & 92.0 & 9.81 & 7.25 & 5.25\\
4096 & {Proposed}, $m=16$ & 0.4 & 95.4 & 94.6 & 9.18 & 6.67 & 4.80\\
\bottomrule\end{tabular}\end{table}

\clearpage

\subsection{Computing times}\label{app:timing}
Serial timing favors the inexpensive diagonal exact solve.  For example,
at $\kappa=300$ and $n=256$, RSGD and exact Newton each take about
$0.045$ ms per complete path, compared with $0.104$ ms for {proposed} with $m=4$.
These measurements include data generation and one-pass inference, using
aggregated coordinate counts for {proposed}.  The observed gain is therefore in
sample-budget performance relative to RSGD, rather than computation per
observation.

\Cref{tab:torus-time} gives medians across 30 blocks of 100 complete paths
per cell, with randomized method order within each block.  These times
include observation generation, updates, averaging, covariance tracking,
and numeric output construction.
Coordinate counts are sampled as binomials, so these timings describe an
aggregated coordinate solver, not $m$ general matrix solves.
\begin{table}[tbp]\centering
\caption{Product torus median serial time per complete path, in milliseconds.
Medians are over 30 randomized-order blocks of 100 paths per cell.  Timing
includes observation generation and uses aggregated coordinate sketches.}
\label{tab:torus-time}
\begin{tabular}{rrrrrrr}
\toprule
$\kappa$ & $n$ & RSGD & Exact Newton & {Proposed}, $m=1$ & {Proposed}, $m=4$ & {Proposed}, $m=16$\\
\midrule
1 & 256 & 0.045 & 0.046 & 0.068 & 0.102 & 0.205\\
1 & 1024 & 0.182 & 0.180 & 0.257 & 0.407 & 0.799\\
1 & 4096 & 0.711 & 0.719 & 1.023 & 1.581 & 3.191\\
30 & 256 & 0.045 & 0.046 & 0.066 & 0.105 & 0.207\\
30 & 1024 & 0.176 & 0.177 & 0.262 & 0.403 & 0.803\\
30 & 4096 & 0.692 & 0.708 & 1.020 & 1.590 & 3.181\\
300 & 256 & 0.045 & 0.045 & 0.067 & 0.104 & 0.206\\
300 & 1024 & 0.182 & 0.176 & 0.259 & 0.401 & 0.806\\
300 & 4096 & 0.708 & 0.704 & 1.020 & 1.592 & 3.197\\
\bottomrule\end{tabular}\end{table}

\subsection{WDBC analysis details}\label{app:wdbc-details}

The Wisconsin Diagnostic Breast Cancer data provide a heterogeneous stream
of correlated image measurements.  The official UCI release contains 569
aspirate records and 30 continuous features derived from digitized images of
cell nuclei, including measures of size, texture, boundary irregularity,
smoothness, symmetry, and fractal dimension
\citep{UCIWDBC,WolbergEtAl1994}.  It reports no missing values and distributes
the data under CC BY 4.0.  The analysis is an unsupervised
dimension reduction illustration, not an evaluation of diagnostic accuracy.

We allocate 169 label free records to calibration and stream the remaining
400.  Calibration means and standard deviations define the feature scale;
the calibration covariance supplies the initial line and a fixed
inverse eigengap preconditioner.  Diagnosis labels play no role in either
step.  The stream uses $\eta_k=0.45k^{-0.65}$, a maximum Grassmann step of
$0.30$ radians, and a 50-observation burn-in.  Conditional on calibration,
the target is the leading eigenspace of the second moment matrix about the
fixed calibration center.  Pointwise Wald intervals use the implemented
direction covariance and the delta method under a working i.i.d. model for
the analysis records.  After burn-in, the average, direction covariance, and
analysis-stream Hessian accumulator are restarted; the Wald multiplier is the
$350$ retained observations.  The Hessian estimate is the retained average of
the PCA score derivative, distinct from the fixed calibration preconditioner
used in the updates.

Individual loading signs are arbitrary, so we summarize the leading
rank one projector ${{\mathrm{\Pi}}}$ by its mass on five scientifically interpretable
feature groups that partition the 30 standardized coordinates.  If ${{\mathrm{D}}}_g$ is
the coordinate-selection projector for group $g$, its mass is
$\theta_g=\tr({{\mathrm{D}}}_g{{\mathrm{\Pi}}})$, so $\sum_g\theta_g=\tr({{\mathrm{\Pi}}})=1$.
Boundary irregularity has estimated mass $0.455$ with interval
$[0.390,0.521]$, while the size group has mass $0.392$ with interval
$[0.254,0.530]$ (\Cref{tab:wdbc-groups}).  Texture contributes relatively
little, with mass $0.023$ and interval $[0.005,0.041]$.  These quantities
describe second moment structure in this sample; they are not causal effects
of cell morphology.

\begin{table}[tbp]
\centering
\caption{WDBC leading-projector mass by morphology-based feature group.  Intervals
are pointwise 95\% delta-method intervals under the conditional working model.}
\label{tab:wdbc-groups}

\begin{tabular}{lrr}
\toprule
Feature group & Projector mass & 95\% interval\\
\midrule
Size (radius/perimeter/area) & 0.392 & [0.254, 0.530]\\
Boundary irregularity & 0.455 & [0.390, 0.521]\\
Texture & 0.023 & [0.005, 0.041]\\
Smoothness and symmetry & 0.092 & [0.041, 0.143]\\
Fractal dimension & 0.038 & [0.004, 0.071]\\
\bottomrule
\end{tabular}

\end{table}

\section{Theory supplements and proofs}\label{app:theory-details}
This appendix develops the local conditions and implementation underlying
the main results, then gives solver comparisons and model-specific criteria.
\Cref{app:all-proofs} contains the proofs and explicit constructions.

\subsection{Averaged randomized Newton inference}\label{app:implementation-details}

The common tangent representation links the randomized recursion to the
influence expansion.  It also specifies the directions and derivative
quantities needed for one-pass inference.

\subsubsection{Target and common tangent representation}

Let $(\M,g)$ be a $d$-dimensional Riemannian manifold.  A population
estimating field assigns a tangent vector $G(x)\in T_x\M$ to each $x$ in
its domain.  We study a selected root $x_\star$ satisfying $G(x_\star)=0$.
For optimization, $G=\operatorname{grad}F$; more generally, $G$ may be a
population estimating equation.  Its local differentiability and
nondegeneracy are specified in \Cref{ass:local-field}.

Let $R$ be a $C^2$ retraction and put $E=T_{x_\star}\M$.  Fix a sufficiently
small working neighborhood $U$ whose closure is contained in a larger
common domain of the retractions and local inverse retractions used below.
The domain is chosen before the reported stochastic approximation phase.
Choose it inside a normal neighborhood of $x_\star$; the
coordinate used here is $R_{x_\star}^{-1}(x)\in E${, with $u_k=R_{x_\star}^{-1}(x_k)$;}
for a generic retraction it need not equal $\Log_{x_\star}(x)$.
All local derivative bounds are taken on compact subsets of the larger
domain, and all inverse-retraction expressions below are restricted to
that common domain.

Fix a smooth {orthonormal frame} $\mathfrak e_x:E\to T_x\M$ with
$\mathfrak e_{x_\star}={{\mathrm{I}}}$, and define
$\T_{x\to y}=\mathfrak e_y\mathfrak e_x^*$.
This identification is an isometry and satisfies
$\T_{y\to z}\T_{x\to y}=\T_{x\to z}$; it is not generally direct
Levi-Civita parallel transport from $x$ to $y$.  A computable frame may be
fixed using the public anchor: identifying its coordinates with $E$ is
only an analytical change of basis and does not require knowledge of
$x_\star$ to compute $\T_{x\to y}$.
For an operator ${{\mathrm{L}}}$, $L^*$ denotes its adjoint.  For vectors in a common
inner-product space, $(v\ot w)z=v\ip{w}{z}$.
We use $\norm{\cdot}$ for the tangent norm, $\norm{\cdot}_{\op}$ for
the induced operator norm, and $\norm {{\mathrm{L}}}_{\mathrm F}^2=\tr(L^*{{\mathrm{L}}})$ for the
Frobenius norm.  For a random operator,
$\norm {{\mathrm{L}}}_{L^2(\op)}=\{\E\norm {{\mathrm{L}}}_{\op}^2\}^{1/2}$, with analogous
notation for the Frobenius and vector norms.

\subsubsection{Recursion and online average}

At time $k+1$, let $\widehat G_{k+1}(x_k)$ be a stochastic score and let
${{\mathrm{P}}}_{k+1}:T_{x_k}\M\to T_{x_k}\M$ be the random linear operator produced by an inner solver.  The update is
\begin{equation}\label{eq:update-main}
 x_{k+1}=R_{x_k}\left[-\eta_{k+1}
 \{{{\mathrm{P}}}_{k+1}\widehat G_{k+1}(x_k)+e_{k+1}\}\right],
 \qquad
 \eta_k=\eta_0 k^{-\alpha},\quad \alpha\in(1/2,1),
\end{equation}
where $e_{k+1}$ is the discrepancy between the implemented direction and a chosen linear reference.  If an inner algorithm is initialized at zero and uses coefficients independent of the current right hand side, a fixed number of linear iterations is exactly represented by ${{\mathrm{P}}}_{k+1}$ and $e_{k+1}=0$.  Data-dependent stopping or residual-based acceleration can be handled by retaining a linear reference and placing the remaining error in $e_{k+1}$.

Two online averages are considered.  The recursive Riemannian average is
\begin{equation}\label{eq:riemannian-average}
 \bar x_1^{\mathrm{rec}}=x_1,
 \qquad
 \bar x_k^{\mathrm{rec}}=R_{\bar x_{k-1}^{\mathrm{rec}}}\left[
  \frac1kR_{\bar x_{k-1}^{\mathrm{rec}}}^{-1}(x_k)\right],
 \qquad k\ge2.
\end{equation}
A fixed anchor chart average, which is often simpler for streaming inference,
is
\begin{equation}\label{eq:anchor-average}
 y_1^a=R_a^{-1}(x_1),
 \qquad
 y_k^a=\left(1-\frac1k\right)y_{k-1}^a
       +\frac1kR_a^{-1}(x_k),
 \qquad
 \bar x_k^a=R_a(y_k^a).
\end{equation}
The anchor $a$ is fixed before the reported averaging phase.  The main theorem applies once the first order equivalence in
\Cref{ass:averaging} is verified.  For both averages, a common-domain $C^2$
retraction and the state rate suffice on the fixed sufficiently small
working domains in \Cref{lem:anchor-average-equivalence,lem:average-equivalence}.

A deterministic burn-in may be omitted when its cumulative state contribution is root-$n$ negligible.  In particular, under \eqref{eq:state-second}, omitting the first $k_0=\lfloor n^{1/2}\rfloor$ iterates is first order negligible because $n^{-1/2}\sum_{k\le k_0}\E\norm{u_k}=O(n^{-\alpha/4})$.
For a proportional burn-in, restarting the average defines a tail estimator
rather than a negligible deletion; \Cref{prop:tail-average} gives its
root-$N_n$ limit in terms of the retained count.

Let $\F_k$ contain the initial point, observations, derivative information,
iterates, and solver randomness through time $k$.  The current score and new
solver seed are not included in $\F_k$.  Their joint conditional law may be
dependent unless a sample split is imposed.  All current scores, solver
operators, inexactness variables, and updated iterates are assumed
$\F_{k+1}$-measurable, and every conditional expectation displayed below is
well defined and integrable to the stated order.

\subsubsection{One-pass covariance tracking}

The following implementation uses a fixed public anchor $a\in U$.  All
running matrices live in $T_a\M$, while actual directions are mapped to
$T_{x_k}\M$.  The displayed derivative branch is for a gradient field, where
the Jacobian is a self-adjoint Hessian and spectral safeguarding is meaningful.
For a generic estimating field with a possibly nonsymmetric derivative ${{\mathrm{H}}}$,
the same statistical theorem applies only after replacing this branch by a
consistent Jacobian estimator and an invertibility/stability safeguard
appropriate to nonsymmetric operators.  The Hessian and score mini-batches
may be literal independent splits, cross-fitted streams, or any construction
satisfying the product conditions below.
The map $\mathcal K_J({{\mathrm{B}}},\Omega)$ sends a right hand side to the output of
$J$ zero-start linear-solver steps, with coefficients fixed by the curvature
operator ${{\mathrm{B}}}$ and seed $\Omega$.  Right-hand-side-dependent coefficients,
including ordinary finite-step conjugate gradients, require an explicit
linear reference and an error term.
For a self-adjoint operator, $\operatorname{SpecClip}_{[h_-,h_+]}$
clips its eigenvalues to the specified positive interval.  The raw
Hessian average is stored separately from this safeguarded estimate.
For a vector $z$, the Welford update of a count-$k$ mean $\mu$
and centered sum of squares ${{\mathrm W}}$ is
\[
\delta=z-\mu,\qquad
\mu^+=\mu+\delta/(k+1),\qquad
{{\mathrm W}}^+={{\mathrm W}}+\delta\ot(z-\mu^+).
\]
The localization map $\mathcal L$ is either the identity on an independently
verified invariant basin, or a specified map with its own drift and error
proof.  It must leave the actual direction in
the common inverse-retraction domain and satisfy the actual-step and
error conditions used by the theorem.  A step cap alone does not supply
this property.  The predictable covariance branch is used only under
\Cref{thm:predictable,cor:predictable-covariance}.

For $c>0$, radial clipping is
$\operatorname{clip}_c(z)=\min\{1,c/\norm z\}z$ for $z\ne0$, and
$\operatorname{clip}_c(0)=0$.

\begin{algorithm}[p]
\caption{Riemannian Newton averaging with online inference}\label{alg:main}
\begin{algorithmic}[1]
\Require $x_0$, public anchor $a$, horizon $n$, common-domain $C^2$
retraction and coherent frame; averaging rule; step sizes $\eta_k$;
solver budgets $J_k,m_k$ and specified inner-error rule; ridge $\lambda_k$; spectral band $[h_-,h_+]$;
safeguarded $\widehat {{\mathrm{H}}}_0^a$; step cutoff $r_0$ and justified localization
map $\mathcal L$; fresh or admissible predictable covariance regime.
\State Initialize the raw Hessian mean ${{\mathrm M}}_0^a=0$, solver mean $\widehat {{\mathrm P}}_0^a=0$, and Welford state $(\mu_0,{{\mathrm W}}_0)=(0,0)$.
\State \textbf{for} $k=0,\ldots,n-1$ \textbf{do}
\State \hspace{1em}Set
      ${{\mathrm{B}}}_k^{(x_k)}=\T_{a\to x_k}\widehat {{\mathrm{H}}}_k^a\T_{x_k\to a}$.
\State \hspace{1em}\textbf{if} the fresh solver regime is selected
      \textbf{then}
\State \hspace{2em}Draw $m_k$ conditionally independent solver seeds,
      independent of the current observation.
\State \hspace{2em}Set
      ${{\mathrm{P}}}_{k+1}=m_k^{-1}\sum_{j=1}^{m_k}
       \mathcal K_{J_k}({{\mathrm{B}}}_k^{(x_k)},\Omega_{k+1}^{(j)})$.
\State \hspace{1em}\textbf{else} choose the specified invertible
      predictable ${{\mathrm{P}}}_{k+1}$, for example $({{\mathrm{B}}}_k^{(x_k)})^{-1}$.
\State \hspace{1em}\textbf{end if}
\State \hspace{1em}Obtain the current score $\widehat G_{k+1}(x_k)$ and
      derivative contribution $\widehat {{\mathrm{B}}}_{k+1}$.
\State \hspace{1em}Form
      $D_{k+1}^0={{\mathrm{P}}}_{k+1}\widehat G_{k+1}(x_k)+e_{k+1}^{\mathrm{inner}}$
      and $D_{k+1}^{\mathrm{cap}}=
      \operatorname{clip}_{r_0/\eta_{k+1}}(D_{k+1}^0)$.
\State \hspace{1em}Set
      $x_{k+1}=\mathcal L\bigl(R_{x_k}[-\eta_{k+1}
                      D_{k+1}^{\mathrm{cap}}]\bigr)$ and
      $D_{k+1}=-\eta_{k+1}^{-1}R_{x_k}^{-1}(x_{k+1})$.
\State \hspace{1em}\textbf{if} $k=0$ \textbf{then} initialize the average
      at $x_1$; \textbf{else} update it by
      \eqref{eq:anchor-average} or \eqref{eq:riemannian-average}.
\State \hspace{1em}Set $\widehat {{\mathrm{B}}}_{k+1}^a=
      \T_{x_k\to a}\widehat {{\mathrm{B}}}_{k+1}\T_{a\to x_k}$ and
      ${{\mathrm{M}}}_{k+1}^a=(k{{\mathrm{M}}}_k^a+\widehat {{\mathrm{B}}}_{k+1}^a)/(k+1)$.
\State \hspace{1em}Set
      $\widehat {{\mathrm{H}}}_{k+1}^a=\operatorname{SpecClip}_{[h_-,h_+]}
      ({{\mathrm{M}}}_{k+1}^a+\lambda_{k+1}{{\mathrm{I}}})$.
\State \hspace{1em}\textbf{if} the fresh solver regime is selected
      \textbf{then}
\State \hspace{2em}Set
      $\widehat {{\mathrm{P}}}_{k+1}^a=
      \{k\widehat {{\mathrm{P}}}_k^a+\T_{x_k\to a}{{\mathrm{P}}}_{k+1}\T_{a\to x_k}\}/(k+1)$
      and $z_{k+1}=\T_{x_k\to a}D_{k+1}$.
\State \hspace{1em}\textbf{else} set
      $z_{k+1}=\T_{x_k\to a}\widehat G_{k+1}(x_k)$.
\State \hspace{1em}\textbf{end if}
\State Apply the count-$k$ Welford update to$z_{k+1}$, obtaining $(\mu_{k+1},{{\mathrm W}}_{k+1})$.
\State \textbf{end for}
\State \textbf{if} fresh \textbf{then} set
      $\widehat {{\mathrm{A}}}_n^a=\widehat {{\mathrm{P}}}_n^a\widehat {{\mathrm{H}}}_n^a$ and, when invertible,
$\widehat {{\mathrm V}}_n^a
=(\widehat {{\mathrm A}}_n^a)^{-1}
({{\mathrm W}}_n/n)
{((\widehat {\mathrm A}_n^a)^{*-1})}.$
\State \textbf{else} set
      $\widehat {{\mathrm V}}_n^a
=(\widehat {{\mathrm H}}_n^a)^{-1}
({{\mathrm W}}_n/n)
{((\widehat {\mathrm H}_n^a)^{*-1})}$.
\State \textbf{end if}
\State \textbf{return} $\bar x_n$, count $n$, and
      $\T_{a\to\bar x_n}\widehat {{\mathrm{V}}}_n^a\T_{\bar x_n\to a}$;
      report unavailable covariance if the required inverse does not exist.
\end{algorithmic}
\end{algorithm}

The full discrepancy in \eqref{eq:update-main} is
$e_{k+1}=D_{k+1}-{{\mathrm{P}}}_{k+1}\widehat G_{k+1}(x_k)$, including inner error,
clipping, and localization.  The displayed Hessian branch concerns a
gradient field; a nonsymmetric estimating field requires a consistent
Jacobian estimator and an appropriate alternative safeguard.
Full dimensional Wald inference additionally requires a positive definite
reported covariance and the relevant inverse-retraction chart.  A singular
covariance can still be returned; a scalar interval requires a positive
estimated variance for its contrast.
For deterministic burn-in, keep the global step size index and restart
reporting accumulators with the retained count.  If the optimizer learns
its Hessian before burn-in, retain its optimizer state separately from a
restarted reporting Hessian average.  A full fresh solver covariance
requires access to a consistent mean solver estimator, not only the
ability to apply the solver to a vector.

When the update uses a predictable preconditioner that is not assumed to
converge, the mean operator branch in the final covariance formula is not
appropriate.  In that regime, use the raw transported score
\begin{equation}\label{eq:raw-score-anchor}
 G_{k+1}^a=\T_{x_k\to a}\widehat G_{k+1}(x_k)
\end{equation}
as input to the Welford accumulator.  Denote its covariance by
$\widehat {{\mathrm S}}_n^a={{\mathrm W}}_n/n.$ and report
\begin{equation}\label{eq:predictable-cov-estimator-algorithm}
 \widehat {{\mathrm{V}}}_n^{a,\mathrm{pred}}
 = (\widehat {{\mathrm{H}}}_n^a)^{-1}\widehat {{\mathrm{S}}}_n^a{((\widehat {\mathrm{H}}_n^a)^{*-1})}.
\end{equation}
\Cref{cor:predictable-covariance} shows that this branch remains consistent
when the predictable preconditioner has no fixed limit.  The direction
covariance branch is used when the solver is randomized at the current
iteration because it retains the corresponding product noise.

Algorithm \ref{alg:main} stores a constant number of $d\times d$ operators,
requiring $O(d^2)$ memory.  A matrix-free direction routine does not by
itself provide the full mean solver estimator.  Structured or contrast-specific
tracking requires a consistent estimator of the corresponding covariance;
a diagonal approximation is justified only when the target structure or
the chosen contrast makes it sufficient.

\subsubsection{Local conditions for inference}\label{sec:assumptions}
The local conditions separate the target field, the current solver score
product, and the errors that accumulate under averaging.  The state rate
criterion in \Cref{prop:state-rate} is verified directly from the actual
update.  The geometric and averaging propositions then supply the corresponding
remainder conditions under their explicit domain and moment hypotheses.

\begin{assumption}
\label{ass:local-field}
The reported phase starts at a deterministic $x_0\in U$, and the actual
updates remain in the fixed working neighborhood $U$ almost surely.  The
reported average belongs to the common inverse-retraction domain with
probability tending to one.  On $U$,
\begin{equation}\label{eq:field-linearization}
 \T_{x\to x_\star}G(x)={{\mathrm{H}}}{R_{x_\star}^{-1}(x)}+r_G(x),\qquad
 \norm{r_G(x)}=o(\norm{{R_{x_\star}^{-1}(x)}})\quad(x\to x_\star),
\end{equation}
where ${{\mathrm{H}}}:E\to E$ is invertible.  The working domain satisfies the uniform
bound $\norm{{{\mathrm{H}}}{R_{x_\star}^{-1}(x)}+r_G(x)}\le C\norm{{R_{x_\star}^{-1}(x)}}$ for $x\in U$.
\end{assumption}

Differentiability at the root gives the uniform field bound on a
sufficiently small neighborhood, which must be selected before imposing
localization.  Alternatively, continuity on a compact larger domain,
together with differentiability at the root, gives this bound on a fixed
compact working basin.  A step-length cap alone does not prove basin
invariance; that property is supplied by a model-specific argument or an
explicit localization mechanism.  An independent initialization phase may
be conditioned on, but global root selection is not part of the local
inference theorem.  At a nondegenerate local minimum of a twice
differentiable objective, ${{\mathrm{H}}}$ is positive definite; a strict minimum alone
does not imply this property.

\begin{assumption}\label{ass:score}
For $x_k\in U$,
\[
 \T_{x_k\to x_\star}\widehat G_{k+1}(x_k)
 ={{\mathrm{H}}}u_k+r_G(x_k)+\xi_{k+1}+q_{k+1},
\]
where the target score $\xi_{k+1}\in E$ and centered score increment
$q_{k+1}\in E$ satisfy
\begin{equation}\label{eq:xi-q-center}
 \E(\xi_{k+1}\mid\F_k)=0,\qquad
 \E(q_{k+1}\mid\F_k)=0.
\end{equation}
\end{assumption}

For fresh i.i.d. observations and a tangent-valued sample field
$\mathsf g(x,Z)$ with $\E\mathsf g(x,Z)=G(x)$, a consistent choice is
\[
 \xi_{k+1}=\mathsf g(x_\star,Z_{k+1}),\qquad
 q_{k+1}=\T_{x_k\to x_\star}\mathsf g(x_k,Z_{k+1})
 -\mathsf g(x_\star,Z_{k+1})-\T_{x_k\to x_\star}G(x_k).
\]
Every term is in $E$.  An $L^2$ local derivative envelope for the
transported sample field gives
$\E(\norm{q_{k+1}}^2\mid\F_k)\le C\norm{u_k}^2$.
To obtain a bound for its product with the solver, one also needs the
operator or joint-product conditions below.

Write
\[
 {{\widetilde{\mathrm{P}}}}_{k+1}=\T_{x_k\to x_\star}{{\mathrm{P}}}_{k+1}\T_{x_\star\to x_k},
 \qquad {{\mathrm{\Pi}}}_k=\E({{\widetilde{\mathrm{P}}}}_{k+1}\mid\F_k),
\]
and define the two conditional product means and their centered parts by
\begin{align}
 b_{k+1}^{\xi}&=\E({{\widetilde{\mathrm{P}}}}_{k+1}\xi_{k+1}\mid\F_k),
 &\zeta_{k+1}&={{\widetilde{\mathrm{P}}}}_{k+1}\xi_{k+1}-b_{k+1}^{\xi},
 \label{eq:zeta-def}\\
 b_{k+1}^{q}&=\E({{\widetilde{\mathrm{P}}}}_{k+1}q_{k+1}\mid\F_k),
 &\chi_{k+1}&={{\widetilde{\mathrm{P}}}}_{k+1}q_{k+1}-b_{k+1}^{q}.
 \label{eq:chi-def}
\end{align}

\begin{assumption}\label{ass:fresh-solver}
There is a deterministic invertible $\bar {{\mathrm{P}}}:E\to E$ such that
${{\mathrm{A}}}=\bar {{\mathrm{P}}} {{\mathrm{H}}}$ is positive stable, meaning that all its eigenvalues have
positive real part.  For a deterministic constant $C$,
\begin{align}
 \sup_k\E(\norm{{{\widetilde{\mathrm{P}}}}_{k+1}}_{\op}^2\mid\F_k)&\le C,
 \label{eq:P-moment-new}\\
 \E(\norm{\chi_{k+1}}^2\mid\F_k)&\le C\norm{u_k}^2.
 \label{eq:chi-moment}
\end{align}
The population remainders satisfy
\begin{align}
 \frac1{\sqrt n}\sum_{k=0}^{n-1}({{\mathrm{\Pi}}}_k-\bar {{\mathrm{P}}}){{\mathrm{H}}}u_k&\pto0,
 \label{eq:mean-P-sum}\\
 \frac1{\sqrt n}\sum_{k=0}^{n-1}{{\mathrm{\Pi}}}_kr_G(x_k)&\pto0.
 \label{eq:field-remainder-sum}
\end{align}
\end{assumption}

Conditional independence of the current solver from the whole current
observation is one sufficient route to product centering and moment
bounds; it is not required by the theorem.  Separate unbiasedness or
separate second moments do not, by themselves, establish the corresponding
properties of a dependent product.  Given the state rate below,
$\norm{{{\mathrm{\Pi}}}_k-\bar {{\mathrm{P}}}}_{L^2(\op)}=O(k^{-\delta})$ with
$\delta>(1-\alpha)/2$, and
$\norm{r_G(x)}\le C\norm{{R_{x_\star}^{-1}(x)}}^{1+\rho}$ with
$\alpha(1+\rho)>1$, are sufficient for the two accumulated remainders.
The direct sum conditions permit weaker routes when additional
cancellation is available.

Decompose the complete implementation discrepancy, including all
safeguard modifications, as
\begin{equation}\label{eq:e-decomp-new}
 \widetilde e_{k+1}=\T_{x_k\to x_\star}e_{k+1}=a_{k+1}+\nu_{k+1},
 \quad a_{k+1}=\E(\widetilde e_{k+1}\mid\F_k),
 \quad\E(\nu_{k+1}\mid\F_k)=0.
\end{equation}
Put $\beta_{k+1}=b_{k+1}^{\xi}+b_{k+1}^{q}+a_{k+1}$.

\begin{assumption}
\label{ass:bias-inexact}
For a deterministic $b\in E$,
\begin{equation}\label{eq:rootn-bias}
 \frac1{\sqrt n}\sum_{k=0}^{n-1}\beta_{k+1}\pto b,
\end{equation}
and
\begin{equation}\label{eq:centered-inexact-small}
 \frac1n\sum_{k=0}^{n-1}\E(\norm{\nu_{k+1}}^2\mid\F_k)\pto0.
\end{equation}
\end{assumption}

The limit $b$ determines a possible Gaussian centering shift.  A centered
inner error with nonvanishing average variance must instead be included
in the leading innovation and its covariance; it cannot be discarded
merely because its conditional mean is zero.

\begin{assumption}
\label{ass:geometry}
In the coordinate $u$, the actual update admits
\begin{align}
 u_{k+1}
 &=u_k-\eta_{k+1}\Big[{{\widetilde{\mathrm{P}}}}_{k+1}
       \{{{\mathrm{H}}}u_k+r_G(x_k)+\xi_{k+1}+q_{k+1}\}
       +a_{k+1}+\nu_{k+1}\Big]\nonumber\\
 &\quad+\eta_{k+1}c_{k+1}+\eta_{k+1}d_{k+1},
 \label{eq:fixed-tangent-raw}
\end{align}
where $\E(c_{k+1}\mid\F_k)=0$ and
\begin{equation}\label{eq:geometry-sums}
 \frac1n\sum_{k=0}^{n-1}\E(\norm{c_{k+1}}^2\mid\F_k)\pto0,
 \qquad
 \frac1{\sqrt n}\sum_{k=0}^{n-1}d_{k+1}\pto0.
\end{equation}
The remainder $d_{k+1}$ need not be predictable.
\end{assumption}

These are accumulated error conditions, not a consequence of smoothness
alone.  A common-domain $C^2$ retraction and frame, the state rate, actual
bounded steps, and the direction and bias moments in
\Cref{prop:geometry-primitive} imply them.

\begin{assumption}
\label{ass:averaging}
With $\widetilde u_n=n^{-1}\sum_{k=1}^nu_k$, the chosen average satisfies
\begin{equation}\label{eq:assumed-average-equivalence}
 \sqrt n\{R_{x_\star}^{-1}(\bar x_n)-\widetilde u_n\}\pto0.
\end{equation}
\end{assumption}

For the fixed anchor and recursive averages, the state rate and common-domain
$C^2$ regularity on the fixed sufficiently small domains imply this equivalence by
\Cref{lem:anchor-average-equivalence,lem:average-equivalence}.  No locally
Lipschitz second derivative or third derivative is required for either
rule.

\begin{assumption}
\label{ass:state-clt}
The local state satisfies
\begin{equation}\label{eq:state-second}
 \E\norm{u_k}^2\le C\eta_k.
\end{equation}
For a deterministic positive semidefinite operator ${{\mathrm{Q}}}:E\to E$,
\begin{equation}\label{eq:Q-limit-new}
 \frac1n\sum_{k=0}^{n-1}
 \E(\zeta_{k+1}\ot\zeta_{k+1}\mid\F_k)\pto {{\mathrm{Q}}},
\end{equation}
and, for every $\epsilon>0$,
\begin{equation}\label{eq:conditional-lindeberg}
 \frac1n\sum_{k=0}^{n-1}\E\!\left[
 \norm{\zeta_{k+1}}^2
 \ind\{\norm{\zeta_{k+1}}>\epsilon\sqrt n\}\mid\F_k\right]\pto0.
\end{equation}
\end{assumption}

The state rate follows from the direct Lyapunov criterion in
\Cref{prop:state-rate}; it is not used to prove that criterion's one step
drift inequality.  The covariance limit and Lindeberg condition are
separate noise requirements.  A uniform conditional $(2+\tau)$ moment
of $\zeta_{k+1}$ is sufficient for Lindeberg, but the main theorem does not
require a fixed moment above two.  Positive semidefinite ${{\mathrm{Q}}}$ permits a
degenerate Gaussian limit; full dimensional Wald inversion requires
positive definiteness.

In Euclidean stochastic approximation with $R_x(v)=x+v$ and identity
transport, the geometric errors vanish and both averages are arithmetic.
The remaining restrictions concern stable drift, martingale noise,
accumulated solver/error bias, and feasible operator estimation.  On a
manifold, common coordinate domains and quadratic geometric remainders
are additionally needed.  The verification order is state rate, population
and geometric remainders, average equivalence, the CLT, and covariance
consistency.  \Cref{ex:invariant-quadratic} gives a fully checkable instance;
the local theorem does not assert that an arbitrary optimizer selects or
remains in the required basin.

\subsection{Theoretical results}\label{app:solver-results}

\subsubsection{Influence representation}

The local recursion has drift ${{\mathrm{A}}}=\bar {{\mathrm{P}}} {{\mathrm{H}}}$, but its leading innovation is
not $\bar {{\mathrm{P}}}\xi_{k+1}$.  It is the centered realized product
$\zeta_{k+1}$ in \eqref{eq:zeta-def}.  The representation in \Cref{thm:main} connects that temporal distinction to inference.

The main-text statement is \Cref{thm:main}.

The mean pull toward the root is governed by ${{\mathrm{A}}}=\bar {{\mathrm{P}}} {{\mathrm{H}}}$, while the
leading fluctuation is the realized product of the solver and target score.
Replacing this product by the mean solver applied to the score would remove
randomness that the algorithm introduces.  The limit $b$ records conditional
product or implementation bias that survives at the root-$n$ scale.  Its
mean shift $-{{\mathrm{A}}}^{-1}b$ cannot be removed by covariance estimation alone:
centered Wald inference requires $b=0$, or an available consistent estimate
of $b$ and the plus-sign correction in \eqref{eq:bias-corrected-average}.

No H\"older exponent, polynomial solver rate, or moment above order two is
primitive to the theorem.  \Cref{prop:remainder-rates} records convenient
rate conditions for the two accumulated remainders, and
\Cref{prop:state-rate} supplies the mean-square state rate without a
fourth moment recursion.

\begin{corollary}\label{cor:centered}
Under \Cref{thm:main}, if
\[
 \frac1{\sqrt n}\sum_{k=0}^{n-1}
 \{b_{k+1}^{\xi}+b_{k+1}^{q}+a_{k+1}\}\pto0,
\]
then
\[
 \sqrt n\,R_{x_\star}^{-1}(\bar x_n)\dto N_E(0,{{\mathrm{V}}}).
\]
In particular, this holds under product orthogonality and root-$n$
negligible inexactness bias.
\end{corollary}

\begin{proposition}\label{prop:tail-average}
Under the centered conditions of \Cref{cor:centered}, let $(b_n)$ be a
deterministic integer sequence with $0\le b_n<n$,
$b_n/n\to\tau\in[0,1)$, and $N_n=n-b_n$.  Restart either online average at
$x_{b_n+1}$ and suppose its tangent coordinate satisfies the tail analogue
of \eqref{eq:assumed-average-equivalence},
\[
 \sqrt{N_n}\left\{
 R_{x_\star}^{-1}(\bar x_{n,b_n})
 -\frac1{N_n}\sum_{k=b_n+1}^{n}u_k
 \right\}\pto0.
\]
Then
\[
 \sqrt{N_n}\,R_{x_\star}^{-1}(\bar x_{n,b_n})
 \dto N_E(0,{{\mathrm{V}}}),\qquad {{\mathrm{V}}}={{\mathrm{A}}}^{-1}{{\mathrm{Q}}}{({\mathrm{A}}^{*-1})}.
\]
The displayed tail equivalence holds for the fixed anchor average under the
same local $C^2$ chart conditions used in
\Cref{lem:anchor-average-equivalence}; it also holds for the restarted
recursive average under the conditions of \Cref{lem:average-equivalence}.
\end{proposition}

\subsubsection{Mean-square localization}

The local rate \eqref{eq:state-second} can be verified without a fourth moment recursion.

\begin{proposition}\label{prop:state-rate}
Let $\bar {{\mathrm{P}}}$ be deterministic with ${{\mathrm{A}}}=\bar {{\mathrm{P}}}{{\mathrm{H}}}$ positive stable.  Suppose
there are a deterministic index $k_\star$ and a deterministic ${{\mathrm{K}}}\succ0$ such
that, on the localized trajectory, almost surely for every $k\ge k_\star$,
\begin{equation}\label{eq:uniform-conditional-stability}
 ({{\mathrm{\Pi}}}_k{{\mathrm{H}}})^*{{\mathrm{K}}}+{{\mathrm{K}}}{{\mathrm{\Pi}}}_k{{\mathrm{H}}}\succeq c_0{{\mathrm{I}}}
\end{equation}
for a deterministic $c_0>0$.
Suppose \Cref{ass:local-field,ass:score} and the error decomposition
\eqref{eq:e-decomp-new} hold.  Without assuming
\Cref{ass:fresh-solver,ass:bias-inexact,ass:geometry,ass:state-clt}, assume that
the retraction and local frame are $C^2$, $\sup_k\norm{{{\mathrm{\Pi}}}_k}_{\op}\le C$,
and the actual direction satisfies
$\eta_{k+1}\norm{D_{k+1}}\le r_0$ and
\eqref{eq:primitive-direction-second}, and
$\E\norm{\beta_{k+1}}^2\le C\eta_{k+1}$ for
$\beta_{k+1}=b_{k+1}^{\xi}+b_{k+1}^{q}+a_{k+1}$.  If $U$ is chosen sufficiently small before imposing the stated localization, then $\E\norm{u_k}^2\le C\eta_k$.
One convenient route is to take ${{\mathrm{K}}}$ as the solution of
${{\mathrm{A}}}^*{{\mathrm{K}}}+{{\mathrm{K}}}{{\mathrm{A}}}={{\mathrm{I}}}$ and impose a deterministic eventual uniform operator-norm
bound on ${{\mathrm{\Pi}}}_k-\bar {{\mathrm{P}}}$.  The rank one closure below instead verifies
\eqref{eq:uniform-conditional-stability} directly with ${{\mathrm{K}}}={{\mathrm{H}}}$.
\end{proposition}

The proposition is deliberately upstream of the geometric reduction:
it uses the direct two-variable retraction expansion and the direction
second moment bound, not \Cref{ass:geometry}.  Once the state rate is known,
\Cref{prop:geometry-primitive} turns the same local expansion into the two
root-$n$ geometric remainder bounds.  This order removes any need to assume
the conclusion of one result while verifying the other.

A local state rate and eventual localization are logically distinct.  The proposition controls the rate once the trajectory is in the stable neighborhood.  Global convergence can be established under the usual coercivity, projection, or truncation conditions for stochastic approximation; such conditions are model-dependent and are not needed for the local influence-function calculation.  Similar localization is standard in Riemannian stochastic approximation \citep{Bonnabel2013,TripuraneniEtAl2018,ZhouSaid2019}.

\subsubsection{Feasible covariance estimation and Wald inference}\label{app:cov-estimation-details}

The influence representation identifies what must be tracked online: the
covariance of the implemented direction in a common frame, together with the
mean solver and derivative.  This is stronger than storing a score covariance
when the current solve is random, but it remains a one-pass calculation.

Let $a\in U$ be the anchor used by \Cref{alg:main}, and define the anchor representations
\begin{align*}
 {{\mathrm{H}}}_a&=\T_{x_\star\to a}{{\mathrm{H}}}\T_{a\to x_\star},
 &\bar {{\mathrm{P}}}_a&=\T_{x_\star\to a}\bar {{\mathrm{P}}}\T_{a\to x_\star},\\
 {{\mathrm{Q}}}_a&=\T_{x_\star\to a}{{\mathrm{Q}}}\T_{a\to x_\star},
 &{{\mathrm{V}}}_a&=(\bar {{\mathrm{P}}}_a{{\mathrm{H}}}_a)^{-1}{{\mathrm{Q}}}_a{((\bar {\mathrm{P}}_a{\mathrm{H}}_a)^{*-1})}.
\end{align*}

The main-text statement is \Cref{ass:cov-estimation}.

The squared bias condition is additional to the signed root-$n$ bias
condition: alternating errors can cancel in an average while remaining
visible in an empirical covariance.  Under the preceding assumptions,
including the uniform field bound on the actual working basin,
\Cref{lem:covariance-closure} derives the required realized-error and
outer-product laws.  No fourth moment is needed for this consistency
argument.  The streaming Hessian and solver propositions provide
sufficient conditions for the two operator estimators.

For the fresh solver branch, define the implemented directions and their
sample covariance in the anchor space by
\begin{align*}
 Y_{k+1}&=\T_{x_k\to a}D_{k+1},
 &\bar Y_n&=\frac1n\sum_{k=0}^{n-1}Y_{k+1},\\
\widehat {{\mathrm Q}}_n^a
&=\frac1n\sum_{k=0}^{n-1}
(Y_{k+1}-\bar Y_n)\ot(Y_{k+1}-\bar Y_n)
={{\mathrm W}}_n/n,
 &\widehat {{\mathrm{A}}}_n^a&=\widehat {{\mathrm{P}}}_n^a\widehat {{\mathrm{H}}}_n^a.
\end{align*}
When $\widehat {{\mathrm{A}}}_n^a$ is invertible, put
$\widehat {{\mathrm{V}}}_n^a=(\widehat {{\mathrm{A}}}_n^a)^{-1}\widehat {{\mathrm{Q}}}_n^a
{((\widehat {\mathrm{A}}_n^a)^{*-1})}$ and
$\widehat {{\mathrm{V}}}_n^{\bar x}=\T_{a\to\bar x_n}\widehat {{\mathrm{V}}}_n^a\T_{\bar x_n\to a}$.
For convergence statements, extend these covariance operators by zero
when the required inverse is unavailable.

The main-text statement is \Cref{thm:cov-estimator}.

The recorded-discrepancy bias construction is given in \Cref{app:recorded-bias}.

The direction covariance, rather than a covariance of the iterates, is the natural estimator for the \linebreak[3]Polyak-Ruppert influence sum.  It automatically includes solver randomness but does not, by itself, estimate the signed local bias $b$.  Thus covariance consistency and bias centering are separate requirements.  In particular, using only an unpreconditioned score covariance and then reporting ${{\mathrm{H}}}^{-1}\widehat {{\mathrm{S}}} {({\mathrm{H}}^{*-1})}$ is inconsistent for a fixed-budget randomized sketch unless the inflation term vanishes.  The estimator in \eqref{eq:Qhat-consistency} also differs from the weighted iterate-covariance estimator used for the last iterate Lyapunov limit of \citet{WangDuNa2026}.

\subsubsection{Deterministic and asymptotically deterministic conditioning}

Let ${{\mathrm{S}}}$ denote the deterministic limit, when it exists, in
\begin{equation}\label{eq:S-def}
 \frac1n\sum_{k=0}^{n-1}
 \E(\xi_{k+1}\ot\xi_{k+1}\mid\F_k)\pto {{\mathrm{S}}}.
\end{equation}

\begin{corollary}\label{cor:sandwich}
Suppose the centered conditions of \Cref{thm:main} hold with $\bar {{\mathrm{P}}}={{\mathrm{P}}}_0$, where ${{\mathrm{P}}}_0$ is deterministic and invertible and ${{\mathrm{P}}}_0{{\mathrm{H}}}$ is positive stable.  Assume
\[
 \frac1n\sum_{k=0}^{n-1}
 \E\{\norm{({{\widetilde{\mathrm{P}}}}_{k+1}-{{\mathrm{P}}}_0)\xi_{k+1}}^2\mid\F_k\}\pto0.
\]
If \eqref{eq:S-def} holds, then
\begin{equation}\label{eq:sandwich-covariance}
 {{\mathrm{V}}}=\Vsand:={{\mathrm{H}}}^{-1}{{\mathrm{S}}}{({\mathrm{H}}^{*-1})}.
\end{equation}
This includes ${{\mathrm{P}}}_0={{\mathrm{I}}}$, exact Newton conditioning ${{\mathrm{P}}}_0={{\mathrm{H}}}^{-1}$ in the optimization case, and every fixed deterministic invertible preconditioner for which ${{\mathrm{P}}}_0{{\mathrm{H}}}$ is positive stable.
\end{corollary}

In the applications below, ``asymptotically deterministic'' means that the
score-product condition in \Cref{cor:sandwich} and the covariance limit
\eqref{eq:S-def} hold, together with the centered conditions of
\Cref{thm:main}.  Convergence ${{\widetilde{\mathrm{P}}}}_{k+1}\pto {{\mathrm{P}}}_0$ alone is insufficient.

This is the established preconditioner-invariance mechanism of averaged
stochastic approximation
\citep{PolyakJuditsky1992,BoyerGodichon2023,AnHuoDynamic2026}; it locates the
additional fresh solver covariance relative to that benchmark.

Under a correctly specified likelihood score, the information identity can
make \eqref{eq:sandwich-covariance} equal to inverse Fisher information.
For a generic loss or estimating equation, the sandwich instead describes
the covariance of the corresponding ${{\mathrm{M}}}$-estimator.

The local contraction comparison is given in \Cref{prop:rsgd-contraction}.

\subsubsection{Randomization in the current solve and covariance inflation}\label{sec:fresh-inflation}

The sandwich cancellation above fails only through directions in which the
solver varies and the target score has energy.  Independence makes this
failure an exact covariance decomposition.

The main-text statement is \Cref{thm:inflation}.

If ${{\mathrm{S}}}$ is positive definite, equality requires ${{\widetilde{\mathrm{P}}}}=\bar {{\mathrm{P}}}$ almost
surely; for singular ${{\mathrm{S}}}$, randomness outside the score's range is harmless.
This comparison fixes the target score and uses independent solver seeds.
Computational benefits of a cheaper solve depend separately on the
finite-budget transient and its implementation cost.

\begin{corollary}\label{cor:multiple-sketch}
Assume the centered conditions of \Cref{thm:main}.  Suppose that,
conditionally on $\F_k$, the target score has the same covariance ${{\mathrm{S}}}$ at
every iteration.  Let $m_k$ be a positive, $\F_k$-measurable integer and let
\[
 {{\widetilde{\mathrm{P}}}}_{k+1}^{(m_k)}
 =\frac1{m_k}\sum_{j=1}^{m_k}{{\widetilde{\mathrm{P}}}}_{k+1}^{(j)},
\]
where the $m_k$ operators are conditionally independent copies from a common
time-homogeneous solver law with mean $\bar {{\mathrm{P}}}$ and variance operator
${{\mathrm{R}}}$ as in \eqref{eq:R-sketch}, independent of the current score.  If
\begin{equation*}
 \frac1n\sum_{k=1}^n\frac1{m_k}\xrightarrow{p}\lambda\in[0,1],
\end{equation*}
then the limiting covariance is
\begin{equation*}
 \Vsand+
 \lambda {{\mathrm{H}}}^{-1}\bar {{\mathrm{P}}}^{-1}{{\mathrm{R}}}{(\bar {\mathrm{P}}^{*-1})}{({\mathrm{H}}^{*-1})}.
\end{equation*}
Thus a fixed $m$ gives the factor $1/m$, while $m_k\xrightarrow{p}\infty$ gives $\lambda=0$ and recovers the sandwich benchmark.  The limiting mean operator need not equal ${{\mathrm{H}}}^{-1}$; it only needs to be invertible and to satisfy the stability conditions.
\end{corollary}

\begin{corollary}
\label{cor:loewner-order}
Under the conditions of \Cref{thm:inflation,cor:multiple-sketch}, define
\[
 {{\mathrm{\Delta}}}={{\mathrm{H}}}^{-1}\bar {{\mathrm{P}}}^{-1}{{\mathrm{R}}}{(\bar {\mathrm{P}}^{*-1})}{({\mathrm{H}}^{*-1})}\succeq0.
\]
For a fixed number $m$ of independent solver replicas,
\begin{equation}\label{eq:V-m-order}
 {{\mathrm{V}}}_m=\Vsand+\frac1m{{\mathrm{\Delta}}}.
\end{equation}
Consequently, if $1\le m_1\le m_2$, then
\[
 {{\mathrm{V}}}_{m_1}\succeq {{\mathrm{V}}}_{m_2}\succeq\Vsand,
 \qquad
 a^*{{\mathrm{V}}}_m a=a^*\Vsand a+\frac1m a^*{{\mathrm{\Delta}}} a
 \quad\text{for every }a\in E.
\]
The first inequality is strict in a contrast $a$ exactly when
$m_1<m_2$ and $a^*{{\mathrm{\Delta}}} a>0$.  Thus solver replication gives a formal
variance ordering
at equal score information; it does not compare computational budgets.
\end{corollary}

\begin{corollary}\label{cor:inner-accuracy}
Suppose the centered conditions of \Cref{thm:main} hold with
$\bar {{\mathrm{P}}}={{\mathrm{P}}}_0$, where ${{\mathrm{P}}}_0$ is deterministic and invertible and ${{\mathrm{P}}}_0{{\mathrm{H}}}$ is
positive stable.  Assume \eqref{eq:S-def} and, in addition to the drift
condition \eqref{eq:mean-P-sum},
\begin{align}
 \frac1n\sum_{k=0}^{n-1}
 \E\{\norm{({{\mathrm{\Pi}}}_k-{{\mathrm{P}}}_0)\xi_{k+1}}^2\mid\F_k\}&\pto0,
 \label{eq:mean-solver-score-small}\\
 \frac1n\sum_{k=0}^{n-1}
 \E\{\norm{({{\widetilde{\mathrm{P}}}}_{k+1}-{{\mathrm{\Pi}}}_k)\xi_{k+1}}^2\mid\F_k\}&\pto0.
 \label{eq:random-solver-score-small}
\end{align}
Then ${{\mathrm{Q}}}={{\mathrm{P}}}_0{{\mathrm{S}}}{{\mathrm{P}}}_0^*$ and ${{\mathrm{V}}}=\Vsand$.  Under conditional solver score
independence and a uniformly bounded conditional second moment of the
score, the single stronger condition
\begin{equation}\label{eq:full-solver-l2-small}
 \frac1n\sum_{k=0}^{n-1}
 \E(\norm{{{\widetilde{\mathrm{P}}}}_{k+1}-{{\mathrm{P}}}_0}_{\op}^2\mid\F_k)\pto0
\end{equation}
is sufficient for
\eqref{eq:mean-solver-score-small}--\eqref{eq:random-solver-score-small}.
Thus a growing inner budget recovers the sandwich when both the conditional
mean error and the new random error vanish in the directions excited by
the score.
\end{corollary}

The root-$n$ drift condition \eqref{eq:mean-P-sum} alone is not enough for
sandwich recovery.  It controls how the conditional mean operator acts on the current state,
but it does not control its error on the target score or the random part of
the solver error on that score.  The two conditions above are therefore
separate requirements for the covariance conclusion.

\subsubsection{Predictable dynamic preconditioning}\label{sec:predictable}

Fresh solver randomness can remain in the leading score direction.
For predictable preconditioners, an alternative representation removes
the preconditioner when its accumulated inverse gain variation is small.
Let ${{\mathrm{C}}}_k:E\to E$ be invertible and $\F_k$-measurable, and consider the recursion
\begin{equation}\label{eq:predictable-recursion}
 u_{k+1}=u_k-\eta_{k+1}
 \{{{\mathrm{C}}}_k({{\mathrm{H}}}u_k+\xi_{k+1})+\varepsilon_{k+1}\},
\end{equation}
where geometric, nonlinear, and estimation terms are included in $\varepsilon_{k+1}$.

The main-text statement is \Cref{thm:predictable}.

\begin{corollary}\label{cor:stabilization}
Suppose $\sup_k\norm{{{\mathrm{M}}}_k}_{L^2(\op)}<\infty$,
$\E\norm{u_k}^2=O(k^{-\alpha})$, and
\[
 \norm{{{\mathrm{M}}}_k-{{\mathrm{M}}}_{k-1}}_{L^2(\op)}=O(k^{-\beta})
 \qquad\text{for some }\quad
 \beta>\frac{1+\alpha}{2}.
\]
Then \eqref{eq:predictable-abel-condition} holds.  In particular, an $O(k^{-1})$ one step variation is sufficient for every $\alpha\in(1/2,1)$.
\end{corollary}

The Euclidean stabilization criterion and its polynomial-rate scope are
developed by \citet{AnHuoDynamic2026}.  In a coherent tangent representation,
the same Abel cancellation applies after the geometric remainders are
controlled.  Predictability alone is insufficient: both accumulated
inverse gain variation and the weighted remainder condition must hold.
Order-one fresh jumps in a predictable operator can violate the criterion.
The distinction therefore concerns variation of the inverse preconditioner,
not just the time at which its random seed is revealed.

\begin{corollary}\label{cor:predictable-covariance}
Suppose \Cref{thm:predictable} holds with a martingale target score and
\begin{equation}\label{eq:predictable-S-limit}
 \frac1n\sum_{k=0}^{n-1}
 \E(\xi_{k+1}\ot\xi_{k+1}\mid\F_k)\pto {{\mathrm{S}}}.
\end{equation}
Let ${{\mathrm{S}}}_a=\T_{x_\star\to a}{{\mathrm{S}}}\T_{a\to x_\star}$ and suppose
$\widehat {{\mathrm{H}}}_n^a\pto {{\mathrm{H}}}_a$.  Assume that the raw anchor score in
\eqref{eq:raw-score-anchor} satisfies
\begin{equation}\label{eq:raw-score-decomposition}
 G_{k+1}^a=\T_{x_\star\to a}\xi_{k+1}+h_{k+1}^G,
 \qquad
 \frac1n\sum_{k=0}^{n-1}\norm{h_{k+1}^G}^2\pto0,
\end{equation}
and that the outer products of $\xi_{k+1}$ obey a martingale weak law.
Then the Welford covariance $\widehat {{\mathrm{S}}}_n^a$ of $G_{k+1}^a$ and the estimator
in \eqref{eq:predictable-cov-estimator-algorithm} satisfy
\begin{equation}\label{eq:predictable-cov-consistency}
 \widehat {{\mathrm{S}}}_n^a\pto {{\mathrm{S}}}_a,
 \qquad
 \widehat {{\mathrm{V}}}_n^{a,\mathrm{pred}}
 \pto {{\mathrm{H}}}_a^{-1}{{\mathrm{S}}}_a{({\mathrm{H}}_a^{*-1})}.
\end{equation}
After transport to $T_{\bar x_n}\M$, the corresponding Wald region has
asymptotic coverage $1-\gamma$ when ${{\mathrm{S}}}$ is positive definite.  No limit of
${{\mathrm{C}}}_k$ is required.
\end{corollary}

The two feasible estimators serve different regimes.  For a solver randomized at the current iteration, the implemented direction covariance estimates the covariance
of ${{\widetilde{\mathrm{P}}}}_{k+1}\xi_{k+1}$.  For a predictable preconditioner without a
fixed limit, the mean direction operator need not converge, but the Abel
identity removes it from the influence function; the raw score covariance
in \eqref{eq:predictable-cov-estimator-algorithm} therefore estimates the
correct sandwich law.

\subsubsection{Hessian averaging and fixed-budget sketch solvers}\label{sec:hessian}

\paragraph{A checkable rank one solver.}
The fixed-budget conditions can be verified for sketch and project without
requiring an increasingly accurate solve at every iteration.  For a unit
sketch $s$ and a positive definite Hessian estimate ${{\mathrm{B}}}$, use
$\mathcal K_{\mathrm{sap}}({{\mathrm{B}}},s)=(s\ot s)/\ip{s}{{{\mathrm{B}}}s}$ and average independent
replicas.  If ${{\mathrm{H}}}$, ${{\mathrm{B}}}$ and the predictable ${{\mathrm{B}}}_k$ share a spectral band
$[h_-,h_+]$, with $h_->0$, then
\[
 \norm{\mathcal K_{\mathrm{sap}}({{\mathrm{B}}},s)}_{\op}\le h_-^{-1},\qquad
 \norm{\mathcal K_{\mathrm{sap}}({{\mathrm{B}}},s)-\mathcal K_{\mathrm{sap}}({{\mathrm{H}}},s)}_{\op}
 \le h_-^{-2}\norm{{{\mathrm{B}}}-{{\mathrm{H}}}}_{\op}.
\]
A sketch law that spans the tangent space in mean makes the mean solver
positive definite, hence its product with ${{\mathrm{H}}}$ positive stable.  The second
bound transfers a Hessian rate $\norm{{{\mathrm{B}}}_k-{{\mathrm{H}}}}_{L^2(\op)}=O(k^{-\delta})$
to the solver.  The state rate then bounds the mean-drift remainder in
$L^1$ by a constant times
$n^{-1/2}\sum_{k=1}^n k^{-\delta-\alpha/2}=o(1)$ when
$\delta>(1-\alpha)/2$.  Under the local score, moment and invariant basin
conditions, \Cref{prop:rank-one-sap} completes the state and covariance
arguments.  This supplies a concrete route from a fixed computational
budget to centered Wald inference; the experiments below use its
coordinate sketch specialization.

Let $\widehat {{\mathrm{B}}}_{k+1}^a=\T_{x_k\to a}\widehat {{\mathrm{B}}}_{k+1}\T_{a\to x_k}$ and consider
\begin{equation*}
 \widehat {{\mathrm{H}}}_n^a=\lambda_n {{\mathrm{I}}}+\frac1n\sum_{k=0}^{n-1}\widehat {{\mathrm{B}}}_{k+1}^a.
\end{equation*}

\begin{proposition}\label{prop:hessian-rate}
Suppose ${{\mathrm{H}}}_a$ and the contributions $\widehat {{\mathrm{B}}}_{k+1}^a$ are self-adjoint, ${{\mathrm{H}}}_a$ is positive definite, and
\begin{align}
 \E(\widehat {{\mathrm{B}}}_{k+1}^a\mid\F_k)&={{\mathrm{H}}}_a+{{\mathrm{R}}}_{k+1}^B,
 &\norm{{{\mathrm{R}}}_{k+1}^B}_{L^2(\op)}&\le C\norm{u_k}_{L^2}^{\rho_{{\mathrm{H}}}},
 \label{eq:H-bias-local}\\
 \E\{\norm{\widehat {{\mathrm{B}}}_{k+1}^a-
 \E(\widehat {{\mathrm{B}}}_{k+1}^a\mid\F_k)}_{\mathrm F}^2\mid\F_k\}&\le C,
 &\lambda_n&=O(n^{-\delta_\lambda})
 \label{eq:H-noise}
\end{align}
for some $\rho_{{\mathrm{H}}}\in(0,1]$ and $\delta_\lambda>0$.  Under \eqref{eq:state-second},
\begin{equation}\label{eq:H-rate-result}
 \norm{\widehat {{\mathrm{H}}}_n^a-{{\mathrm{H}}}_a}_{L^2(\op)}
 =O(n^{-\delta_{{\mathrm{H}}}}),
 \qquad
 \delta_{{\mathrm{H}}}=\min\left\{\frac12,\frac{\alpha\rho_{{\mathrm{H}}}}{2},\delta_\lambda\right\}.
\end{equation}
If the spectrum of ${{\mathrm{H}}}_a$ belongs to a fixed closed interval $[h_-,h_+]$, $h_->0$, then spectral projection of the self-adjoint estimate onto that interval preserves the rate in \eqref{eq:H-rate-result}, and the inverse of the safeguarded estimate obeys the same rate.  In particular, for a locally Lipschitz Hessian ($\rho_{{\mathrm{H}}}=1$) and regularization satisfying $\delta_\lambda\ge\alpha/2$, $\delta_{{\mathrm{H}}}=\alpha/2>(1-\alpha)/2$.
\end{proposition}

\begin{corollary}\label{cor:predictable-hessian}
Let $\widetilde {{\mathrm{H}}}_k^a=\lambda_k{{\mathrm{I}}}+k^{-1}\sum_{j=1}^k\widehat {{\mathrm{B}}}_j^a$
be the self-adjoint anchor Hessian average.  Suppose its individual
contributions have uniformly bounded $L^2$ norm, $\sup_k|\lambda_k|<\infty$,
and $|\lambda_k-\lambda_{k-1}|=O(k^{-1})$.  Take $\widehat {{\mathrm{H}}}_k^a$ either
as $\widetilde {{\mathrm{H}}}_k^a$, assumed uniformly positive definite, or as its
spectral projection onto a fixed interval $[h_-,h_+]$, $h_->0$.
Use the exact predictable operator
\[
 {{\mathrm{C}}}_k=\T_{a\to x_\star}(\widehat {{\mathrm{H}}}_k^a)^{-1}\T_{x_\star\to a}
\]
in the recursion in the reference tangent space.  Then
$\norm{{{\mathrm{C}}}_k^{-1}-{{\mathrm{C}}}_{k-1}^{-1}}_{L^2(\op)}=O(k^{-1})$, so
\Cref{cor:stabilization} applies.  The sandwich benchmark follows even without invoking the convergence rate \eqref{eq:H-rate-result}, provided the state and remainder conditions hold.
\end{corollary}

\begin{proposition}\label{prop:fixed-sketch}
Let
\begin{equation}\label{eq:sketch-operator}
 {{\widetilde{\mathrm{P}}}}_{k+1}=\mathcal K_J({{\mathrm{B}}}_k,\Omega_{k+1}),
\end{equation}
where ${{\mathrm{B}}}_k$ is $\F_k$-measurable in coordinates of the reference tangent space,
$\Omega_{k+1}$ is conditionally independent of the current target score
given $\F_k$, with a fixed deterministic conditional law
$\mathcal L(\Omega_{k+1}\mid\F_k)=\mu$, and $J$ is fixed.
Here $\E_\Omega$ denotes expectation with respect to $\mu$.  Put
${{\mathrm{\Sigma}}}_k=\E(\xi_{k+1}\ot\xi_{k+1}\mid\F_k)$.  Suppose the local field,
state, bias, geometric, score increment, centered inexactness, and averaging
conditions of \Cref{thm:main} hold, including \Cref{ass:averaging} and the
direct population remainder condition \eqref{eq:field-remainder-sum}, and assume
\begin{align}
 \norm{{{\mathrm{B}}}_k-{{\mathrm{H}}}}_{L^2(\op)}&=O(k^{-\delta}),
 &\delta&>\frac{1-\alpha}{2},
 \label{eq:fixed-sketch-B-rate}\\
 \sup_k\E(\norm{\xi_{k+1}}^2\mid\F_k)&\le C,
 &\sup_{{{\mathrm{B}}}\in\mathcal N_{{\mathrm{H}}}}\E_\Omega
 \norm{\mathcal K_J({{\mathrm{B}}},\Omega)}_{\op}^{2}&\le C,
 \notag\\
 \left\{\E_\Omega\norm{\mathcal K_J({{\mathrm{B}}},\Omega)
 -\mathcal K_J({{\mathrm{H}}},\Omega)}_{\op}^{2}\right\}^{1/2}
 &\le L\norm{{{\mathrm{B}}}-{{\mathrm{H}}}}_{\op},
 \label{eq:fixed-sketch-lipschitz}\\
 \sup_k\norm{{{\mathrm{\Sigma}}}_k}_{\op}&\le C\quad\text{almost surely},
 \notag\\
 \frac1n\sum_{k=0}^{n-1}\E\norm{{{\mathrm{\Sigma}}}_k-{{\mathrm{S}}}}_{\op}&\longrightarrow0
 \label{eq:fixed-sketch-score-cov}
\end{align}
for a deterministic positive semidefinite ${{\mathrm{S}}}$ and an operator neighborhood
$\mathcal N_{{\mathrm{H}}}$ that contains ${{\mathrm{B}}}_k$ almost surely for every $k$ in the
reported phase.  Assume also
the direct product Lindeberg condition: for every $\epsilon>0$,
\begin{equation}\label{eq:fixed-sketch-lindeberg}
 \frac1n\sum_{k=0}^{n-1}
 \E\!\left[
 \norm{\mathcal K_J({{\mathrm{B}}}_k,\Omega_{k+1})\xi_{k+1}}^2
 \ind\{\norm{\mathcal K_J({{\mathrm{B}}}_k,\Omega_{k+1})\xi_{k+1}}
       >\epsilon\sqrt n\}\mid\F_k\right]\pto0.
\end{equation}
Define
\[
 {{\widetilde{\mathrm{P}}}}_J=\mathcal K_J({{\mathrm{H}}},\Omega),
 \qquad
 \bar {{\mathrm{P}}}_J=\E_\Omega({{\widetilde{\mathrm{P}}}}_J).
\]
If $\bar {{\mathrm{P}}}_J{{\mathrm{H}}}$ is positive stable, then the mean operator remainder,
conditional Lindeberg condition, and covariance limit in
\Cref{thm:main} hold with
${{\mathrm{Q}}}_J=\E_\Omega({{\widetilde{\mathrm{P}}}}_J{{\mathrm{S}}}{{\widetilde{\mathrm{P}}}}_J^*)$.
Consequently, \Cref{thm:main} applies with $\bar {{\mathrm{P}}}=\bar {{\mathrm{P}}}_J$ and
${{\mathrm{Q}}}={{\mathrm{Q}}}_J$, and \Cref{thm:inflation} gives the independent solver covariance
decomposition.  For $J=J_k\to\infty$, sandwich recovery follows from
\Cref{cor:inner-accuracy} when both solver errors applied to the target score
\eqref{eq:mean-solver-score-small}--\eqref{eq:random-solver-score-small}
vanish.
\end{proposition}

A uniform conditional $(2+\tau)$ moment of
$\mathcal K_J({{\mathrm{B}}}_k,\Omega_{k+1})\xi_{k+1}$ is a convenient sufficient
condition for \eqref{eq:fixed-sketch-lindeberg}, but is not required by the
proposition.  A fixed number of Richardson, randomized Kaczmarz,
sketch and project, or fixed coefficient accelerated iterations started
from zero produces a linear map of the right hand side and is covered by
\eqref{eq:sketch-operator}.  If coefficients or stopping times are selected
from the current residual, the map need not be linear; the discrepancy from
a reference linear operator must then be included in $e_{k+1}$ and checked
through \Cref{ass:bias-inexact,ass:cov-estimation}.

The generic proposition becomes fully checkable for the following named
instance of the sketch and project construction of
\citet{GowerRichtarik2015}.  It uses one rank one sketch per replica; increasing
$m$ averages independent replicas but does not change the one step inner
budget.

\begin{proposition}
\label{prop:rank-one-sap}
Suppose ${{\mathrm{H}}}={{\mathrm{H}}}^*\succ0$, and let ${{\mathrm{B}}}_k={{\mathrm{B}}}_k^*$ be $\F_k$-measurable.  Assume
throughout the reported phase, almost surely,
\begin{equation}\label{eq:sap-spectral-band}
 h_-{{\mathrm{I}}}\preceq {{\mathrm{H}}}\preceq h_+{{\mathrm{I}}},
 \qquad
 h_-{{\mathrm{I}}}\preceq {{\mathrm{B}}}_k\preceq h_+{{\mathrm{I}}},
 \qquad 0<h_-<h_+<\infty.
\end{equation}
For a unit vector $s$, define the one step sketch and project operator
\begin{equation}\label{eq:sap-operator}
 \mathcal K_{\mathrm{sap}}({{\mathrm{B}}},s)
 =\frac{s\ot s}{\ip{s}{{{\mathrm{B}}}s}}.
\end{equation}
At iteration $k+1$, draw a fixed number $m$ of conditionally independent
sketches $s_{k+1}^{(1)},\ldots,s_{k+1}^{(m)}$ from a time-homogeneous law
$\mu$ on the unit sphere, independently of the current observation, and use
\begin{equation*}
 {{\widetilde{\mathrm{P}}}}_{k+1}^{(m)}
 =\frac1m\sum_{j=1}^m
 \mathcal K_{\mathrm{sap}}({{\mathrm{B}}}_k,s_{k+1}^{(j)}).
\end{equation*}
Assume the sketch law spans $E$ in mean:
\begin{equation}\label{eq:sap-spanning}
 \E_\mu(s\ot s)\succeq\kappa {{\mathrm{I}}}
 \quad\text{for some }\kappa>0.
\end{equation}
Then, for self-adjoint ${{\mathrm{B}}},{{\mathrm{B}}}'$ whose spectra lie in $[h_-,h_+]$ as in
\eqref{eq:sap-spectral-band},
\begin{align}
 \norm{\mathcal K_{\mathrm{sap}}({{\mathrm{B}}},s)}_{\op}
 &\le h_-^{-1},
 &
 \norm{\mathcal K_{\mathrm{sap}}({{\mathrm{B}}},s)
       -\mathcal K_{\mathrm{sap}}({{\mathrm{B}}}',s)}_{\op}
 &\le h_-^{-2}\norm{{{\mathrm{B}}}-{{\mathrm{B}}}'}_{\op},
 \label{eq:sap-bounds}\\
 \frac{\kappa}{h_+}{{\mathrm{I}}}
 \preceq\bar {{\mathrm{P}}}_{\mathrm{sap}}
 :=\E_\mu\{\mathcal K_{\mathrm{sap}}({{\mathrm{H}}},s)\}
 &\preceq\frac1{h_-}{{\mathrm{I}}}.
 \label{eq:sap-mean-bounds}
\end{align}
In particular, $\bar {{\mathrm{P}}}_{\mathrm{sap}}{{\mathrm{H}}}$ is positive stable.

Assume further that the score decomposition in \Cref{ass:score} holds and,
for some
$\delta>(1-\alpha)/2$, $\rho\in(0,1]$, and $\tau>0$,
\begin{align}
 \norm{{{\mathrm{B}}}_k-{{\mathrm{H}}}}_{L^2(\op)}&=O(k^{-\delta}),
 \label{eq:sap-B-rate}\\
 \norm{r_G(x)}&\le C\norm{{R_{x_\star}^{-1}(x)}}^{1+\rho},
 & \alpha(1+\rho)&>1,
 \label{eq:sap-field-rate}\\
 \sup_k\E(\norm{\xi_{k+1}}^{2+\tau}\mid\F_k)&\le C,
 &
 \E(\norm{q_{k+1}}^{2+\tau}\mid\F_k)
 &\le C\norm{u_k}^{2+\tau},
 \label{eq:sap-score-moments}\\
 \frac1n\sum_{k=0}^{n-1}
 \E\norm{{{\mathrm{\Sigma}}}_k-{{\mathrm{S}}}}_{\op}&\longrightarrow0,
 &{{\mathrm{\Sigma}}}_k&=\E(\xi_{k+1}\ot\xi_{k+1}\mid\F_k).
 \label{eq:sap-score-covariance}
\end{align}
Suppose the actual clipped recursion, with no additional pre-clipping
inexactness or localization modification, remains almost surely in the
fixed sufficiently small working basin.  The retraction and frame are
$C^2$, and the fixed anchor average is used.  Let the derivative estimator be consistent, and let the
mean solver and direction covariance estimators be the one-pass averages in
\Cref{alg:main}.  Then the centered conclusions of
\Cref{thm:main,thm:cov-estimator} hold.  With
\begin{equation*}
 {{\mathrm{R}}}_{\mathrm{sap}}
 =\E_\mu\{(\mathcal K_{\mathrm{sap}}({{\mathrm{H}}},s)-\bar {{\mathrm{P}}}_{\mathrm{sap}})
 {{\mathrm{S}}}(\mathcal K_{\mathrm{sap}}({{\mathrm{H}}},s)-\bar {{\mathrm{P}}}_{\mathrm{sap}})^*\},
\end{equation*}
the limiting covariance is
\begin{equation}\label{eq:sap-covariance}
 {{\mathrm{V}}}_m=\Vsand+\frac1m{{\mathrm{H}}}^{-1}\bar {{\mathrm{P}}}_{\mathrm{sap}}^{-1}
 {{\mathrm{R}}}_{\mathrm{sap}}{(\bar {\mathrm{P}}_{\mathrm{sap}}^{*-1})}{({\mathrm{H}}^{*-1})}.
\end{equation}
For the uniform law on the unit sphere, \eqref{eq:sap-spanning} holds with
${\E_\mu(s\ot s)}={{\mathrm{I}}}/d$.
\end{proposition}

Thus the fixed-budget theory covers a named random linear solver on an
invariant localized reporting phase.  Its common stability inequality is
verified before using a Hessian convergence rate.  A direction cap alone
does not establish basin invariance; any additional localization
modification needs its own drift and implementation-error bounds.

The complete invariant model and its verification are in \Cref{ex:invariant-quadratic}.

\subsubsection{Relation to established limits}\label{sec:special-cases}

When $\M=\R^d$, $R_x(v)=x+v$, and $\T_{x\to y}={{\mathrm{I}}}$, all geometric terms
vanish and \Cref{thm:main} becomes
\[
 \sqrt n(\bar x_n-x_\star)
 =-(\bar {{\mathrm{P}}} {{\mathrm{H}}})^{-1}\frac1{\sqrt n}\sum_{k=1}^n\zeta_k
  -(\bar {{\mathrm{P}}} {{\mathrm{H}}})^{-1}b+o_p(1).
\]
For an asymptotically deterministic preconditioner this is the classical
sandwich law; for an operator randomized with the current score it gives
\eqref{eq:V-inflation}.  If instead ${{\mathrm{P}}}_{k+1}={{\mathrm{I}}}$ and $e_{k+1}=0$, then
\[
 \sqrt nR_{x_\star}^{-1}(\bar x_n)
 \dto N_E(0,{{\mathrm{H}}}^{-1}{{\mathrm{S}}}{({\mathrm{H}}^{*-1})}),
\]
which is the local averaged-RSGD covariance of
\citet{TripuraneniEtAl2018} under the corresponding centered conditions.
Thus deterministic and predictable conditioning lead to the usual averaged
stochastic approximation covariance
\citep{PolyakJuditsky1992,BoyerGodichon2023,AnHuoDynamic2026}.  Current
randomization instead gives the ordered family in
\Cref{thm:inflation,cor:loewner-order}.  This comparison is a theorem about
first order covariance.  Last iterate randomized Newton methods have a
different Lyapunov covariance and are not special cases of the present
average \citep{WangDuNa2026}.

\subsection{Non-Euclidean applications}\label{app:model-details}

This section specializes the main theorem to parameters for which the
manifold structure is intrinsic to the statistical problem.  The purpose is
twofold: to identify the derivative ${{\mathrm{H}}}$ and target score covariance ${{\mathrm{S}}}$, and
to state primitive conditions under which the abstract remainder assumptions
hold.  The solver contribution is then obtained from
\Cref{thm:inflation,cor:loewner-order} without a separate central limit
theorem for each model.

\subsubsection{Intrinsic Fr\'echet means}\label{sec:application-frechet}

Let $Y$ be manifold-valued and consider
\begin{equation*}
 F(x)=\frac12\E d^2(x,Y),
 \qquad
 G(x)=-\E\Log_x(Y).
\end{equation*}
Let $x_\star$ be an isolated Fr\'echet mean and set
\begin{equation*}
 {{\mathrm{H}}}_F=\operatorname{Hess} F(x_\star),
 \qquad
 {{\mathrm{S}}}_F=\cov\{\Log_{x_\star}(Y)\}
 =\E\{\Log_{x_\star}(Y)\ot\Log_{x_\star}(Y)\}.
\end{equation*}
The stochastic score is
$\widehat G_{k+1}(x)=-\Log_x(Y_{k+1})$.

\begin{corollary}
\label{cor:frechet}
Assume that ${{\mathrm{H}}}_F\succ0$, the recursion is localized in a neighborhood on
which the relevant logarithm maps are single-valued, and
\begin{equation}\label{eq:frechet-l2-envelope}
 \norm{\T_{x\to x_\star}\Log_x(Y)-\Log_{x_\star}(Y)}
 \le L(Y)\norm{{R_{x_\star}^{-1}(x)}},
 \qquad \E L(Y)^2<\infty.
\end{equation}
Suppose $\E\norm{\Log_{x_\star}(Y)}^2<\infty$ and all remaining conditions
of \Cref{thm:main} hold with $b=0$.  Then
\begin{equation}\label{eq:frechet-clt}
 \sqrt n\,R_{x_\star}^{-1}(\bar x_n)
 \dto N_E\!\left(0,
 (\bar {{\mathrm{P}}} {{\mathrm{H}}}_F)^{-1}{{\mathrm{Q}}}_F{((\bar {\mathrm{P}} {\mathrm{H}}_F)^{*-1})}\right),
\end{equation}
where
\begin{equation}\label{eq:frechet-Q}
 {{\mathrm{Q}}}_F=\lim_{n\to\infty}\frac1n\sum_{k=0}^{n-1}
 \cov\!\left(
 {{\widetilde{\mathrm{P}}}}_{k+1}\Log_{x_\star}(Y_{k+1})
 \mid\F_k\right).
\end{equation}
The covariance, rather than the raw conditional second moment, is required
when the current solver and score may be dependent; under exact conditional
product centering the two coincide.
An asymptotically deterministic solver gives
${{\mathrm{H}}}_F^{-1}{{\mathrm{S}}}_F{{\mathrm{H}}}_F^{-1}$.  An independent nondegenerate current solver adds
the positive semidefinite term in \eqref{eq:V-inflation}.
\end{corollary}

Condition \eqref{eq:frechet-l2-envelope} is a local $L^2$ condition on the
derivative of the logarithm map.  A compact support/domain pair uniformly
separated from the cut and conjugate loci gives a sufficient condition by
uniform boundedness of the transported logarithm derivative.  It
is not necessary.  General central limit theorems for intrinsic means can
allow support at the cut locus under more delicate density and geometric
conditions \citep{BhattacharyaPatrangenaru2005,HotzLeWood2024}; the present
online argument uses the regular local regime in which the score itself has
the expansion required by \Cref{ass:score}.

\paragraph{The unit sphere}

For $\M=\mathbb S^q$, write
\[
 r=d(x_\star,Y),
 \qquad
 e=\frac{\Log_{x_\star}(Y)}{r}\quad (r>0),
\]
and interpret the expressions below by continuity at $r=0$.

\begin{corollary}\label{cor:sphere-frechet}
Take $R=\Exp$.  Suppose $x_\star$ is the unique intrinsic mean,
$r\le r_0<\pi/2$ almost surely, and
$\E\Log_{x_\star}(Y)=0$.  Then
\begin{align}
 {{\mathrm{H}}}_{\mathbb S}
 &=\E\!\left[
 r\cot(r){{\mathrm{I}}}+\{1-r\cot(r)\}e\ot e\right],
 \label{eq:sphere-H}\\
 {{\mathrm{S}}}_{\mathbb S}
 &=\E(r^2 e\ot e),
 \label{eq:sphere-S}
\end{align}
and ${{\mathrm{H}}}_{\mathbb S}\succeq r_0\cot(r_0){{\mathrm{I}}}$.  Let ${{\mathrm{Q}}}_{\mathbb S}$ denote the
sphere instance of the conditional product-covariance limit
\eqref{eq:frechet-Q}.  Under the algorithmic
conditions of \Cref{cor:frechet},
\begin{equation*}
 \sqrt n\,\Log_{x_\star}(\bar x_n)
 \dto N\!\left(0,
 (\bar {{\mathrm{P}}} {{\mathrm{H}}}_{\mathbb S})^{-1}{{\mathrm{Q}}}_{\mathbb S}
 {((\bar {\mathrm{P}} {\mathrm{H}}_{\mathbb S})^{*-1})}\right).
\end{equation*}
In particular, an asymptotically deterministic solver has covariance
${{\mathrm{H}}}_{\mathbb S}^{-1}{{\mathrm{S}}}_{\mathbb S}{{\mathrm{H}}}_{\mathbb S}^{-1}$, while an independent
current solver obeys \eqref{eq:V-inflation} and
\eqref{eq:V-m-order}.
\end{corollary}

The support condition is a transparent sufficient condition for regularity
of the logarithm map and a uniformly nonsingular Hessian; uniqueness remains
the separate assumption in \Cref{cor:sphere-frechet}.  Related
convexity radii for Riemannian centers of mass are studied by
\citet{Afsari2011}.  It is stronger than necessary and is imposed here to
keep the online score expansion regular.

\paragraph{Affine-invariant covariance centers}

Let $\mathcal P_q$ be the manifold of $q\times q$ symmetric positive definite
matrices with affine-invariant metric
$\ip{{{\mathrm{U}}}}{{{\mathrm{V}}}}_{{\mathrm{X}}}=\tr({{\mathrm{X}}}^{-1}{{\mathrm{U}}}{{\mathrm{X}}}^{-1}{{\mathrm{V}}})$.
Its exponential and logarithm maps are
\begin{align}
 \Exp_{{\mathrm{X}}}({{\mathrm{U}}})
 &={{\mathrm{X}}}^{1/2}\exp({{\mathrm{X}}}^{-1/2}{{\mathrm{U}}}{{\mathrm{X}}}^{-1/2}){{\mathrm{X}}}^{1/2},
 \label{eq:spd-exp}\\
 \Log_{{\mathrm{X}}}({{\mathrm{Y}}})
 &={{\mathrm{X}}}^{1/2}\log({{\mathrm{X}}}^{-1/2}{{\mathrm{Y}}}{{\mathrm{X}}}^{-1/2}){{\mathrm{X}}}^{1/2}.
 \label{eq:spd-log}
\end{align}

\begin{corollary}
\label{cor:spd-frechet}
Let ${{\mathrm{X}}}_\star$ be the affine-invariant Fr\'echet mean of a random matrix
${{\mathrm{Y}}}\in\mathcal P_q$.  Assume that the support is contained in a bounded
geodesic ball, ${{\mathrm{H}}}_{\mathrm{SPD}}=\operatorname{Hess} F({{\mathrm{X}}}_\star)\succ0$, and the remaining
conditions of \Cref{cor:frechet} hold.  Let ${{\mathrm{Q}}}_{\mathrm{SPD}}$ be the SPD
instance of \eqref{eq:frechet-Q} and define
\[
 {{\mathrm{V}}}_{\mathrm{SPD}}
 = (\bar {{\mathrm{P}}} {{\mathrm{H}}}_{\mathrm{SPD}})^{-1}{{\mathrm{Q}}}_{\mathrm{SPD}}
   {((\bar {\mathrm{P}} {\mathrm{H}}_{\mathrm{SPD}})^{*-1})}.
\]
With
\[
 {{\mathrm{S}}}_{\mathrm{SPD}}
 =\E\{\Log_{{{\mathrm{X}}}_\star}({{\mathrm{Y}}})\ot\Log_{{{\mathrm{X}}}_\star}({{\mathrm{Y}}})\},
\]
the limit in \eqref{eq:frechet-clt} holds with
${{\mathrm{H}}}_F={{\mathrm{H}}}_{\mathrm{SPD}}$ and covariance ${{\mathrm{V}}}_{\mathrm{SPD}}$.  In particular, in the
asymptotically deterministic-solver regime this reduces to
${{\mathrm{H}}}_{\mathrm{SPD}}^{-1}{{\mathrm{S}}}_{\mathrm{SPD}}{{\mathrm{H}}}_{\mathrm{SPD}}^{-1}$.  For
$\psi({{\mathrm{X}}})=\log\det {{\mathrm{X}}}$,
\begin{equation}\label{eq:spd-logdet-clt}
 \sqrt n\{\log\det(\bar {{\mathrm{X}}}_n)-\log\det({{\mathrm{X}}}_\star)\}
 \dto N(0,\ell {{\mathrm{V}}}_{\mathrm{SPD}}\ell^*),
 \qquad
 \ell({{\mathrm{U}}})=\tr({{\mathrm{X}}}_\star^{-1}{{\mathrm{U}}}).
\end{equation}
\end{corollary}

The affine-invariant positive definite manifold is complete, simply
connected, and nonpositively curved, so it has no cut-locus obstruction to
\eqref{eq:spd-log}; see \citet{Bhatia2007}.  Bounded support is used only as
a convenient route to the derivative envelope and can be replaced by an
integrable envelope on the logarithm and its derivative.

\subsubsection{Streaming principal subspaces}\label{sec:application-pca}

Let $X\in\R^p$ have finite second moment, and write its uncentered second
moment as
\begin{equation}\label{eq:pca-spectrum}
 {{\mathrm{M}}}=\E(XX^*)={{\mathrm{U}}}_1{{\mathrm{\Lambda}}}_1{{\mathrm{U}}}_1^*+{{\mathrm{U}}}_2{{\mathrm{\Lambda}}}_2{{\mathrm{U}}}_2^*,
\end{equation}
where $1\le r<p$, $[{{\mathrm{U}}}_1,{{\mathrm{U}}}_2]$ is an orthogonal matrix,
${{\mathrm{U}}}_1\in\R^{p\times r}$, and ${{\mathrm{U}}}_2\in\R^{p\times(p-r)}$.
Order the eigenvalues as $\lambda_1\ge\cdots\ge\lambda_p$, set
${{\mathrm{\Lambda}}}_1=\diag(\lambda_1,\ldots,\lambda_r)$ and
${{\mathrm{\Lambda}}}_2=\diag(\lambda_{r+1},\ldots,\lambda_p)$, and assume
$\lambda_r>\lambda_{r+1}$.  Thus ${{\mathrm{U}}}_1$ spans the leading eigenspace.  The parameter is the point
$[{{\mathrm{U}}}_1]\in\operatorname{Gr}(r,p)$ minimizing
$F([{{\mathrm{U}}}])=-\frac12\E\norm{{{\mathrm{U}}}^*X}^2$.
Represent a tangent vector by ${{\mathrm{K}}}\in\R^{(p-r)\times r}$ through ${{\mathrm{U}}}_2{{\mathrm{K}}}$, and
write $z_1={{\mathrm{U}}}_1^*X\in\R^r$ and $z_2={{\mathrm{U}}}_2^*X\in\R^{p-r}$.  The manifold dimension is
$r(p-r)$.  Column-stacking $\mathrm{vec}$ identifies tangent matrices
isometrically with $\R^{r(p-r)}$; all covariance products below represent
the tangent operators in these coordinates.  We use the graph/polar retraction
$R_{[{{\mathrm{U}}}_1]}({{\mathrm{U}}}_2{{\mathrm{K}}})=[({{\mathrm{U}}}_1+{{\mathrm{U}}}_2{{\mathrm{K}}})({{\mathrm{I}}}+{{\mathrm{K}}}^*{{\mathrm{K}}})^{-1/2}]$.  For any other $C^2$
retraction with the same differential, its inverse-retraction coordinate and
the graph coordinate differ quadratically; hence their difference at the
reported average is $o_p(n^{-1/2})$.

\begin{corollary}
\label{cor:grassmann-pca}
Suppose $\E\norm X^4<\infty$, the eigengap in
\eqref{eq:pca-spectrum} is positive, and the recursion is localized and
satisfies the centered conditions of \Cref{thm:main}.  In the graph chart at
$[{{\mathrm{U}}}_1]$, the population derivative and target score are
\begin{equation*}
 \mathcal H[{{\mathrm{K}}}]={{\mathrm{K}}}{{\mathrm{\Lambda}}}_1-{{\mathrm{\Lambda}}}_2{{\mathrm{K}}},
 \qquad
 \xi=-z_2z_1^*.
\end{equation*}
If
\begin{equation*}
 {{\mathrm{Q}}}_{\mathrm{PCA}}
 =\lim_{n\to\infty}\frac1n\sum_{k=0}^{n-1}
 \cov\!\left(
 \mathrm{vec}\{{{\widetilde{\mathrm{P}}}}_{k+1}(z_{2,k+1}z_{1,k+1}^*)\}
 \mid\F_k\right),
\end{equation*}
then, for the tangent graph coordinate $\bar {{\mathrm{K}}}_n$ of the reported average,
\begin{equation}\label{eq:pca-clt}
 \sqrt n\,\mathrm{vec}(\bar {{\mathrm{K}}}_n)
 \dto N\!\left(0,
 (\bar {{\mathrm{P}}}\mathcal H)^{-1}{{\mathrm{Q}}}_{\mathrm{PCA}}
 {((\bar {\mathrm{P}}\mathcal H)^{*-1})}\right).
\end{equation}
For an asymptotically deterministic solver, the covariance is
$\mathcal H^{-1}{{\mathrm{S}}}_{\mathrm{PCA}}\mathcal H^{-1}$, where
\[
 {{\mathrm{S}}}_{\mathrm{PCA}}=\cov\{\mathrm{vec}(z_2z_1^*)\}.
\]
 In this asymptotically deterministic-solver regime, if $X$ is centered Gaussian in the eigenbasis, let $Z$ denote the limiting Gaussian tangent matrix.  Its coordinates
are uncorrelated and
\begin{equation}\label{eq:pca-coordinate-variance}
 \operatorname{Var}(Z_{ij})
 =\frac{\lambda_{r+i}\lambda_j}
        {(\lambda_j-\lambda_{r+i})^2}.
\end{equation}
For the spectral projector ${{\mathrm{\Pi}}}_\star={{\mathrm{U}}}_1{{\mathrm{U}}}_1^*$,
\begin{equation}\label{eq:pca-projector-derivative}
 D{{\mathrm{\Pi}}}_{[{{\mathrm{U}}}_1]}[{{\mathrm{K}}}]={{\mathrm{U}}}_2{{\mathrm{K}}}{{\mathrm{U}}}_1^*+{{\mathrm{U}}}_1{{\mathrm{K}}}^*{{\mathrm{U}}}_2^*.
\end{equation}
\end{corollary}

For centered $X$, ${{\mathrm{M}}}$ in \eqref{eq:pca-spectrum} is the covariance.  Without
centering, the same score and Hessian derivation targets the second moment
about the chosen origin; centering is needed only to identify ${{\mathrm{M}}}$ with the
covariance.  The eigengap is not a technical convenience: it identifies a unique leading
subspace and makes $\mathcal H$ invertible.  Without it, a regular root-$n$
limit for a fixed subspace is not available without changing the target.
The fourth moment of $X$ is a second moment condition for the quadratic score
$z_2z_1^*$ and its local random derivative; it is not a fourth moment
condition on the iterate.  The local geometry agrees with the standard
Grassmann formulation of invariant-subspace optimization
\citep{EdelmanEtAl1998,TripuraneniEtAl2018,LiJordan2023}.
Entrywise and high-dimensional uncertainty quantification for Oja-type
streaming PCA has been studied by \citet{KumarEtAl2025} and
\citet{XuLi2026}.  The result here instead concerns a fixed-dimensional
Riemannian Polyak-Ruppert average with randomized Newton directions.

\subsection{Primitive conditions for the high-level assumptions}
\label{app:primitive-conditions}

This section records three reusable ways to verify the product, remainder,
and geometric conditions in the main text.  The propositions are sufficient
routes, not additional requirements of \Cref{thm:main}.

The verification order is one-way.  Product and local direction bounds first
control the raw update; \Cref{prop:state-rate} then obtains the mean-square
state rate directly from the local update map.  The accumulated population
remainders and the two root-$n$ geometric sums are verified only afterward
through \Cref{prop:remainder-rates,prop:geometry-primitive}.  In particular,
the state rate argument assumes neither conclusion.

\begin{proposition}
\label{prop:sample-split}
Suppose, conditionally on $\F_k$, that ${{\widetilde{\mathrm{P}}}}_{k+1}$ is independent of
$(\xi_{k+1},q_{k+1})$.  If
\[
 \sup_k\E(\norm{{{\widetilde{\mathrm{P}}}}_{k+1}}_{\op}^{2}\mid\F_k)\le C,
 \qquad
 \sup_k\E(\norm{\xi_{k+1}}^2\mid\F_k)\le C,
\]
then $b_{k+1}^{\xi}=0$ and
$\E(\norm{{{\widetilde{\mathrm{P}}}}_{k+1}\xi_{k+1}}^2\mid\F_k)\le C$.  If, in addition,
$\E(\norm{q_{k+1}}^2\mid\F_k)\le C\norm{u_k}^2$ and the conditional second
moment of ${{\widetilde{\mathrm{P}}}}_{k+1}$ is bounded as above, then $b_{k+1}^{q}=0$ and
\eqref{eq:chi-moment} holds.  A sufficient tail condition for
\eqref{eq:conditional-lindeberg} is
\begin{equation}\label{eq:product-moment}
 \sup_k\E(\norm{{{\widetilde{\mathrm{P}}}}_{k+1}\xi_{k+1}}^{2+\tau}\mid\F_k)\le C
 \quad\text{for some }\tau>0.
\end{equation}
Under conditional independence, \eqref{eq:product-moment} follows from
separate $(2+\tau_{{\mathrm{P}}})$ and $(2+\tau_\xi)$ moments whenever
$\tau\le\min(\tau_{{\mathrm{P}}},\tau_\xi)$.  Without independence, conditional H\"older
still yields the product moment, but not product centering, if
\[
 \frac{2+\tau}{2+\tau_{{\mathrm{P}}}}+\frac{2+\tau}{2+\tau_\xi}\le1.
\]
\end{proposition}

Conditional independence is sufficient, not necessary.  A model-specific
identity may give $b_{k+1}^{\xi}=0$ when the score and Hessian use the same
observation.  Conversely, separate conditional unbiasedness of the solver
and score does not imply $\E({{\widetilde{\mathrm{P}}}}_{k+1}\xi_{k+1}\mid\F_k)=0$.

\begin{proposition}
\label{prop:remainder-rates}
Assume \eqref{eq:state-second} and \eqref{eq:P-moment-new}.
\begin{enumerate}[label=(\roman*),leftmargin=2em]
\item If
$\{\E\norm{{{\mathrm{\Pi}}}_k-\bar {{\mathrm{P}}}}_{\op}^2\}^{1/2}=O(k^{-\delta})$ with
$\delta>(1-\alpha)/2$, then \eqref{eq:mean-P-sum} holds.
\item If, for some $\rho\in(0,1]$,
$\norm{r_G(x)}\le L\norm{{R_{x_\star}^{-1}(x)}}^{1+\rho}$ locally and
$\alpha(1+\rho)>1$, then \eqref{eq:field-remainder-sum} holds.
\end{enumerate}
Both conclusions may instead be verified directly when neither a polynomial
rate nor a H\"older modulus is available.
\end{proposition}

\begin{proposition}
\label{prop:geometry-primitive}
Assume \Cref{ass:local-field,ass:score,ass:fresh-solver,ass:bias-inexact}
and \eqref{eq:state-second}.  Suppose $R$ and the local frame are $C^2$ on a
relatively compact neighborhood and use the safeguard
$\eta_{k+1}\norm{D_{k+1}}\le r_0$, where
$D_{k+1}={{\mathrm{P}}}_{k+1}\widehat G_{k+1}(x_k)+e_{k+1}$.  Suppose
\begin{equation}\label{eq:primitive-direction-second}
 \E(\norm{\T_{x_k\to x_\star}D_{k+1}}^2\mid\F_k)
 \le C\{1+\norm{u_k}^2+\norm{\beta_{k+1}}^2\},
 \qquad
 \beta_{k+1}=b_{k+1}^{\xi}+b_{k+1}^{q}+a_{k+1},
\end{equation}
with $\E\norm{\beta_{k+1}}^2\le C\eta_{k+1}$.  Then \Cref{ass:geometry} holds for the actual safeguarded recursion.
A uniform conditional $(2+\tau)$ moment of its raw pre-clipping direction
is sufficient for the clipping discrepancy to have negligible root-$n$
mean and vanishing average conditional variance along that recursion.
An uncapped method requires its own common-domain expansion and localization
argument; clipping-tail bounds alone do not transfer a state rate result
between different trajectories.
\end{proposition}

\subsection{Verification of the high-level conditions}
\label{app:verification-guide}

The main theorem is stated in terms of the accumulated quantities that enter
the root-$n$ expansion.  This section records common primitive routes to
those conditions without making them part of every application.

\paragraph{Population linearization and solver mean.}
Differentiation under the expectation, justified by an integrable local
derivative envelope, gives the derivative ${{\mathrm{H}}}$.  If
\[
 \norm{r_G(x)}\le C\norm{{R_{x_\star}^{-1}(x)}}^{1+\rho},
 \qquad \alpha(1+\rho)>1,
\]
then \Cref{prop:remainder-rates} gives
\eqref{eq:field-remainder-sum}.  Likewise,
$\norm{{{\mathrm{\Pi}}}_k-\bar {{\mathrm{P}}}}_{L^2}=O(k^{-\delta})$ with
$\delta>(1-\alpha)/2$ is sufficient for \eqref{eq:mean-P-sum}.  Neither
polynomial rate is necessary if the corresponding sum can be bounded
directly.

\paragraph{Scores, products, and inexactness.}
For fresh i.i.d. observations with tangent-valued estimating function
$\mathsf g(x,Z)$, $\E\mathsf g(x,Z)=G(x)$, take
\[
 \xi_{k+1}=\mathsf g(x_\star,Z_{k+1}),\qquad
 q_{k+1}=\T_{x_k\to x_\star}\mathsf g(x_k,Z_{k+1})
 -\mathsf g(x_\star,Z_{k+1})-\T_{x_k\to x_\star}G(x_k).
\]
A local $L^2$ Lipschitz envelope for the transported score gives
$\E(\norm{q_{k+1}}^2\mid\F_k)\le C\norm{u_k}^2$.
Together with a bounded solver operator, suitable joint-product bounds,
or the conditional independence and moment assumptions of
\Cref{prop:sample-split}, this verifies \eqref{eq:chi-moment}.

Without independence, one must verify the cumulative
conditional product mean in \eqref{eq:rootn-bias}.  A deterministic bias
bound $O(k^{-\kappa})$ with $\kappa>1/2$ gives $b=0$, whereas an order
$k^{-1/2}$ contribution may produce the local shift in \Cref{thm:main}.
The centered inner error is negligible when its average conditional second
moment tends to zero.

\paragraph{State and geometry.}
A local Lyapunov inequality for ${{\mathrm{\Pi}}}_k{{\mathrm{H}}}$, bounded conditional second moments
of the implemented direction, and sufficiently small geometric remainder
give $\E\norm{u_k}^2=O(\eta_k)$ by \Cref{prop:state-rate}.  A $C^2$
retraction and frame then yield \Cref{ass:geometry} after a local direction
safeguard.  For the fixed anchor average, $C^2$ chart transitions and
$\alpha>1/2$ make the quadratic averaging error root-$n$ negligible.  The recursive average uses the same $C^2$ regularity on its fixed
small coordinate ball; \eqref{eq:average-local-expansion} supplies the
quadratic error and proves invariance of that average's ball.

\paragraph{Covariance limits.}
A direct predictable covariance limit and conditional uniform integrability
verify \eqref{eq:Q-limit-new}--\eqref{eq:conditional-lindeberg}.  A bounded
conditional $(2+\tau)$ moment of ${{\widetilde{\mathrm{P}}}}_{k+1}\xi_{k+1}$ is one simple
sufficient condition.  Consistency of the one-pass covariance estimator additionally requires
consistent derivative and mean solver estimates and vanishing Ces\`aro
squared conditional bias.  \Cref{lem:covariance-closure} derives the
realized-direction and outer-product laws from these and the preceding
conditions; no additional fourth moment is required.

\paragraph{Recovery of the sandwich covariance.}
For increasing inner accuracy, convergence of the solver mean on the state
is not enough.  The two score-direction conditions
\eqref{eq:mean-solver-score-small}--\eqref{eq:random-solver-score-small}
control, respectively, the conditional-mean error and fresh random error.
For a predictable nonconvergent preconditioner, the alternative route is the
one step variation condition in \Cref{cor:stabilization}; the covariance is
then estimated from raw scores as in \Cref{cor:predictable-covariance}.

\subsection{Unified proofs}\label{app:all-proofs}
\subsubsection{Local mean contraction}\label{app:contraction-norm}
Only in \Cref{prop:rsgd-contraction} and its proof, $\norm{\cdot}_{\op}$
denotes the operator norm induced by the ${{\mathrm{P}}}^{-1}$ energy norm:
for $v\in E$ and a linear operator ${{\mathrm{L}}}:E\to E$,
\[
 \norm{v}_{{{\mathrm{P}}}^{-1}}^2=\ip{v}{{{\mathrm{P}}}^{-1}v},
 \qquad
 \norm{{{\mathrm{L}}}}_{\op}
 :=\sup_{v\ne0}\frac{\norm{{{\mathrm{L}}}v}_{{{\mathrm{P}}}^{-1}}}
 {\norm{v}_{{{\mathrm{P}}}^{-1}}}.
\]
\begin{proof}[Proof of \Cref{prop:rsgd-contraction}]
Conjugation by ${{\mathrm{P}}}^{1/2}$ gives
\[
 {{\mathrm{P}}}^{-1/2}({{\mathrm{I}}}-\eta {{\mathrm{P}}}{{\mathrm{H}}}){{\mathrm{P}}}^{1/2}
 ={{\mathrm{I}}}-\eta {{\mathrm{P}}}^{1/2}{{\mathrm{H}}}{{\mathrm{P}}}^{1/2}.
\]
The spectral theorem therefore yields
$\norm{{{\mathrm{I}}}-\eta {{\mathrm{P}}}{{\mathrm{H}}}}_{\op}
=\max_{\lambda\in[\lambda_{\min},\lambda_{\max}]}|1-\eta\lambda|$.
The minimum is attained when the errors at the two endpoints have equal
magnitude, which gives the displayed step and contraction factor.  The RSGD
and exact Newton cases follow by substitution.
\end{proof}
\subsubsection{Recorded-discrepancy bias estimation}\label{app:recorded-bias}
One sufficient route to bias estimation is available when
$b_{k+1}^{\xi}=b_{k+1}^{q}=0$ and the full discrepancy $e_{k+1}$ is recorded.
Then $\widehat b_n^a=n^{-1/2}\sum_{k=0}^{n-1}\T_{x_k\to a}e_{k+1}$
converges in probability to $b_a$: coherent transport identifies its
predictable part with the signed bias sum, while its centered part vanishes
by the martingale variance condition in \Cref{ass:bias-inexact}.
This construction requires one additional running vector; it does not
estimate an unknown solver score product bias.

\subsubsection{An invariant bounded noise construction}
\begin{example}\label{ex:invariant-quadratic}
On $\R^d$, let $G(x)={{\mathrm{H}}}x$ with ${{\mathrm{H}}}=\diag(h_1,\ldots,h_d)\succ0$ and
$x_\star=0$.  Let the observations supply centered bounded i.i.d. noise
$\xi$ with $|\xi_i|\le h_i r_i$, and take $x_0$ in the box
$|x_i|\le r_i$.  Use ${{\mathrm{B}}}_k={{\mathrm{H}}}$, sketches $s=e_i$ with probabilities
$p_i>0$, one sketch per observation, and $0<\eta_k\le1$.
When coordinate $i$ is selected, its update is
\[
 x_{k+1,i}=(1-\eta_{k+1})x_{k,i}
          +\eta_{k+1}(-\xi_{k+1,i}/h_i),
\]
while all other coordinates remain unchanged.  Convexity makes the box
invariant.  The solver is bounded with positive definite mean, its noise
has all moments, and the population and geometric remainders vanish.
The common Lyapunov inequality above proves the state rate without assuming
it.  A constant Hessian estimator and the empirical mean solver satisfy
the remaining covariance requirements.  A step cutoff large enough for
this bounded box is inactive.  This example verifies a complete primitive
route; a nonlinear manifold application must separately establish its
working basin and model-specific moment bounds.
\end{example}
\subsubsection{Reduction to the reference tangent space and proof of the main theorem}\label{app:main-proof}

Throughout the appendices, all norms on the finite-dimensional space $E$ are equivalent.  Constants denoted by $C$ may change from line to line.  All local estimates use the fixed working domains chosen before the reporting phase. A sufficiently-small-radius requirement is part of the respective localization hypothesis; containment in a larger basin does not imply containment in a smaller one.

\paragraph{A reduced stochastic approximation recursion}

Set
\begin{align}
 \beta_{k+1}&=b_{k+1}^{\xi}+b_{k+1}^{q}+a_{k+1},
 \label{eq:beta-appendix}\\
 \omega_{k+1}
 &=({{\widetilde{\mathrm{P}}}}_{k+1}-{{\mathrm{\Pi}}}_k)\{{{\mathrm{H}}}u_k+r_G(x_k)\},
 \notag\\
 m_{k+1}&=\omega_{k+1}+\chi_{k+1}+\nu_{k+1}-c_{k+1},
 \notag\\
 r_{k+1}&=({{\mathrm{\Pi}}}_k-\bar {{\mathrm{P}}}){{\mathrm{H}}}u_k+{{\mathrm{\Pi}}}_kr_G(x_k)-d_{k+1}.
 \label{eq:r-appendix}
\end{align}

\begin{lemma}\label{lem:reduced-recursion}
Under \Cref{ass:score,ass:geometry},
\begin{equation}\label{eq:reduced-recursion}
 u_{k+1}=u_k-\eta_{k+1}
 \{{{\mathrm{A}}}u_k+\zeta_{k+1}+m_{k+1}+r_{k+1}+\beta_{k+1}\},
 \qquad {{\mathrm{A}}}=\bar {{\mathrm{P}}}{{\mathrm{H}}}.
\end{equation}
Moreover, $\E(m_{k+1}\mid\F_k)=0$.
\end{lemma}

\begin{proof}
The transported solver contribution to the population part is
\begin{align*}
 {{\widetilde{\mathrm{P}}}}_{k+1}\{{{\mathrm{H}}}u_k+r_G(x_k)\}
 &={{\mathrm{\Pi}}}_k\{{{\mathrm{H}}}u_k+r_G(x_k)\}
 +({{\widetilde{\mathrm{P}}}}_{k+1}-{{\mathrm{\Pi}}}_k)\{{{\mathrm{H}}}u_k+r_G(x_k)\}\\
 &={{\mathrm{A}}}u_k+({{\mathrm{\Pi}}}_k-\bar {{\mathrm{P}}}){{\mathrm{H}}}u_k+{{\mathrm{\Pi}}}_kr_G(x_k)+\omega_{k+1}.
\end{align*}
By \eqref{eq:zeta-def}--\eqref{eq:chi-def},
\[
 {{\widetilde{\mathrm{P}}}}_{k+1}\xi_{k+1}=\zeta_{k+1}+b_{k+1}^{\xi},
 \qquad
 {{\widetilde{\mathrm{P}}}}_{k+1}q_{k+1}=\chi_{k+1}+b_{k+1}^{q}.
\]
Insert these identities and \eqref{eq:e-decomp-new} into
\eqref{eq:fixed-tangent-raw}, move the two positive geometric terms into the bracket with negative signs, and collect terms according to
\eqref{eq:beta-appendix}--\eqref{eq:r-appendix}.  This gives
\eqref{eq:reduced-recursion} exactly.

The vector ${{\mathrm{H}}}u_k+r_G(x_k)$ is $\F_k$-measurable, so the definition of
${{\mathrm{\Pi}}}_k$ gives $\E(\omega_{k+1}\mid\F_k)=0$.  The other three terms in
$m_{k+1}$ are conditionally centered by \eqref{eq:chi-def},
\eqref{eq:e-decomp-new}, and \Cref{ass:geometry}.  Hence
$\E(m_{k+1}\mid\F_k)=0$.
\end{proof}

\begin{lemma}\label{lem:secondary-martingale}
Under \Cref{ass:local-field,ass:fresh-solver,ass:bias-inexact,ass:geometry} and the state bound \eqref{eq:state-second},
\begin{equation}\label{eq:m-small}
 \frac1{\sqrt n}\sum_{k=0}^{n-1}m_{k+1}\pto0.
\end{equation}
\end{lemma}

\begin{proof}
The uniform field bound in \Cref{ass:local-field} gives $\norm{{{\mathrm{H}}}u_k+r_G(x_k)}\le C\norm{u_k}$.  Conditional Jensen's inequality and \eqref{eq:P-moment-new} give a uniform conditional second moment bound for ${{\widetilde{\mathrm{P}}}}_{k+1}-{{\mathrm{\Pi}}}_k$.  Therefore
\begin{equation*}
 \E(\norm{\omega_{k+1}}^2\mid\F_k)
 \le C\norm{u_k}^2.
\end{equation*}
By \eqref{eq:chi-moment}, the same bound holds for $\chi_{k+1}$.  Using
$\norm{z_1+\cdots+z_4}^2\le4\sum_{j=1}^4\norm{z_j}^2$, we obtain
\begin{align*}
 \frac1n\sum_{k=0}^{n-1}\E(\norm{m_{k+1}}^2\mid\F_k)
 &\le \frac Cn\sum_{k=0}^{n-1}\norm{u_k}^2
 +\frac Cn\sum_{k=0}^{n-1}\E(\norm{\nu_{k+1}}^2\mid\F_k)\\
 &\quad+\frac Cn\sum_{k=0}^{n-1}\E(\norm{c_{k+1}}^2\mid\F_k).
\end{align*}
The expectation of the first term is bounded by
$Cn^{-1}\sum_{k=1}^{n}k^{-\alpha}=O(n^{-\alpha})$ by
\eqref{eq:state-second}; hence it converges to zero in probability.  The
last two terms converge to zero in probability by
\eqref{eq:centered-inexact-small} and \eqref{eq:geometry-sums}.  Thus the
predictable quadratic variation of the finite-dimensional martingale
$n^{-1/2}\sum m_{k+1}$ converges to zero in probability.  Applying
Lenglart's inequality coordinatewise, or the martingale isometry after a
standard localization at a bounded quadratic variation, yields
\eqref{eq:m-small}.
\end{proof}

\begin{lemma}
\label{lem:remainder-small}
Under the assumptions of \Cref{thm:main},
\begin{equation}\label{eq:r-small}
 \frac1{\sqrt n}\sum_{k=0}^{n-1}r_{k+1}\pto0,
 \qquad
 \frac1{\sqrt n}\sum_{k=0}^{n-1}\beta_{k+1}\pto b.
\end{equation}
\end{lemma}

\begin{proof}
The second conclusion is \eqref{eq:rootn-bias}.  From
\eqref{eq:r-appendix},
\[
 \frac1{\sqrt n}\sum_{k=0}^{n-1}r_{k+1}
 =\frac1{\sqrt n}\sum_{k=0}^{n-1}({{\mathrm{\Pi}}}_k-\bar {{\mathrm{P}}}){{\mathrm{H}}}u_k
 +\frac1{\sqrt n}\sum_{k=0}^{n-1}{{\mathrm{\Pi}}}_kr_G(x_k)
 -\frac1{\sqrt n}\sum_{k=0}^{n-1}d_{k+1}.
\]
The first two terms converge to zero by
\eqref{eq:mean-P-sum}--\eqref{eq:field-remainder-sum}; the last converges
to zero by \eqref{eq:geometry-sums}.  This proves \eqref{eq:r-small}.
\end{proof}

\paragraph{Abel boundary and online averaging}

\begin{lemma}\label{lem:abel-boundary}
If $\eta_k=\eta_0k^{-\alpha}$ with $\alpha\in(1/2,1)$ and
\eqref{eq:state-second} holds, then
\begin{equation*}
 \frac1{\sqrt n}\sum_{k=0}^{n-1}
 \frac{u_k-u_{k+1}}{\eta_{k+1}}\pto0.
\end{equation*}
\end{lemma}

\begin{proof}
A direct summation by parts gives the exact identity
\begin{equation*}
 \sum_{k=0}^{n-1}\frac{u_k-u_{k+1}}{\eta_{k+1}}
 =\frac{u_0}{\eta_1}-\frac{u_n}{\eta_n}
 +\sum_{k=1}^{n-1}
 \left(\frac1{\eta_{k+1}}-\frac1{\eta_k}\right)u_k.
\end{equation*}
The first term divided by $\sqrt n$ converges to zero because $u_0$ is
fixed after localization.  For the terminal term,
\[
 \frac1{\sqrt n}\left\|\frac{u_n}{\eta_n}\right\|_{L^2}
 \le Cn^{-1/2}n^{\alpha}n^{-\alpha/2}
 =Cn^{(\alpha-1)/2}\longrightarrow0.
\]
By the mean-value theorem,
$\eta_{k+1}^{-1}-\eta_k^{-1}\le Ck^{\alpha-1}$.  Hence
\begin{align*}
 \frac1{\sqrt n}
 \sum_{k=1}^{n-1}
 \left(\frac1{\eta_{k+1}}-\frac1{\eta_k}\right)
 \norm{u_k}_{L^2}
 &\le \frac C{\sqrt n}\sum_{k=1}^{n-1}k^{\alpha/2-1}\\
 &=O\{n^{(\alpha-1)/2}\}=o(1).
\end{align*}
The $L^1$ bound and Markov's inequality complete the proof.
\end{proof}

For the averaging argument, put
$\widetilde u_n=n^{-1}\sum_{k=1}^nu_k$.

\begin{lemma}
\label{lem:anchor-average-equivalence}
Let $a\in U$ be fixed, suppose the chart transition
$R_a^{-1}\circ R_{x_\star}$ and its local inverse are $C^2$, and let
$\bar x_n^a$ be defined by \eqref{eq:anchor-average}.  If
$\E\norm{u_k}^2\le Ck^{-\alpha}$ with $\alpha>1/2$, then
\begin{equation}\label{eq:anchor-average-equivalence}
 \sqrt n\{R_{x_\star}^{-1}(\bar x_n^a)-\widetilde u_n\}\pto0.
\end{equation}
\end{lemma}

\begin{proof}
Define the local chart transition
\[
 f(u)=R_a^{-1}\{R_{x_\star}(u)\},
 \qquad y_\star=f(0),
 \qquad {{\mathrm{J}}}=Df(0).
\]
Because $f$ is a diffeomorphism between local coordinate neighborhoods,
${{\mathrm{J}}}$ is invertible.  A second-order Taylor expansion on a compact local set
gives
\begin{equation}\label{eq:anchor-f-expansion}
 f(u)=y_\star+{{\mathrm{J}}}u+r_f(u),
 \qquad \norm{r_f(u)}\le C\norm u^2.
\end{equation}
The recursion \eqref{eq:anchor-average} is the ordinary arithmetic mean in
$T_a\M$, hence
\begin{equation}\label{eq:anchor-y-average}
 y_n^a-y_\star
 ={{\mathrm{J}}}\widetilde u_n+\bar r_n,
 \qquad
 \bar r_n=\frac1n\sum_{k=1}^nr_f(u_k).
\end{equation}
By \eqref{eq:anchor-f-expansion} and the state rate,
\[
 \sqrt n\,\E\norm{\bar r_n}
 \le \frac C{\sqrt n}\sum_{k=1}^nk^{-\alpha}
 =O(n^{1/2-\alpha})\longrightarrow0.
\]
Thus $\sqrt n\,\bar r_n\pto0$.  Moreover, Minkowski's inequality gives
\[
 \norm{\widetilde u_n}_{L^2}
 \le\frac1n\sum_{k=1}^n\norm{u_k}_{L^2}
 =O(n^{-\alpha/2}),
\]
so \eqref{eq:anchor-y-average} implies
$y_n^a-y_\star=O_p(n^{-\alpha/2})$.

Let $g=R_{x_\star}^{-1}\circ R_a=f^{-1}$.  Then
$Dg(y_\star)={{\mathrm{J}}}^{-1}$, and another second-order Taylor expansion gives
\[
 g(y_\star+h)={{\mathrm{J}}}^{-1}h+r_g(h),
 \qquad \norm{r_g(h)}\le C\norm h^2.
\]
Using \eqref{eq:anchor-y-average},
\begin{align*}
 R_{x_\star}^{-1}(\bar x_n^a)-\widetilde u_n
 &={{\mathrm{J}}}^{-1}\bar r_n+r_g(y_n^a-y_\star).
\end{align*}
The first term is $o_p(n^{-1/2})$, while
$\sqrt n\norm{r_g(y_n^a-y_\star)}
 =O_p(n^{1/2-\alpha})=o_p(1)$.  This proves
\eqref{eq:anchor-average-equivalence}.
\end{proof}

For the recursive rule, write
$\bar u_k^{\mathrm{rec}}=R_{x_\star}^{-1}(\bar x_k^{\mathrm{rec}})$.

\begin{lemma}
\label{lem:average-equivalence}
Suppose the retraction and relevant local inverse retractions are $C^2$
on a common domain.  There is a sufficiently small coordinate ball,
chosen before the recursion, such that the following holds.  If all
$u_k$ lie in that ball and $\E\norm{u_k}^2\le Ck^{-\alpha}$ with
$1/2<\alpha<1$, then the recursive average in
\eqref{eq:riemannian-average} is well defined, remains in the ball, and
\[
 \sqrt n\{\bar u_n^{\mathrm{rec}}-\widetilde u_n\}\pto0.
\]
Moreover, $\E\norm{\bar u_k^{\mathrm{rec}}}^2=O(k^{-\alpha})$.
\end{lemma}

\begin{proof}
For $x(v)=R_{x_\star}(v)$ define
\[
 f_v(w)=R_{x_\star}^{-1}
   \bigl(R_{x(v)}[\T_{x_\star\to x(v)}w]\bigr).
\]
The inverse function theorem, uniformly on a compact neighborhood, gives
$w_{v,u}$ with $f_v(w_{v,u})=u$ and
$\norm{w_{v,u}}\le C\norm{u-v}$.  The averaging map is
$\mathcal A(v,u,t)=f_v(tw_{v,u})$.  Its values at $t=0,1$ are $v,u$,
and bounded second derivatives of $f_v$ give
\[
 \sup_{0\le t\le1}\norm{\partial_{tt}\mathcal A(v,u,t)}
 \le C\norm{u-v}^2.
\]
For a twice differentiable vector-valued function $g$ on $[0,1]$,
subtract its linear endpoint interpolant.  Integrating $g''$ twice,
with zero boundary values for the difference, gives the Green-kernel
bound $\norm{g(t)-(1-t)g(0)-tg(1)}
\le \tfrac12t(1-t)\sup_s\norm{g''(s)}$.  Therefore
\begin{equation}\label{eq:average-local-expansion}
 \mathcal A(v,u,t)=(1-t)v+tu+\varrho(v,u,t),\qquad
 \norm{\varrho(v,u,t)}\le Ct(1-t)\norm{u-v}^2.
\end{equation}
Only $C^2$ regularity has been used.

Let $\norm{u},\norm{v}\le r$, put $w=(1-t)v+tu$ and
$D=t(1-t)\norm{u-v}^2\le r^2$.  Since
\[
 \norm{w}^2=(1-t)\norm{v}^2+t\norm{u}^2-D,
\]
we have
\[
 \norm{\mathcal A(v,u,t)}^2
 \le (1-t)\norm{v}^2+t\norm{u}^2
       -\{1-2Cr-C^2r^2\}D.
\]
Choose $r$ within the common domain and with $2Cr+C^2r^2\le1$.
This proves invariance of the coordinate ball for the average, and with
$t=1/k$ gives
\[
 \E\norm{\bar u_k^{\mathrm{rec}}}^2
 \le \frac1k\sum_{j=1}^k\E\norm{u_j}^2=O(k^{-\alpha}).
\]
The recursive identity is
\[
 \bar u_k^{\mathrm{rec}}
  =\left(1-\frac1k\right)\bar u_{k-1}^{\mathrm{rec}}
    +\frac1ku_k+\varrho_k,\qquad
 \norm{\varrho_k}\le\frac Ck
      \norm{u_k-\bar u_{k-1}^{\mathrm{rec}}}^2.
\]
Subtracting the arithmetic-mean recursion and using the common
initialization at $k=1$ yields
\[
 n(\bar u_n^{\mathrm{rec}}-\widetilde u_n)
   =\sum_{k=2}^n k\varrho_k.
\]
Consequently
\[
 \E\norm{\sqrt n(\bar u_n^{\mathrm{rec}}-\widetilde u_n)}
 \le\frac C{\sqrt n}\sum_{k=2}^n
 \bigl\{\E\norm{u_k}^2+
         \E\norm{\bar u_{k-1}^{\mathrm{rec}}}^2\bigr\}
 =O(n^{1/2-\alpha})\longrightarrow0.
\]
Markov's inequality proves the claim.
\end{proof}

\paragraph{Proof of the main theorem}

\begin{proof}[Proof of Theorem~\ref{thm:main}]
Divide \eqref{eq:reduced-recursion} by $\eta_{k+1}$ and rearrange:
\[
 {{\mathrm{A}}}u_k=\frac{u_k-u_{k+1}}{\eta_{k+1}}
 -\zeta_{k+1}-m_{k+1}-r_{k+1}-\beta_{k+1}.
\]
Sum this identity from $k=0$ to $n-1$ and divide by $\sqrt n$.  By
\Cref{lem:abel-boundary,lem:secondary-martingale,lem:remainder-small},
\begin{equation}\label{eq:arithmetic-linear-representation}
 {{\mathrm{A}}}\frac1{\sqrt n}\sum_{k=0}^{n-1}u_k
 =-\frac1{\sqrt n}\sum_{k=0}^{n-1}\zeta_{k+1}-b+o_p(1).
\end{equation}
The arithmetic averages based on $u_0,\ldots,u_{n-1}$ and
$u_1,\ldots,u_n$ differ, after multiplication by $\sqrt n$, by
$(u_n-u_0)/\sqrt n=o_p(1)$.  By \eqref{eq:assumed-average-equivalence}, the latter
average is root-$n$ equivalent to $\bar u_n=R_{x_\star}^{-1}(\bar x_n)$.
Since ${{\mathrm{A}}}$ is invertible, \eqref{eq:arithmetic-linear-representation} gives
\eqref{eq:main-representation}.

The martingale central limit theorem applies to
$n^{-1/2}\sum_{k=0}^{n-1}\zeta_{k+1}$.  Its predictable covariance converges
to ${{\mathrm{Q}}}$ by \eqref{eq:Q-limit-new}, and the conditional Lindeberg condition is
assumed.  Alternatively, under \eqref{eq:product-moment}, for every
$\epsilon>0$,
\begin{align*}
 \frac1n\sum_{k=0}^{n-1}
 \E\{\norm{\zeta_{k+1}}^2
 \ind(\norm{\zeta_{k+1}}>\epsilon\sqrt n)\mid\F_k\}
 &\le \frac1{\epsilon^\tau n^{1+\tau/2}}
 \sum_{k=0}^{n-1}\E(\norm{\zeta_{k+1}}^{2+\tau}\mid\F_k)\\
 &\le Cn^{-\tau/2}\longrightarrow0.
\end{align*}
The finite-dimensional martingale CLT \citep{HallHeyde1980} therefore yields
$n^{-1/2}\sum\zeta_{k+1}\dto N_E(0,{{\mathrm{Q}}})$.  Slutsky's theorem proves
\eqref{eq:main-clt-new}.

Finally, in the retraction chart,
\[
 \psi\{R_{x_\star}(u)\}=\psi(x_\star)+D\psi_{x_\star}u+o(\norm u).
\]
The established central limit theorem implies $\bar u_n=O_p(n^{-1/2})$.
Multiplication by $\sqrt n$ and another application of Slutsky's theorem
give \eqref{eq:delta-method-new}.
\end{proof}

\begin{proof}[Proof of \Cref{prop:tail-average}]
Sum the rearranged recursion in the proof of \Cref{thm:main} from
$k=b_n$ to $n-1$ and divide by $\sqrt{N_n}$.  The two endpoint terms are
$o_p(1)$ by \Cref{lem:abel-boundary}, applied at $n$ and at $b_n$; this is
unchanged when $\tau=0$ because the initial endpoint is then
root-$N_n$ negligible.  The secondary martingale, remainder, and centered
bias sums over the tail are differences of the corresponding prefix sums,
so \Cref{lem:secondary-martingale,lem:remainder-small} and the centered
condition give
\[
 {{\mathrm{A}}}\frac1{\sqrt{N_n}}\sum_{k=b_n}^{n-1}u_k
 =-\frac1{\sqrt{N_n}}\sum_{k=b_n}^{n-1}\zeta_{k+1}+o_p(1).
\]
Prefix convergence of the predictable covariance averages implies
\[
 \frac1{N_n}\sum_{k=b_n}^{n-1}
 \E(\zeta_{k+1}\zeta_{k+1}^*\mid\F_k)\pto {{\mathrm{Q}}},
\]
and the prefix Lindeberg condition implies its tail version.  The martingale
central limit theorem therefore gives the stated Gaussian limit.  Replacing
the arithmetic tail average by the reported average uses the assumed tail
equivalence.  For the two concrete averages, restarting their recursions at
$b_n+1$ and repeating the proofs of
\Cref{lem:anchor-average-equivalence,lem:average-equivalence} with $N_n$
in place of $n$ establishes that equivalence.
\end{proof}

\begin{proof}[Proof of \Cref{cor:centered}]
Take $b=0$ in \Cref{thm:main}.
\end{proof}

\subsubsection{Product and geometric conditions}\label{app:geometry}

\paragraph{Proofs of the product and remainder criteria}

\begin{proof}[Proof of \Cref{prop:sample-split}]
Conditional independence and \eqref{eq:xi-q-center} give
\[
 b_{k+1}^{\xi}
 =\E({{\widetilde{\mathrm{P}}}}_{k+1}\mid\F_k)
  \E(\xi_{k+1}\mid\F_k)=0.
\]
Moreover,
\begin{align*}
 \E(\norm{{{\widetilde{\mathrm{P}}}}_{k+1}\xi_{k+1}}^2\mid\F_k)
 &\le
 \E(\norm{{{\widetilde{\mathrm{P}}}}_{k+1}}_{\op}^2\mid\F_k)
 \E(\norm{\xi_{k+1}}^2\mid\F_k)
 \le C.
\end{align*}
The same factorization gives $b_{k+1}^q=0$ and
\[
 \E(\norm{\chi_{k+1}}^2\mid\F_k)
 =\E(\norm{{{\widetilde{\mathrm{P}}}}_{k+1}q_{k+1}}^2\mid\F_k)
 \le C\norm{u_k}^2.
\]

If \eqref{eq:product-moment} holds, conditional centering and Jensen's
inequality give a uniform conditional $(2+\tau)$ moment of
$\zeta_{k+1}$.  Hence, for every $\epsilon>0$,
\begin{align*}
 &\frac1n\sum_{k=0}^{n-1}
 \E\!\left[\norm{\zeta_{k+1}}^2
 \ind\{\norm{\zeta_{k+1}}>\epsilon\sqrt n\}\mid\F_k\right]\\
 &\qquad\le
 \frac{C}{\epsilon^\tau n^{1+\tau/2}}
 \sum_{k=0}^{n-1}
 \E(\norm{\zeta_{k+1}}^{2+\tau}\mid\F_k)
 \le Cn^{-\tau/2}\longrightarrow0.
\end{align*}
Under conditional independence, the separate moments factorize and imply
\eqref{eq:product-moment}.  Without independence, conditional H\"older
with exponents determined by the reciprocal-sum condition gives the same
product moment.  That argument does not factor the first moment and
therefore does not establish $b_{k+1}^{\xi}=0$.
\end{proof}

\begin{proof}[Proof of \Cref{prop:remainder-rates}]
For part (i), Cauchy-Schwarz and \eqref{eq:state-second} give
\begin{align*}
 \E\norm{({{\mathrm{\Pi}}}_k-\bar {{\mathrm{P}}}){{\mathrm{H}}}u_k}
 &\le \norm {{\mathrm{H}}}_{\op}
 \{\E\norm{{{\mathrm{\Pi}}}_k-\bar {{\mathrm{P}}}}_{\op}^2\}^{1/2}
 \{\E\norm{u_k}^2\}^{1/2}\\
 &\le Ck^{-\delta-\alpha/2}.
\end{align*}
Consequently,
\[
 \E\left\{
 \frac1{\sqrt n}\sum_{k=1}^{n-1}
 \norm{({{\mathrm{\Pi}}}_k-\bar {{\mathrm{P}}}){{\mathrm{H}}}u_k}\right\}
 \le\frac C{\sqrt n}\sum_{k=1}^{n-1}
 k^{-\delta-\alpha/2}=o(1)
\]
when $\delta>(1-\alpha)/2$.  Markov's inequality proves
\eqref{eq:mean-P-sum}.

For part (ii), conditional Jensen and \eqref{eq:P-moment-new} give a
uniform bound on $\norm{{{\mathrm{\Pi}}}_k}_{\op}$.  Since $1+\rho\le2$, Lyapunov's
inequality and \eqref{eq:state-second} imply
\[
 \E\norm{u_k}^{1+\rho}
 \le\{\E\norm{u_k}^2\}^{(1+\rho)/2}
 \le Ck^{-\alpha(1+\rho)/2}.
\]
It follows that
\[
 \E\left\{
 \frac1{\sqrt n}\sum_{k=1}^{n-1}
 \norm{{{\mathrm{\Pi}}}_kr_G(x_k)}\right\}
 \le\frac C{\sqrt n}\sum_{k=1}^{n-1}
 k^{-\alpha(1+\rho)/2}=o(1)
\]
when $\alpha(1+\rho)>1$.  Markov's inequality proves
\eqref{eq:field-remainder-sum}.
\end{proof}

\paragraph{A two-variable retraction expansion}

For $u,z\in E$ and $t\ge0$, define
\begin{equation}\label{eq:Phi-definition}
 \Phi(u,z,t)=R_{x_\star}^{-1}\left[
 R_{R_{x_\star}(u)}
 \{-t\T_{x_\star\to R_{x_\star}(u)}z\}\right].
\end{equation}

\begin{lemma}\label{lem:local-update-map}
On a sufficiently small relatively compact coordinate neighborhood, there is
a $C^1$ operator map ${{\mathrm{L}}}(u):E\to E$ and a remainder
$\mathcal R_2(u,z,t)$ such that, whenever $t\norm z\le r_0$,
\begin{equation}\label{eq:Phi-expansion}
 \Phi(u,z,t)=u-tz-t{{\mathrm{L}}}(u)z+t^2\mathcal R_2(u,z,t),
\end{equation}
with
\begin{equation}\label{eq:L-R-bounds}
 {{\mathrm{L}}}(0)=0,
 \qquad \norm{{{\mathrm{L}}}(u)}_{\op}\le C\norm u,
 \qquad \norm{\mathcal R_2(u,z,t)}\le C\norm z^2.
\end{equation}
\end{lemma}

\begin{proof}
The map in \eqref{eq:Phi-definition} is $C^2$ because $R$, its local inverse,
and the frame transport are $C^2$.  It satisfies $\Phi(u,z,0)=u$.  Its
$t$-derivative at zero is linear in $z$; write $\partial_t\Phi(u,z,0)=-{{\mathrm{B}}}(u)z$.  At $u=0$, the retraction identities $R_{x_\star}(0)=x_\star$ and
$DR_{x_\star}(0)={{\mathrm{I}}}$ imply ${{\mathrm{B}}}(0)={{\mathrm{I}}}$.  Since ${{\mathrm{B}}}$ is $C^1$ on a compact local
set, ${{\mathrm{L}}}(u):={{\mathrm{B}}}(u)-{{\mathrm{I}}}$ satisfies ${{\mathrm{L}}}(0)=0$ and
$\norm{{{\mathrm{L}}}(u)}_{\op}\le C\norm u$ by the mean-value theorem.

Taylor's theorem in $t$, with integral remainder, gives
\[
 \Phi(u,z,t)=u-t{{\mathrm{B}}}(u)z
 +t^2\int_0^1(1-s)\partial_{tt}\Phi(u,z,st)\,ds.
\]
The second derivative is bilinear in $z$ up to bounded chart coefficients.
Compactness and the safeguard $t\norm z\le r_0$ therefore give
$\norm{\partial_{tt}\Phi(u,z,st)}\le C\norm z^2$.  Substituting
${{\mathrm{B}}}(u)={{\mathrm{I}}}+{{\mathrm{L}}}(u)$ proves \eqref{eq:Phi-expansion}--\eqref{eq:L-R-bounds}.
\end{proof}

\paragraph{Proof of the geometric proposition}

\begin{proof}[Proof of \Cref{prop:geometry-primitive}]
Put
\[
 Z_{k+1}=\T_{x_k\to x_\star}D_{k+1},
 \qquad
 \mu_{k+1}=\E(Z_{k+1}\mid\F_k).
\]
Coherence of the transport gives
$D_{k+1}=\T_{x_\star\to x_k}Z_{k+1}$.  Applying
\Cref{lem:local-update-map} with $t=\eta_{k+1}$ yields
\begin{equation}\label{eq:geometry-direct-expansion}
 u_{k+1}=u_k-\eta_{k+1}Z_{k+1}
 -\eta_{k+1}L(u_k)Z_{k+1}
 +\eta_{k+1}^2\mathcal R_{2,k+1}.
\end{equation}
Define
\begin{align*}
 c_{k+1}&=-L(u_k)\{Z_{k+1}-\mu_{k+1}\},
 \\
 d_{k+1}&=-L(u_k)\mu_{k+1}
 +\eta_{k+1}\mathcal R_{2,k+1}.
\end{align*}
Then \eqref{eq:geometry-direct-expansion} has exactly the form
\eqref{eq:fixed-tangent-raw}, and
$\E(c_{k+1}\mid\F_k)=0$.

We next verify the two summability statements.  By
\eqref{eq:L-R-bounds} and conditional variance domination,
\begin{align*}
 \E(\norm{c_{k+1}}^2\mid\F_k)
 &\le C\norm{u_k}^2
 \E(\norm{Z_{k+1}-\mu_{k+1}}^2\mid\F_k)\\
 &\le C\norm{u_k}^2
 \{1+\norm{u_k}^2+\norm{\beta_{k+1}}^2\}.
\end{align*}
On the localized neighborhood, $\norm{u_k}$ is uniformly small and bounded.
Using \eqref{eq:state-second} and
$\E\norm{\beta_{k+1}}^2\le C\eta_{k+1}$, we obtain
\[
 \E\left[\frac1n\sum_{k=0}^{n-1}
 \E(\norm{c_{k+1}}^2\mid\F_k)\right]
 \le \frac Cn\sum_{k=0}^{n-1}\eta_{k+1}\longrightarrow0.
\]
Thus the first statement of \eqref{eq:geometry-sums} holds by Markov's
inequality.

From the score and error decompositions,
\begin{equation}\label{eq:mu-primitive}
 \mu_{k+1}={{\mathrm{\Pi}}}_k\{{{\mathrm{H}}}u_k+r_G(x_k)\}+\beta_{k+1}.
\end{equation}
Conditional Jensen and \eqref{eq:P-moment-new} bound $\norm{{{\mathrm{\Pi}}}_k}_{\op}$
uniformly.  The uniform field bound in \Cref{ass:local-field} therefore gives
\[
 \norm{\mu_{k+1}}
 \le C\norm{u_k}+\norm{\beta_{k+1}}.
\]
Equations \eqref{eq:L-R-bounds}, \eqref{eq:primitive-direction-second}, and
Cauchy-Schwarz now imply
\begin{align*}
 \E\norm{d_{k+1}}
 &\le C\E\{\norm{u_k}\norm{\mu_{k+1}}\}
 +C\eta_{k+1}\E\norm{Z_{k+1}}^2\\
 &\le C\E\norm{u_k}^2
 +C\{\E\norm{u_k}^2\}^{1/2}
       \{\E\norm{\beta_{k+1}}^2\}^{1/2}
 +C\eta_{k+1}\\
 &\le C\eta_{k+1}.
\end{align*}
Because $\alpha>1/2$,
\[
 \E\left\|\frac1{\sqrt n}\sum_{k=0}^{n-1}d_{k+1}\right\|
 \le \frac C{\sqrt n}\sum_{k=1}^nk^{-\alpha}
 =O(n^{1/2-\alpha})\longrightarrow0.
\]
This proves the second statement of \eqref{eq:geometry-sums}.

It remains to justify clipping.  Let $Z_{k+1}^{0}$ be the raw pre-clipping direction evaluated along
the actual safeguarded trajectory, and let $Z_{k+1}^{\mathrm c}$ be that direction
clipped at radius $r_0/\eta_{k+1}$.  With
$\Delta_{k+1}=Z_{k+1}^{0}-Z_{k+1}^{\mathrm c}$ and a uniform conditional
$(2+\tau)$ moment,
\begin{align*}
 \E(\norm{\Delta_{k+1}}\mid\F_k)
 &\le \E\{\norm{Z_{k+1}^{0}}
 \ind(\eta_{k+1}\norm{Z_{k+1}^{0}}>r_0)\mid\F_k\}\\
 &\le r_0^{-(1+\tau)}\eta_{k+1}^{1+\tau}
 \E(\norm{Z_{k+1}^{0}}^{2+\tau}\mid\F_k)\\
 &\le C\eta_{k+1}^{1+\tau}.
\end{align*}
Moreover,
\[
 \E(\norm{\Delta_{k+1}}^2\mid\F_k)
 \le r_0^{-\tau}\eta_{k+1}^{\tau}
 \E(\norm{Z_{k+1}^{0}}^{2+\tau}\mid\F_k)
 \le C\eta_{k+1}^{\tau}.
\]
Hence
\[
 \E\left\{\frac1{\sqrt n}\sum_{k=0}^{n-1}
 \norm{\Delta_{k+1}}\right\}
 \le \frac C{\sqrt n}\sum_{k=1}^nk^{-\alpha(1+\tau)}=o(1),
\]
because $\alpha(1+\tau)>1/2$, while
$\frac1n\sum_k\E(\norm{\Delta_{k+1}}^2\mid\F_k)\pto0$.
Writing $\Delta_{k+1}$ as its conditional mean plus a centered part also
gives the average squared bias condition by conditional Jensen.  Thus the
clipping discrepancy satisfies all three inexactness requirements used by
the safeguarded expansion.  This proves the claim without asserting that
the safeguarded and unsafeguarded state paths are asymptotically coupled.
\end{proof}

\subsubsection{Lyapunov proof of the state rate}\label{app:state-rate}

\begin{lemma}\label{lem:quadratic-drift}
Under the assumptions of \Cref{prop:state-rate}, there exist constants
$c,C>0$ such that, for all sufficiently large $k$,
\begin{equation}\label{eq:quadratic-drift}
 \E\{\ip{u_{k+1}}{{{\mathrm{K}}}u_{k+1}}\}
 \le(1-c\eta_{k+1})\E\{\ip{u_k}{{{\mathrm{K}}}u_k}\}
 +C\eta_{k+1}^2.
\end{equation}
\end{lemma}

\begin{proof}
Let $V(u)=\ip{u}{{{\mathrm{K}}}u}$.  The local retraction expansion in
\Cref{lem:local-update-map} and the direction moment assumptions imply a
one step representation
\begin{equation}\label{eq:state-generic-increment}
 u_{k+1}=u_k-\eta_{k+1}Z_{k+1}^{\circ},
\end{equation}
where, conditionally on $\F_k$,
\begin{align}
 \E(Z_{k+1}^{\circ}\mid\F_k)
 &={{\mathrm{\Pi}}}_k{{\mathrm{H}}}u_k+\beta_{k+1}+s_{k+1},
 \notag\\
 \E(\norm{Z_{k+1}^{\circ}}^2\mid\F_k)
 &\le C\{1+\norm{u_k}^2+\norm{\beta_{k+1}}^2\},
 \label{eq:state-second-dir}
\end{align}
with a geometric/nonlinear mean remainder satisfying, for a fixed
$\varepsilon>0$ when $U$ is chosen sufficiently small before localization,
\begin{equation}\label{eq:state-mean-rem}
 \norm{s_{k+1}}
 \le \varepsilon\norm{u_k}
 +C\{\norm{u_k}^2+\eta_{k+1}
 +\norm{u_k}\norm{\beta_{k+1}}
 +\eta_{k+1}\norm{\beta_{k+1}}^2\}.
\end{equation}
To justify the second moment bound without a fourth moment, set
$Z_{k+1}=\T_{x_k\to x_\star}D_{k+1}$ and
$\eta=\eta_{k+1}$.  The local update expansion gives
\[
 Z_{k+1}^{\circ}
 =({{\mathrm{I}}}+L(u_k))Z_{k+1}-\eta\mathcal R_{2,k+1},\qquad
 \norm{\mathcal R_{2,k+1}}\le C\norm{Z_{k+1}}^2.
\]
Since the actual step satisfies $\eta\norm{Z_{k+1}}\le r_0$,
\[
 \norm{Z_{k+1}^{\circ}}
 \le (1+C\norm{u_k}+Cr_0)\norm{Z_{k+1}}.
\]
Thus its conditional second moment is bounded by a constant multiple of
that of the actual direction.  The domain and its radius are fixed before
the reporting phase; all sufficiently-small-domain requirements in this
proof refer to that same choice.

To bound the mean remainder, use \eqref{eq:mu-primitive}: the population
nonlinearity contributes ${{\mathrm{\Pi}}}_kr_G(x_k)$, the first chart correction
contributes $L(u_k)\mu_{k+1}$, and the second chart derivative contributes
$\eta_{k+1}\E(\mathcal R_{2,k+1}\mid\F_k)$.  Bounds
\eqref{eq:field-linearization}, \eqref{eq:L-R-bounds}, and
\eqref{eq:primitive-direction-second} give
\eqref{eq:state-mean-rem}; the last displayed term is the contribution of
the conditional second-order chart remainder.  Centered score, solver, inexactness, and chart
parts give \eqref{eq:state-second-dir}.

Since $V$ is quadratic, \eqref{eq:state-generic-increment} gives exactly
\begin{align}
 \E\{V(u_{k+1})\mid\F_k\}
 &=V(u_k)-2\eta_{k+1}
 \ip{{{\mathrm{K}}}u_k}{{{\mathrm{\Pi}}}_k{{\mathrm{H}}}u_k+\beta_{k+1}+s_{k+1}}
 \nonumber\\
 &\quad+\eta_{k+1}^2
 \E\{\ip{Z_{k+1}^{\circ}}{{{\mathrm{K}}}Z_{k+1}^{\circ}}\mid\F_k\}.
 \label{eq:V-exact-expand}
\end{align}
By \eqref{eq:uniform-conditional-stability},
\[
 2\ip{{{\mathrm{K}}}u_k}{{{\mathrm{\Pi}}}_k{{\mathrm{H}}}u_k}
 =\ip{u_k}{\{{{\mathrm{K}}}{{\mathrm{\Pi}}}_k{{\mathrm{H}}}+({{\mathrm{\Pi}}}_k{{\mathrm{H}}})^*{{\mathrm{K}}}\}u_k}
 \ge c_0\norm{u_k}^2.
\]
Young's inequality gives, for a sufficiently small fixed $\epsilon>0$,
\[
 2\norm {{\mathrm{K}}}_{\op}\norm{u_k}\norm{\beta_{k+1}}
 \le \epsilon\norm{u_k}^2+C_\epsilon\norm{\beta_{k+1}}^2.
\]
The terms in \eqref{eq:state-mean-rem} are handled similarly.  Choose the
coefficient $\varepsilon$ there small relative to the stable drift.  The
product $\norm{u_k}^3$ is at most a small constant times
$\norm{u_k}^2$ on this preselected sufficiently small $U$.  The term
$\norm{u_k}^2\norm{\beta_{k+1}}$ is bounded by the same Young inequality
because $\norm{u_k}$ is uniformly small.  The remaining
$\eta_{k+1}\norm{u_k}$ term is bounded by
$\epsilon\norm{u_k}^2+C\eta_{k+1}^2$.  Because the safeguarded reporting
phase keeps $u_k$ uniformly bounded, the unconditional contribution of
$\eta_{k+1}\norm{\beta_{k+1}}^2$ to the drift in
\eqref{eq:V-exact-expand} is $O(\eta_{k+1}^3)$ under
$\E\norm{\beta_{k+1}}^2\le C\eta_{k+1}$.  Finally,
\eqref{eq:state-second-dir} bounds the quadratic increment in
\eqref{eq:V-exact-expand}.

Choose $U$ and the Young constants so that all $\norm{u_k}^2$ remainders
consume at most $c_0/2$ of the stable drift.  Taking expectations, using
$\E\norm{\beta_{k+1}}^2\le C\eta_{k+1}$, and absorbing the harmless
$\eta_{k+1}^3$ terms gives
\[
 \E V(u_{k+1})\le \E V(u_k)-c\eta_{k+1}\E\norm{u_k}^2
 +C\eta_{k+1}^2.
\]
The eigenvalues of ${{\mathrm{K}}}$ are bounded above and below by positive constants, so
this is equivalent, after changing $c$ and $C$, to
\eqref{eq:quadratic-drift}.
\end{proof}

\begin{proof}[Proof of \Cref{prop:state-rate}]
Put $y_k=\E V(u_k)$.  By \Cref{lem:quadratic-drift}, $y_{k+1}\le(1-c\eta_{k+1})y_k+C\eta_{k+1}^2$.  We compare this recursion with $B\eta_k$.  Since $\alpha<1$,
\begin{equation}\label{eq:eta-difference-small}
 \frac{\eta_k-\eta_{k+1}}{\eta_{k+1}^2}
 =O(k^{\alpha-1})\longrightarrow0.
\end{equation}
Choose $B>2C/c$.  For all sufficiently large $k$, \eqref{eq:eta-difference-small}
implies
\[
 B(\eta_k-\eta_{k+1})
 \le \frac{cB-C}{2}\eta_{k+1}^2.
\]
If $y_k\le B\eta_k$, then, using $\eta_k/\eta_{k+1}\to1$,
\begin{align*}
 y_{k+1}
 &\le B\eta_k-cB\eta_{k+1}\eta_k+C\eta_{k+1}^2\\
 &\le B\eta_{k+1}
 +B(\eta_k-\eta_{k+1})-(cB-C+o(1))\eta_{k+1}^2\\
 &\le B\eta_{k+1}.
\end{align*}
Increasing $B$ handles the finite initial segment.  Thus
$y_k=O(\eta_k)$.  Equivalence of $V(u)$ and $\norm u^2$ proves
\eqref{eq:state-second}.
\end{proof}

\subsubsection{Proof of online covariance consistency}\label{app:covariance}

\begin{lemma}
\label{lem:nonnegative-compensator}
Let $X_{n,k}\ge0$ be an integrable adapted triangular array and let
${{\mathrm{A}}}_n=\sum_k\E(X_{n,k}\mid\F_{k-1})$.  If ${{\mathrm{A}}}_n\pto0$, then
$\sum_k X_{n,k}\pto0$.
\end{lemma}
\begin{proof}
Fix $a>0$ and retain term $k$ only if the predictable cumulative
compensator, including term $k$, is at most $a$.  This is an
$\F_{k-1}$-measurable decision.  Nonnegativity implies that the expected
sum of retained terms is at most $a$, while every term is retained on
$\{{{\mathrm{A}}}_n\le a\}$.  Hence, for $\epsilon>0$,
\[
 \Pr\left(\sum_kX_{n,k}>\epsilon\right)
 \le\Pr({{\mathrm{A}}}_n>a)+a/\epsilon.
\]
Let $n\to\infty$ and then $a\downarrow0$.
\end{proof}

\begin{lemma}
\label{lem:optional-covariance}
Let $(\zeta_k)$ be a square-integrable martingale difference sequence in
a fixed finite-dimensional inner-product space.  If its predictable
covariance averages converge in probability to a finite deterministic
operator ${{\mathrm{Q}}}$ and it satisfies the conditional Lindeberg condition at
scale $\sqrt n$, then
\begin{equation}\label{eq:outer-product-wlln}
 \frac1n\sum_{k=1}^n
 \left\{\zeta_k\ot\zeta_k
       -\E(\zeta_k\ot\zeta_k\mid\F_{k-1})\right\}\pto0.
\end{equation}
\end{lemma}
\begin{proof}
Fix $\epsilon>0$ and split each outer product according to
$\norm{\zeta_k}>\epsilon\sqrt n$.  The sum of conditional norms of the
normalized tail terms tends to zero by Lindeberg.  Its realized sum also
tends to zero by \Cref{lem:nonnegative-compensator}.

For the remaining terms put
\[
 {{\mathrm{B}}}_{n,k}=\frac1n\left\{\zeta_k\ot\zeta_k
       \ind\{\norm{\zeta_k}\le\epsilon\sqrt n\}
 -\E\bigl(\zeta_k\ot\zeta_k
       \ind\{\norm{\zeta_k}\le\epsilon\sqrt n\}
       \mid\F_{k-1}\bigr)\right\}.
\]
They are matrix-valued martingale differences, and
\[
 \E(\norm{{{\mathrm{B}}}_{n,k}}_{\mathrm F}^2\mid\F_{k-1})
 \le\frac{\epsilon^2}{n}
       \E(\norm{\zeta_k}^2\mid\F_{k-1}).
\]
The total predictable quantity
$T_n=n^{-1}\sum_k\E(\norm{\zeta_k}^2\mid\F_{k-1})$
is tight, since it converges to $\tr {{\mathrm{Q}}}$.  Retain a term only when the
predictable cumulative quantity defining $T_n$, including that term,
is at most $K$.  The retained martingale has total expected squared
Frobenius norm at most $\epsilon^2K$.  Thus, for $\delta>0$,
\[
 \Pr\left(\norm{\sum_k {{\mathrm{B}}}_{n,k}}_{\mathrm F}>\delta\right)
 \le\Pr(T_n>K)+\epsilon^2K/\delta^2.
\]
For fixed $K$, first take $n\to\infty$ and then
$\epsilon\downarrow0$; finally let $K\to\infty$.  Combining this bound
with the tail argument proves the weak law.
\end{proof}

\begin{lemma}
\label{lem:covariance-closure}
Under \Cref{ass:local-field,ass:score,ass:fresh-solver,ass:bias-inexact,ass:geometry,ass:averaging,ass:state-clt,ass:cov-estimation},
\begin{equation}\label{eq:realized-inexact-square}
 \frac1n\sum_{k=0}^{n-1}\norm{\nu_{k+1}}^2\pto0.
\end{equation}
The actual anchor space directions satisfy
\begin{equation}\label{eq:Y-zeta-decomposition}
 Y_{k+1}=\T_{x_\star\to a}\zeta_{k+1}+h_{k+1},\qquad
 \frac1n\sum_{k=0}^{n-1}\norm{h_{k+1}}^2\pto0,
\end{equation}
and \eqref{eq:outer-product-wlln} holds.
\end{lemma}
\begin{proof}
Apply \Cref{lem:nonnegative-compensator} with
$X_{n,k+1}=\norm{\nu_{k+1}}^2/n$ and use
\eqref{eq:centered-inexact-small} to obtain
\eqref{eq:realized-inexact-square}.  Coherence of the frame and the score
and inexactness decompositions give the exact identity
\[
 h_{k+1}=\T_{x_\star\to a}
 \left[{{\widetilde{\mathrm{P}}}}_{k+1}\{{{\mathrm{H}}}u_k+r_G(x_k)\}
       +\chi_{k+1}+\beta_{k+1}+\nu_{k+1}\right].
\]
By the uniform field bound in \Cref{ass:local-field},
\eqref{eq:P-moment-new}, and the state rate,
\[
 \E\norm{{{\widetilde{\mathrm{P}}}}_{k+1}\{{{\mathrm{H}}}u_k+r_G(x_k)\}}^2
 \le C\E\norm{u_k}^2=O(\eta_k).
\]
The same estimate holds for $\chi_{k+1}$ by
\eqref{eq:chi-moment}.  Their Ces\`aro squared norms vanish in
probability by Markov's inequality.  The squared bias condition and
\eqref{eq:realized-inexact-square}, together with the finite-sum
squared-norm inequality, prove \eqref{eq:Y-zeta-decomposition}.
Finally \Cref{lem:optional-covariance} applies by
\eqref{eq:Q-limit-new} and \eqref{eq:conditional-lindeberg}.
\end{proof}

\begin{lemma}\label{lem:direction-cov-lln}
Under the assumptions of \Cref{lem:covariance-closure},
\begin{equation*}
 \frac1n\sum_{k=0}^{n-1}Y_{k+1}\ot Y_{k+1}\pto {{\mathrm{Q}}}_a,
 \qquad
 \bar Y_n=\frac1n\sum_{k=0}^{n-1}Y_{k+1}\pto0.
\end{equation*}
\end{lemma}

\begin{proof}
Write $Z_{k+1}=\T_{x_\star\to a}\zeta_{k+1}$, so
$Y_{k+1}=Z_{k+1}+h_{k+1}$.  The weak law in \Cref{lem:optional-covariance} and
\eqref{eq:Q-limit-new} imply
\begin{equation}\label{eq:Z-cov-lln}
 \frac1n\sum_{k=0}^{n-1}Z_{k+1}\ot Z_{k+1}\pto {{\mathrm{Q}}}_a.
\end{equation}
In particular, $n^{-1}\sum\norm{Z_{k+1}}^2=O_p(1)$.  The tensor inequality
$\norm{v\ot w}_{\op}\le\norm v\norm w$ and Cauchy-Schwarz give
\begin{align*}
 &\left\|\frac1n\sum_{k=0}^{n-1}
 \{Y_{k+1}\ot Y_{k+1}-Z_{k+1}\ot Z_{k+1}\}\right\|_{\op}\\
 &\quad\le
 2\left(\frac1n\sum\norm{Z_{k+1}}^2\right)^{1/2}
  \left(\frac1n\sum\norm{h_{k+1}}^2\right)^{1/2}
 +\frac1n\sum\norm{h_{k+1}}^2=o_p(1).
\end{align*}
Together with \eqref{eq:Z-cov-lln}, this proves the first statement.

For the mean, \eqref{eq:Q-limit-new} makes the predictable quadratic
variation of $n^{-1/2}\sum Z_{k+1}$ bounded in probability.  Lenglart's
inequality therefore gives $n^{-1}\sum Z_{k+1}=O_p(n^{-1/2})$.  Also
\[
 \left\|\frac1n\sum h_{k+1}\right\|
 \le\left(\frac1n\sum\norm{h_{k+1}}^2\right)^{1/2}=o_p(1).
\]
Hence $\bar Y_n\pto0$.
\end{proof}

\begin{proof}[Proof of \Cref{thm:cov-estimator}]
Welford's algebraic identity is
\begin{equation*}
 \widehat {{\mathrm{Q}}}_n^a
 =\frac1n\sum_{k=0}^{n-1}
 (Y_{k+1}-\bar Y_n)\ot(Y_{k+1}-\bar Y_n)
 =\frac1n\sum_{k=0}^{n-1}Y_{k+1}\ot Y_{k+1}
 -\bar Y_n\ot\bar Y_n.
\end{equation*}
Thus \Cref{lem:direction-cov-lln} proves
$\widehat {{\mathrm{Q}}}_n^a\pto {{\mathrm{Q}}}_a$.  By \eqref{eq:H-P-consistency},
$\widehat {{\mathrm{A}}}_n^a=\widehat {{\mathrm{P}}}_n^a\widehat {{\mathrm{H}}}_n^a\pto\bar {{\mathrm{P}}}_a{{\mathrm{H}}}_a$.
The limit is invertible.  Matrix inversion and multiplication are continuous
on the set of invertible operators, so
\[
 \widehat {{\mathrm{V}}}_n^a=(\widehat {{\mathrm{A}}}_n^a)^{-1}\widehat {{\mathrm{Q}}}_n^a
 {((\widehat {\mathrm{A}}_n^a)^{*-1})}\pto {{\mathrm{V}}}_a.
\]
Coherence of the frame transport gives
\[
 \T_{a\to\bar x_n}{{\mathrm{V}}}_a\T_{\bar x_n\to a}
 =\T_{x_\star\to\bar x_n}{{\mathrm{V}}}\T_{\bar x_n\to x_\star}.
\]
Transport is isometric, so transporting the preceding convergence proves the
third statement in \eqref{eq:Qhat-consistency}.

We now identify the centering boundary for coverage.  Let
$\bar u_n=R_{x_\star}^{-1}(\bar x_n)$ and
$s_n=R_{\bar x_n}^{-1}(x_\star)$.  A $C^2$ expansion of the two-point inverse
retraction map at $(x_\star,x_\star)$ yields
\begin{equation}\label{eq:inverse-retraction-symmetry}
 \T_{\bar x_n\to x_\star}s_n=-\bar u_n+O_p(\norm{\bar u_n}^2).
\end{equation}
Indeed, the map vanishes on the diagonal and its two first derivatives at the
diagonal are $-{{\mathrm{I}}}$ and ${{\mathrm{I}}}$, respectively.  The main theorem gives
$\bar u_n=O_p(n^{-1/2})$, so the remainder in
\eqref{eq:inverse-retraction-symmetry} is $o_p(n^{-1/2})$.

Use the isometry of transport to write the Wald statistic at the true
parameter as
\begin{align*}
 W_n
 &=n\ip{s_n}{(\widehat {{\mathrm{V}}}_n^{\bar x})^{-1}s_n}_{\bar x_n}\\
 &=n\ip{\T_{\bar x_n\to x_\star}s_n}
 {\{\T_{\bar x_n\to x_\star}\widehat {{\mathrm{V}}}_n^{\bar x}
 \T_{x_\star\to\bar x_n}\}^{-1}
 \T_{\bar x_n\to x_\star}s_n}_{x_\star}.
\end{align*}
The covariance inside braces converges to ${{\mathrm{V}}}$.  By
\eqref{eq:inverse-retraction-symmetry} and \Cref{thm:main},
\[
 \sqrt n\T_{\bar x_n\to x_\star}s_n
 \dto N_E({{\mathrm{A}}}^{-1}b,{{\mathrm{V}}}).
\]
If ${{\mathrm{Q}}}$ is positive definite, so is ${{\mathrm{V}}}$.  Slutsky's theorem therefore gives
$W_n\dto\chi_d^2(\lambda_b)$ with \eqref{eq:uncorrected-noncentrality}.
When $b=0$, this is the central chi-squared law and proves
\eqref{eq:wald-coverage}; when $b\ne0$, it proves the stated noncentral
boundary.

It remains to verify the correction in
\eqref{eq:bias-corrected-average}.  Set
$\delta_n^{\bar x}=\T_{a\to\bar x_n}
(\widehat{\mathrm A}_n^a)^{-1}\widehat b_n^a\in T_{\bar x_n}\M$.
Coherence, consistency of
$\widehat {{\mathrm{A}}}_n^a$, and $\widehat b_n^a\pto b_a$ imply
\begin{equation*}
 \T_{\bar x_n\to x_\star}\delta_n^{\bar x}
 =\T_{a\to x_\star}(\widehat {{\mathrm{A}}}_n^a)^{-1}\widehat b_n^a
 \pto {{\mathrm{A}}}^{-1}b.
\end{equation*}
A $C^2$ expansion of the two-point retraction map gives
\[
 R_{x_\star}^{-1}(\bar x_n^{\mathrm{bc}})
 =\bar u_n+n^{-1/2}\T_{\bar x_n\to x_\star}
   \delta_n^{\bar x}+o_p(n^{-1/2}).
\]
Combining this identity with \eqref{eq:main-clt-new} yields the centered
limit asserted for $\bar x_n^{\mathrm{bc}}$.  Transport continuity also
gives consistency of $\widehat {{\mathrm{V}}}_n^{\mathrm{bc}}$.  Repeating the preceding
inverse-retraction argument at $\bar x_n^{\mathrm{bc}}$ proves coverage for
the corrected Wald region.

For a scalar functional, apply the delta method in \Cref{thm:main} to the
centered or corrected estimator, replace the limiting derivative and
covariance by their consistent transported estimators, and use the standard
normal quantile.  The resulting studentized statistic converges to $N(0,1)$.
\end{proof}

\subsubsection{Randomized solvers, predictable preconditioners, and Hessian averaging}
\label{app:solver-proofs}

\paragraph{Deterministic and randomized-solver covariance formulas}

\begin{proof}[Proof of \Cref{cor:sandwich}]
Let
$\Delta_{k+1}=({{\widetilde{\mathrm{P}}}}_{k+1}-{{\mathrm{P}}}_0)\xi_{k+1}$.  Since
$\E(\xi_{k+1}\mid\F_k)=0$,
\[
 \zeta_{k+1}={{\mathrm{P}}}_0\xi_{k+1}
 +\Delta_{k+1}-\E(\Delta_{k+1}\mid\F_k).
\]
The conditional second moment of the last two terms is bounded by
$4\E(\norm{\Delta_{k+1}}^2\mid\F_k)$.  The Ces\`aro average of this bound
converges to zero by assumption.  Conditional Cauchy-Schwarz then shows that
the cross-covariances with ${{\mathrm{P}}}_0\xi_{k+1}$ also vanish in Ces\`aro mean.
Therefore
\[
 {{\mathrm{Q}}}={{\mathrm{P}}}_0{{\mathrm{S}}}{{\mathrm{P}}}_0^*.
\]
Since ${{\mathrm{A}}}={{\mathrm{P}}}_0{{\mathrm{H}}}$,
\[
 {{\mathrm{A}}}^{-1}{{\mathrm{Q}}}{({\mathrm{A}}^{*-1})}
 ={{\mathrm{H}}}^{-1}{{\mathrm{P}}}_0^{-1}({{\mathrm{P}}}_0{{\mathrm{S}}}{{\mathrm{P}}}_0^*){({\mathrm{P}}_0^{*-1})}{({\mathrm{H}}^{*-1})}
 ={{\mathrm{H}}}^{-1}{{\mathrm{S}}}{({\mathrm{H}}^{*-1})}.
\]
\end{proof}

\begin{proof}[Proof of \Cref{thm:inflation}]
Independence and centering imply $\E({{\widetilde{\mathrm{P}}}}\xi)=0$, so the leading
innovation is ${{\widetilde{\mathrm{P}}}}\xi$.  Conditional on ${{\widetilde{\mathrm{P}}}}$, $\E\{({{\widetilde{\mathrm{P}}}}\xi)\ot({{\widetilde{\mathrm{P}}}}\xi)\mid{{\widetilde{\mathrm{P}}}}\}={{\widetilde{\mathrm{P}}}} {{\mathrm{S}}}{{\widetilde{\mathrm{P}}}}^*$.  Write ${{\widetilde{\mathrm{P}}}}=\bar {{\mathrm{P}}}+({{\widetilde{\mathrm{P}}}}-\bar {{\mathrm{P}}})$ and take expectations.  Both
cross terms vanish because $\E({{\widetilde{\mathrm{P}}}}-\bar {{\mathrm{P}}})=0$, giving
\[
 {{\mathrm{Q}}}=\bar {{\mathrm{P}}} {{\mathrm{S}}}\bar {{\mathrm{P}}}^*
 +\E\{({{\widetilde{\mathrm{P}}}}-\bar {{\mathrm{P}}}){{\mathrm{S}}}({{\widetilde{\mathrm{P}}}}-\bar {{\mathrm{P}}})^*\}
 =\bar {{\mathrm{P}}} {{\mathrm{S}}}\bar {{\mathrm{P}}}^*+{{\mathrm{R}}}.
\]
For every $v\in E$,
\[
 \ip v{{{\mathrm{R}}}v}
 =\E\norm{{{\mathrm{S}}}^{1/2}({{\widetilde{\mathrm{P}}}}-\bar {{\mathrm{P}}})^*v}^2\ge0,
\]
so ${{\mathrm{R}}}\succeq0$.  Moreover, ${{\mathrm{R}}}=0$ if and only if
${{\mathrm{S}}}^{1/2}({{\widetilde{\mathrm{P}}}}-\bar {{\mathrm{P}}})^*v=0$ almost surely for every vector in a finite
basis, which is equivalent to
$({{\widetilde{\mathrm{P}}}}-\bar {{\mathrm{P}}}){{\mathrm{S}}}^{1/2}=0$ almost surely.

Finally, ${{\mathrm{A}}}=\bar {{\mathrm{P}}}{{\mathrm{H}}}$, and hence
\begin{align*}
 {{\mathrm{V}}}
 &=(\bar {{\mathrm{P}}}{{\mathrm{H}}})^{-1}(\bar {{\mathrm{P}}}{{\mathrm{S}}}\bar {{\mathrm{P}}}^*+{{\mathrm{R}}}){((\bar {\mathrm{P}}{\mathrm{H}})^{*-1})}\\
 &={{\mathrm{H}}}^{-1}{{\mathrm{S}}}{({\mathrm{H}}^{*-1})}
 +{{\mathrm{H}}}^{-1}\bar {{\mathrm{P}}}^{-1}{{\mathrm{R}}}{(\bar {\mathrm{P}}^{*-1})}{({\mathrm{H}}^{*-1})}.
\end{align*}
Congruence preserves positive semidefiniteness, proving every claim.
\end{proof}

\begin{proof}[Proof of \Cref{cor:multiple-sketch}]
The average operator has mean $\bar {{\mathrm{P}}}$.  Conditional independence of its
replicas gives
\begin{align*}
 &\E\left[
 \{{{\widetilde{\mathrm{P}}}}_{k+1}^{(m_k)}-\bar {{\mathrm{P}}}\}{{\mathrm{S}}}
 \{{{\widetilde{\mathrm{P}}}}_{k+1}^{(m_k)}-\bar {{\mathrm{P}}}\}^*\mid\F_k\right]\\
 &\quad=\frac1{m_k^2}\sum_{j=1}^{m_k}
 \E\{({{\widetilde{\mathrm{P}}}}_{k+1}^{(j)}-\bar {{\mathrm{P}}}){{\mathrm{S}}}
 ({{\widetilde{\mathrm{P}}}}_{k+1}^{(j)}-\bar {{\mathrm{P}}})^*\mid\F_k\}
 =\frac{{{\mathrm{R}}}}{m_k};
\end{align*}
the cross terms are zero.  The predictable covariance average in the
martingale CLT is therefore
\[
 \bar {{\mathrm{P}}}{{\mathrm{S}}}\bar {{\mathrm{P}}}^*+
 \left(\frac1n\sum_{k=1}^n\frac1{m_k}\right){{\mathrm{R}}}
 \xrightarrow{p} \bar {{\mathrm{P}}}{{\mathrm{S}}}\bar {{\mathrm{P}}}^*+\lambda{{\mathrm{R}}}.
\]
Apply the same congruence calculation as in \Cref{thm:inflation}.  If
 $m_k\xrightarrow{p}\infty$, boundedness of $m_k^{-1}$ and Ces\`aro averaging
 give $\lambda=0$ in probability.
\end{proof}

\begin{proof}[Proof of \Cref{cor:loewner-order}]
For fixed $m$, \Cref{cor:multiple-sketch} gives
${{\mathrm{V}}}_m=\Vsand+m^{-1}{{\mathrm{\Delta}}}$.  If $m_1\le m_2$, then
\[
 {{\mathrm{V}}}_{m_1}-{{\mathrm{V}}}_{m_2}
 =\left(\frac1{m_1}-\frac1{m_2}\right){{\mathrm{\Delta}}}\succeq0,
 \qquad
 {{\mathrm{V}}}_{m_2}-\Vsand=\frac1{m_2}{{\mathrm{\Delta}}}\succeq0.
\]
Taking the quadratic form with $a$ gives the contrastwise identity and its
strictness criterion.
\end{proof}

\begin{proof}[Proof of \Cref{cor:inner-accuracy}]
Write
\begin{align*}
 D_{k+1}^{(1)}&=({{\mathrm{\Pi}}}_k-{{\mathrm{P}}}_0)\xi_{k+1},\\
 D_{k+1}^{(2)}&=({{\widetilde{\mathrm{P}}}}_{k+1}-{{\mathrm{\Pi}}}_k)\xi_{k+1}.
\end{align*}
Because ${{\mathrm{\Pi}}}_k$ is $\F_k$-measurable and
$\E(\xi_{k+1}\mid\F_k)=0$,
$\E(D_{k+1}^{(1)}\mid\F_k)=0$.  Therefore
\[
 \zeta_{k+1}={{\mathrm{P}}}_0\xi_{k+1}+D_{k+1}^{(1)}
 +D_{k+1}^{(2)}-\E(D_{k+1}^{(2)}\mid\F_k).
\]
The Ces\`aro conditional second moments of $D_{k+1}^{(1)}$ vanish by
\eqref{eq:mean-solver-score-small}.  Conditional variance domination and
\eqref{eq:random-solver-score-small} give the same conclusion for the
centered second term.  Conditional Cauchy-Schwarz shows that all cross
covariances involving either error have vanishing Ces\`aro norm.  Hence
\[
 \frac1n\sum_{k=0}^{n-1}
 \E(\zeta_{k+1}\ot\zeta_{k+1}\mid\F_k)
 -\frac1n\sum_{k=0}^{n-1}
 {{\mathrm{P}}}_0\E(\xi_{k+1}\ot\xi_{k+1}\mid\F_k){{\mathrm{P}}}_0^*
 \pto0.
\]
By \eqref{eq:S-def}, ${{\mathrm{Q}}}={{\mathrm{P}}}_0{{\mathrm{S}}}{{\mathrm{P}}}_0^*$.  Since ${{\mathrm{A}}}={{\mathrm{P}}}_0{{\mathrm{H}}}$, the cancellation in
\Cref{cor:sandwich} yields ${{\mathrm{V}}}=\Vsand$.

Under conditional solver score independence,
\begin{align*}
 \E\{\norm{({{\mathrm{\Pi}}}_k-{{\mathrm{P}}}_0)\xi_{k+1}}^2\mid\F_k\}
 &\le \norm{{{\mathrm{\Pi}}}_k-{{\mathrm{P}}}_0}_{\op}^2
       \E(\norm{\xi_{k+1}}^2\mid\F_k),\\
 \E\{\norm{({{\widetilde{\mathrm{P}}}}_{k+1}-{{\mathrm{\Pi}}}_k)\xi_{k+1}}^2\mid\F_k\}
 &\le
 \E(\norm{{{\widetilde{\mathrm{P}}}}_{k+1}-{{\mathrm{\Pi}}}_k}_{\op}^2\mid\F_k)
 \E(\norm{\xi_{k+1}}^2\mid\F_k).
\end{align*}
Conditional Jensen gives
$\norm{{{\mathrm{\Pi}}}_k-{{\mathrm{P}}}_0}_{\op}^2\le
\E(\norm{{{\widetilde{\mathrm{P}}}}_{k+1}-{{\mathrm{P}}}_0}_{\op}^2\mid\F_k)$, while
\[
 \E(\norm{{{\widetilde{\mathrm{P}}}}_{k+1}-{{\mathrm{\Pi}}}_k}_{\op}^2\mid\F_k)
 \le4\E(\norm{{{\widetilde{\mathrm{P}}}}_{k+1}-{{\mathrm{P}}}_0}_{\op}^2\mid\F_k).
\]
Thus \eqref{eq:full-solver-l2-small} and the bounded score second moment
imply both conditions on solver errors applied to the target score.
\end{proof}

\paragraph{The predictable Abel identity}

\begin{proof}[Proof of \Cref{thm:predictable}]
Multiply \eqref{eq:predictable-recursion} by
${{\mathrm{M}}}_k={{\mathrm{C}}}_k^{-1}$ and divide by $\eta_{k+1}$:
\begin{equation}\label{eq:predictable-isolation}
 {{\mathrm{H}}}u_k={{\mathrm{\Gamma}}}_k(u_k-u_{k+1})-\xi_{k+1}-{{\mathrm{M}}}_k\varepsilon_{k+1},
 \qquad {{\mathrm{\Gamma}}}_k=\frac{{{\mathrm{M}}}_k}{\eta_{k+1}}.
\end{equation}
Summing the first term on the right gives the exact vector Abel identity
\begin{equation*}
 \sum_{k=0}^{n-1}{{\mathrm{\Gamma}}}_k(u_k-u_{k+1})
 ={{\mathrm{\Gamma}}}_0u_0-{{\mathrm{\Gamma}}}_{n-1}u_n+
 \sum_{k=1}^{n-1}({{\mathrm{\Gamma}}}_k-{{\mathrm{\Gamma}}}_{k-1})u_k.
\end{equation*}
By \eqref{eq:predictable-abel-condition}, the right hand side divided by
$\sqrt n$ is $o_p(1)$.  Sum \eqref{eq:predictable-isolation}, divide by
$\sqrt n$, and use \eqref{eq:predictable-rem-condition} to obtain
\[
 {{\mathrm{H}}}\frac1{\sqrt n}\sum_{k=0}^{n-1}u_k
 =-\frac1{\sqrt n}\sum_{k=0}^{n-1}\xi_{k+1}+o_p(1).
\]
The assumed root-$n$ equivalence of the recursive and arithmetic averages,
and the negligible endpoint change from $u_0,\ldots,u_{n-1}$ to
$u_1,\ldots,u_n$, give \eqref{eq:predictable-clt}.  The stated Gaussian
limit follows from the score CLT and Slutsky's theorem.
\end{proof}

\begin{proof}[Proof of \Cref{cor:stabilization}]
The endpoint terms are negligible.  Indeed,
\begin{align*}
 \frac1{\sqrt n}\E\norm{{{\mathrm{\Gamma}}}_{n-1}u_n}
 &\le \frac1{\sqrt n\eta_n}
 \norm{{{\mathrm{M}}}_{n-1}}_{L^2(\op)}\norm{u_n}_{L^2}\\
 &\le Cn^{(\alpha-1)/2}\longrightarrow0,
\end{align*}
and ${{\mathrm{\Gamma}}}_0u_0/\sqrt n\to0$.

Decompose the one step difference exactly as
\begin{equation}\label{eq:inverse-gain-difference}
 {{\mathrm{\Gamma}}}_k-{{\mathrm{\Gamma}}}_{k-1}
 =\frac{{{\mathrm{M}}}_k-{{\mathrm{M}}}_{k-1}}{\eta_{k+1}}
 +{{\mathrm{M}}}_{k-1}\left(\frac1{\eta_{k+1}}-\frac1{\eta_k}\right).
\end{equation}
For the first part, Cauchy-Schwarz and the stated rates yield
\begin{align*}
 \frac1{\sqrt n}\sum_{k=1}^{n-1}
 \E\left\|
 \frac{({{\mathrm{M}}}_k-{{\mathrm{M}}}_{k-1})u_k}{\eta_{k+1}}\right\|
 &\le \frac C{\sqrt n}\sum_{k=1}^{n-1}
 k^{\alpha}k^{-\beta}k^{-\alpha/2}\\
 &\le \begin{cases}
 O\{n^{(1+\alpha)/2-\beta}\},&\beta<1+\alpha/2,\\
 O(n^{-1/2}\log n),&\beta=1+\alpha/2,\\
 O(n^{-1/2}),&\beta>1+\alpha/2,
 \end{cases}
 =o(1).
\end{align*}
For the second part, the mean-value theorem gives
$\eta_{k+1}^{-1}-\eta_k^{-1}=O(k^{\alpha-1})$, so
\begin{align*}
 \frac1{\sqrt n}\sum_{k=1}^{n-1}
 \E\left\|{{\mathrm{M}}}_{k-1}
 \left(\frac1{\eta_{k+1}}-\frac1{\eta_k}\right)u_k\right\|
 &\le \frac C{\sqrt n}\sum_{k=1}^{n-1}k^{\alpha/2-1}\\
 &=O\{n^{(\alpha-1)/2}\}=o(1).
\end{align*}
Markov's inequality and \eqref{eq:inverse-gain-difference} prove
\eqref{eq:predictable-abel-condition}.
\end{proof}

\begin{proof}[Proof of \Cref{cor:predictable-covariance}]
Put $Z_{k+1}^a=\T_{x_\star\to a}\xi_{k+1}$.  The assumed martingale weak
law and \eqref{eq:predictable-S-limit} give
\[
 \frac1n\sum_{k=0}^{n-1}Z_{k+1}^a\ot Z_{k+1}^a\pto {{\mathrm{S}}}_a,
 \qquad
 \frac1n\sum_{k=0}^{n-1}Z_{k+1}^a\pto0.
\]
By \eqref{eq:raw-score-decomposition} and Cauchy-Schwarz,
\begin{align*}
 &\left\|\frac1n\sum_{k=0}^{n-1}
 G_{k+1}^a\ot G_{k+1}^a
 -\frac1n\sum_{k=0}^{n-1}Z_{k+1}^a\ot Z_{k+1}^a\right\|_{\op}\\
 &\quad\le
 2\left(\frac1n\sum\norm{Z_{k+1}^a}^2\right)^{1/2}
  \left(\frac1n\sum\norm{h_{k+1}^G}^2\right)^{1/2}
 +\frac1n\sum\norm{h_{k+1}^G}^2=o_p(1).
\end{align*}
The sample mean of $G_{k+1}^a$ is also $o_p(1)$.  Welford's identity
therefore gives $\widehat {{\mathrm{S}}}_n^a\pto {{\mathrm{S}}}_a$.  Together with
$\widehat {{\mathrm{H}}}_n^a\pto {{\mathrm{H}}}_a$ and continuity of matrix inversion, this yields
\eqref{eq:predictable-cov-consistency}.

The influence representation in \Cref{thm:predictable} has covariance
${{\mathrm{H}}}_a^{-1}{{\mathrm{S}}}_a{({\mathrm{H}}_a^{*-1})}$ after transport to the anchor.  The same two-point
inverse-retraction expansion used in the proof of \Cref{thm:cov-estimator}
then gives convergence of the Wald statistic to $\chi_d^2$.  No step uses a
limit of ${{\mathrm{C}}}_k$.
\end{proof}

\paragraph{Streaming Hessian estimates and fixed-budget sketches}

\begin{proof}[Proof of \Cref{prop:hessian-rate}]
Let
\[
 \widetilde {{\mathrm{H}}}_n^a=\lambda_n {{\mathrm{I}}}+\frac1n\sum_{k=0}^{n-1}\widehat {{\mathrm{B}}}_{k+1}^a
\]
be the unsafeguarded estimate.  Decompose
\begin{align}
 \widetilde {{\mathrm{H}}}_n^a-{{\mathrm{H}}}_a
 &=\lambda_n{{\mathrm{I}}}+\frac1n\sum_{k=0}^{n-1}{{\mathrm{R}}}_{k+1}^B
 +\frac1n\sum_{k=0}^{n-1}{{\mathrm{M}}}_{k+1}^B,
 \label{eq:H-decomposition-proof}\\
 {{\mathrm{M}}}_{k+1}^B
 &=\widehat {{\mathrm{B}}}_{k+1}^a-
 \E(\widehat {{\mathrm{B}}}_{k+1}^a\mid\F_k).\notag
\end{align}
The final sum is a matrix-valued martingale.  In fixed dimension, the
Frobenius martingale isometry and \eqref{eq:H-noise} give
\[
 \left\|\frac1n\sum_{k=0}^{n-1}{{\mathrm{M}}}_{k+1}^B\right\|_{L^2(\mathrm F)}^2
 =\frac1{n^2}\sum_{k=0}^{n-1}\E\norm{{{\mathrm{M}}}_{k+1}^B}_{\mathrm F}^2
 \le\frac Cn.
\]
For the bias term, Minkowski's inequality, \eqref{eq:H-bias-local}, and the
state rate imply
\begin{align*}
 \left\|\frac1n\sum_{k=0}^{n-1}{{\mathrm{R}}}_{k+1}^B\right\|_{L^2(\op)}
 &\le\frac Cn\sum_{k=1}^n
 \norm{u_{k-1}}_{L^2}^{\rho_{{\mathrm{H}}}}\\
 &\le\frac Cn\sum_{k=1}^nk^{-\alpha\rho_{{\mathrm{H}}}/2}
 =O(n^{-\alpha\rho_{{\mathrm{H}}}/2}).
\end{align*}
The regularization term is $O(n^{-\delta_\lambda})$.  Combining the three
bounds in \eqref{eq:H-decomposition-proof}, and using
$\norm {{\mathrm{B}}}_{\op}\le\norm {{\mathrm{B}}}_{\mathrm F}$, proves the stated rate for
$\widetilde {{\mathrm{H}}}_n^a$.  Without safeguarding, set
$\widehat {{\mathrm{H}}}_n^a=\widetilde {{\mathrm{H}}}_n^a$.  With safeguarding, let
$\widehat {{\mathrm{H}}}_n^a$ be its spectral projection onto the closed convex set of
self-adjoint operators with spectrum in $[h_-,h_+]$.  Since the spectrum of ${{\mathrm{H}}}_a$ belongs to this interval, ${{\mathrm{H}}}_a$ is fixed by the projection.  Nonexpansiveness of metric projection in Frobenius norm gives
$\norm{\widehat {{\mathrm{H}}}_n^a-{{\mathrm{H}}}_a}_{L^2(\op)}\le\norm{\widetilde {{\mathrm{H}}}_n^a-{{\mathrm{H}}}_a}_{L^2(\mathrm F)}$, so the safeguarded estimate has the same rate (in fixed dimension).  Also
$\norm{(\widehat {{\mathrm{H}}}_n^a)^{-1}}_{\op}\le h_-^{-1}$ and
$\norm{{{\mathrm{H}}}_a^{-1}}_{\op}<\infty$.  The resolvent identity
\[
 (\widehat {{\mathrm{H}}}_n^a)^{-1}-{{\mathrm{H}}}_a^{-1}
 =(\widehat {{\mathrm{H}}}_n^a)^{-1}({{\mathrm{H}}}_a-\widehat {{\mathrm{H}}}_n^a){{\mathrm{H}}}_a^{-1}
\]
then shows that the inverse has the same rate.
\end{proof}

\begin{proof}[Proof of \Cref{cor:predictable-hessian}]
Write
$\overline {{\mathrm{B}}}_k^a=k^{-1}\sum_{j=1}^k\widehat {{\mathrm{B}}}_j^a$, so
$\widetilde {{\mathrm{H}}}_k^a=\lambda_k{{\mathrm{I}}}+\overline {{\mathrm{B}}}_k^a$.  The exact recursion
\[
 \overline {{\mathrm{B}}}_k^a-\overline {{\mathrm{B}}}_{k-1}^a
 =\frac1k\{\widehat {{\mathrm{B}}}_k^a-\overline {{\mathrm{B}}}_{k-1}^a\}
\]
and the bounded $L^2$ norms of the individual contributions imply
$\norm{\overline {{\mathrm{B}}}_k^a-\overline {{\mathrm{B}}}_{k-1}^a}_{L^2(\op)}=O(k^{-1})$.
The regularization increment has the same order by assumption.  In fixed
dimension this also gives an $O(k^{-1})$ increment in $L^2(\mathrm F)$.
For the safeguarded branch, nonexpansiveness of spectral projection in
Frobenius norm yields
$\norm{\widehat {{\mathrm{H}}}_k^a-\widehat {{\mathrm{H}}}_{k-1}^a}_{L^2(\op)}
\le\norm{\widetilde {{\mathrm{H}}}_k^a-\widetilde {{\mathrm{H}}}_{k-1}^a}_{L^2(\mathrm F)}=O(k^{-1})$.
Coherent transport is isometric and independent of $k$ between the fixed spaces
$T_a\M$ and $E$, so
\[
 \norm{{{\mathrm{C}}}_k^{-1}-{{\mathrm{C}}}_{k-1}^{-1}}_{L^2(\op)}
 =\norm{\widehat {{\mathrm{H}}}_k^a-\widehat {{\mathrm{H}}}_{k-1}^a}_{L^2(\op)}
 =O(k^{-1}).
\]
Minkowski's inequality and the bounded contribution and ridge norms give
the uniform $L^2$ bound on ${{\mathrm{C}}}_k^{-1}$ in the raw branch; spectral clipping
gives that bound directly in the safeguarded branch.  Uniform positive definiteness
ensures that ${{\mathrm{C}}}_k$ exists.  Apply \Cref{cor:stabilization}.
\end{proof}

\begin{proof}[Proof of \Cref{prop:fixed-sketch}]
Write
\[
 {{\mathrm{K}}}_k(\Omega)=\mathcal K_J({{\mathrm{B}}}_k,\Omega),
 \qquad
 {{\mathrm{K}}}_\star(\Omega)=\mathcal K_J({{\mathrm{H}}},\Omega).
\]
By the fixed conditional seed law, the conditional mean operator is
${{\mathrm{\Pi}}}_k=\E_\Omega {{\mathrm{K}}}_k(\Omega)$.  Minkowski's inequality and
\eqref{eq:fixed-sketch-lipschitz} give
\[
 \norm{{{\mathrm{\Pi}}}_k-\bar {{\mathrm{P}}}_J}_{L^2(\op)}
 \le L\norm{{{\mathrm{B}}}_k-{{\mathrm{H}}}}_{L^2(\op)}=O(k^{-\delta}).
\]
Thus \Cref{prop:remainder-rates}(i) verifies
\eqref{eq:mean-P-sum}.  The second moment assumptions and conditional
independence of seed and score give \eqref{eq:P-moment-new},
$b_{k+1}^{\xi}=0$, and a bounded conditional second moment of the product
by \Cref{prop:sample-split}.  The required tail condition is imposed
directly in \eqref{eq:fixed-sketch-lindeberg}.  Conditional independence and
$\E(\xi_{k+1}\mid\F_k)=0$ give
$\E\{{{\mathrm{K}}}_k(\Omega_{k+1})\xi_{k+1}\mid\F_k\}=0$, so the product itself is
$\zeta_{k+1}$ and \eqref{eq:conditional-lindeberg} follows without an
additional centering argument.

It remains to identify the covariance.  Conditional independence and
centering give
\begin{equation*}
 {{\mathrm{Q}}}_k:=\E(\zeta_{k+1}\ot\zeta_{k+1}\mid\F_k)
 =\E_\Omega\{{{\mathrm{K}}}_k(\Omega){{\mathrm{\Sigma}}}_k{{\mathrm{K}}}_k(\Omega)^*\}.
\end{equation*}
Add and subtract
$\E_\Omega\{{{\mathrm{K}}}_\star(\Omega){{\mathrm{\Sigma}}}_k{{\mathrm{K}}}_\star(\Omega)^*\}$.  The operator
inequality
$\norm{{{\mathrm{U}}}{{\mathrm{V}}}{{\mathrm{W}}}^*}_{\op}\le\norm {{\mathrm{U}}}_{\op}\allowbreak\norm {{\mathrm{V}}}_{\op}\allowbreak\norm {{\mathrm{W}}}_{\op}$ and
Cauchy-Schwarz in $\Omega$ yield
\begin{align*}
 &\norm{{{\mathrm{Q}}}_k-
 \E_\Omega\{{{\mathrm{K}}}_\star(\Omega){{\mathrm{\Sigma}}}_k{{\mathrm{K}}}_\star(\Omega)^*\}}_{\op}
 \nonumber\\
 &\quad\le C
 \left\{\E_\Omega\norm{{{\mathrm{K}}}_k(\Omega)-{{\mathrm{K}}}_\star(\Omega)}_{\op}^2
 \right\}^{1/2}
 \left[
 \{\E_\Omega\norm{{{\mathrm{K}}}_k(\Omega)}_{\op}^2\}^{1/2}
 +\{\E_\Omega\norm{{{\mathrm{K}}}_\star(\Omega)}_{\op}^2\}^{1/2}
 \right]
 \nonumber\\
 &\quad\le C\norm{{{\mathrm{B}}}_k-{{\mathrm{H}}}}_{\op}.
\end{align*}
Also,
\begin{equation*}
 \norm{\E_\Omega\{{{\mathrm{K}}}_\star(\Omega)
 ({{\mathrm{\Sigma}}}_k-{{\mathrm{S}}}){{\mathrm{K}}}_\star(\Omega)^*\}}_{\op}
 \le C\norm{{{\mathrm{\Sigma}}}_k-{{\mathrm{S}}}}_{\op}.
\end{equation*}
Taking Ces\`aro averages, using the $L^2$ rate in
\eqref{eq:fixed-sketch-B-rate} and
\eqref{eq:fixed-sketch-score-cov}, gives
$n^{-1}\sum_{k<n}{{\mathrm{Q}}}_k\pto {{\mathrm{Q}}}_J$.  Thus
\eqref{eq:Q-limit-new} holds.  Positive stability of $\bar {{\mathrm{P}}}_J{{\mathrm{H}}}$ and the
remaining assumptions complete the application of \Cref{thm:main}.
Because ${{\mathrm{K}}}_\star(\Omega)$ is independent of a target score with covariance
${{\mathrm{S}}}$, \Cref{thm:inflation} gives the covariance decomposition.  The
growing-$J_k$ statement follows from \Cref{cor:inner-accuracy} under its two
conditions on solver errors applied to the target score.
\end{proof}

\begin{proof}[Proof of \Cref{prop:rank-one-sap}]
For a fixed positive definite ${{\mathrm{B}}}$ and right hand side $g$, one
sketch and project step from zero is the solution of
\begin{equation*}
 \min_v\ \frac12\ip{v}{{{\mathrm{B}}}v}
 \quad\text{subject to}\quad
 \ip{s}{{{\mathrm{B}}}v}=\ip{s}{g}.
\end{equation*}
The Lagrange equations give $v=cs$, and the constraint gives
$c=\ip{s}{g}/\ip{s}{{{\mathrm{B}}}s}$.  Thus the update is precisely
$\mathcal K_{\mathrm{sap}}({{\mathrm{B}}},s)g$.

Since $\norm{s}=1$ and $\ip{s}{{{\mathrm{B}}}s}\ge h_-$, the rank one operator in
\eqref{eq:sap-operator} has norm at most $h_-^{-1}$.  Moreover,
\begin{align*}
 &\norm{\mathcal K_{\mathrm{sap}}({{\mathrm{B}}},s)
       -\mathcal K_{\mathrm{sap}}({{\mathrm{B}}}',s)}_{\op}\\
 &\quad=\left|
 \frac{\ip{s}{({{\mathrm{B}}}'-{{\mathrm{B}}})s}}
      {\ip{s}{{{\mathrm{B}}}s}\ip{s}{{{\mathrm{B}}}'s}}
 \right|
 \le h_-^{-2}\norm{{{\mathrm{B}}}-{{\mathrm{B}}}'}_{\op}.
\end{align*}
For every $v\in E$,
\begin{equation*}
 \frac1{h_+}\ip{v}{{\E_\mu(s\ot s)} v}
 \le \ip{v}{\bar {{\mathrm{P}}}_{\mathrm{sap}}v}
 =\E_\mu\frac{\ip{v}{s}^2}{\ip{s}{{{\mathrm{H}}}s}}
 \le\frac1{h_-}\ip{v}{{\E_\mu(s\ot s)} v}.
\end{equation*}
Because $\norm{s}=1$, also ${\E_\mu(s\ot s)}\preceq {{\mathrm{I}}}$.  Together with
\eqref{eq:sap-spanning}, this proves
\eqref{eq:sap-bounds}--\eqref{eq:sap-mean-bounds}.  In
particular, $\bar {{\mathrm{P}}}_{\mathrm{sap}}$ is positive definite.  The product
$\bar {{\mathrm{P}}}_{\mathrm{sap}}{{\mathrm{H}}}$ is similar to the positive definite self-adjoint
operator ${{\mathrm{H}}}^{1/2}\bar {{\mathrm{P}}}_{\mathrm{sap}}{{\mathrm{H}}}^{1/2}$ and is therefore positive
stable.  Averaging a fixed number of replicas preserves the mean and all
uniform and Lipschitz bounds.

We next verify the stochastic approximation conditions without assuming the
state rate or the geometric reduction.  Let
\begin{equation*}
 {{\mathrm{\Pi}}}_k=\E_\mu\{\mathcal K_{\mathrm{sap}}({{\mathrm{B}}}_k,s)\}.
\end{equation*}
The pointwise Lipschitz bound and \eqref{eq:sap-B-rate} give
\begin{equation*}
 \norm{{{\mathrm{\Pi}}}_k-\bar {{\mathrm{P}}}_{\mathrm{sap}}}_{L^2(\op)}
 \le h_-^{-2}\norm{{{\mathrm{B}}}_k-{{\mathrm{H}}}}_{L^2(\op)}=O(k^{-\delta}).
\end{equation*}
The spectral band and the spanning condition give, for every $k$,
\[
 {{\mathrm{\Pi}}}_k\succeq\frac{\kappa}{h_+}{{\mathrm{I}}}.
\]
Taking ${{\mathrm{K}}}={{\mathrm{H}}}$ therefore yields the deterministic bound
\begin{equation*}
 ({{\mathrm{\Pi}}}_k{{\mathrm{H}}})^*{{\mathrm{H}}}+{{\mathrm{H}}}{{\mathrm{\Pi}}}_k{{\mathrm{H}}}
 =2{{\mathrm{H}}}{{\mathrm{\Pi}}}_k{{\mathrm{H}}}
 \succeq\frac{2\kappa}{h_+}{{\mathrm{H}}}^2
 \succeq\frac{2\kappa h_-^2}{h_+}{{\mathrm{I}}}.
\end{equation*}
Thus \eqref{eq:uniform-conditional-stability} holds uniformly before the
state rate is invoked.  The Lipschitz solver bound and the $L^2$ rate of ${{\mathrm{B}}}_k$ give
${{\mathrm{\Pi}}}_k{{\mathrm{H}}}\to\bar {{\mathrm{P}}}_{\mathrm{sap}}{{\mathrm{H}}}$ in $L^2$ for the subsequent
mean operator and covariance limits.

Conditional independence of solver and observation and \Cref{ass:score} give
exact centering of the untruncated solver products.  The uniform operator
bound together with \eqref{eq:sap-score-moments} gives the direction
second moment bound \eqref{eq:primitive-direction-second} and a uniform
conditional $(2+\tau)$ moment for the raw direction on the localized
neighborhood.  If $Z_{k+1}$ is the raw transported direction and
$\Delta_{k+1}$ is its discrepancy from clipping at
$r_0/\eta_{k+1}$, conditional Markov bounds give directly
\begin{equation*}
 \E(\norm{\Delta_{k+1}}\mid\F_k)\le C\eta_{k+1}^{1+\tau},
 \qquad
 \E(\norm{\Delta_{k+1}}^2\mid\F_k)\le C\eta_{k+1}^{\tau}.
\end{equation*}
Thus the discrepancy has root-$n$ negligible conditional mean, vanishing
average conditional variance, and vanishing average squared conditional
mean.  In particular, its predictable contribution satisfies
$\E\norm{\beta_{k+1}}^2\le C\eta_{k+1}$ and accumulates to $b=0$.
Consequently, \Cref{prop:state-rate} applies before
\Cref{prop:geometry-primitive}; it yields
$\E\norm{u_k}^2=O(\eta_k)$, after which the latter proposition supplies
\Cref{ass:geometry}.  The fixed anchor equivalence follows from
\Cref{lem:anchor-average-equivalence}.

The mean operator remainder follows from the displayed $L^2$ rate and
\Cref{prop:remainder-rates}(i).  The field remainder follows from
\eqref{eq:sap-field-rate} and \Cref{prop:remainder-rates}(ii).  The local
score increment moment in \eqref{eq:sap-score-moments} gives
\eqref{eq:chi-moment}.  Boundedness of the sketch operator, the
$(2+\tau)$ score moment, and \eqref{eq:sap-score-covariance} verify the
product Lindeberg condition and, by the same covariance comparison used in
the proof of \Cref{prop:fixed-sketch}, the limit
\begin{equation*}
 {{\mathrm{Q}}}_m=\E\{{{\widetilde{\mathrm{P}}}}_{\mathrm{sap}}^{(m)}
 {{\mathrm{S}}}({{\widetilde{\mathrm{P}}}}_{\mathrm{sap}}^{(m)})^*\},
 \qquad
 {{\widetilde{\mathrm{P}}}}_{\mathrm{sap}}^{(m)}
 =\frac1m\sum_{j=1}^m\mathcal K_{\mathrm{sap}}({{\mathrm{H}}},s^{(j)}).
\end{equation*}
This proves the centered conclusion of \Cref{thm:main}.

The same primitives also verify the feasible covariance conditions.  Decompose
the empirical solver mean into
\begin{equation*}
 \frac1n\sum_{k=0}^{n-1}({{\widetilde{\mathrm{P}}}}_{k+1}^{(m)}-{{\mathrm{\Pi}}}_k)
 +\frac1n\sum_{k=0}^{n-1}({{\mathrm{\Pi}}}_k-\bar {{\mathrm{P}}}_{\mathrm{sap}}).
\end{equation*}
The first term is a bounded martingale average and the second tends to zero
by \eqref{eq:sap-B-rate}; hence the one-pass mean solver estimate is
consistent.  The derivative estimate is consistent by assumption.  After
transport to the anchor, the implemented direction equals the centered
target product plus population, state dependent-score, and clipping terms.
Their Ces\`aro squared norm tends to zero by the state rate,
\eqref{eq:sap-score-moments}, the local field bound, and the clipping bounds.
Finally, the uniform $(2+\tau)$ product moment yields the outer-product
martingale weak law.  These observations verify
\Cref{ass:cov-estimation}, so \Cref{thm:cov-estimator} applies.

Independent replica centering gives
\begin{equation*}
 {{\mathrm{Q}}}_m=\bar {{\mathrm{P}}}_{\mathrm{sap}}{{\mathrm{S}}}\bar {{\mathrm{P}}}_{\mathrm{sap}}^*
 +\frac1m{{\mathrm{R}}}_{\mathrm{sap}}.
\end{equation*}
Substitution into $(\bar {{\mathrm{P}}}_{\mathrm{sap}}{{\mathrm{H}}})^{-1}{{\mathrm{Q}}}_m
{((\bar {\mathrm{P}}_{\mathrm{sap}}{\mathrm{H}})^{*-1})}$ proves \eqref{eq:sap-covariance}.  The
uniform-sphere identity $\E(s\ot s)={{\mathrm{I}}}/d$ follows from rotational invariance
and unit trace.
\end{proof}

\subsubsection{Verification for the applications}\label{app:applications}

\begin{proof}[Proof of \Cref{cor:frechet}]
For $\widehat G(x,Y)=-\Log_x(Y)$, set
\[
 \xi=-\Log_{x_\star}(Y)
\]
and let $q(x,Y)$ be the centered version of
$-\T_{x\to x_\star}\Log_x(Y)+\Log_{x_\star}(Y)$.
The first order condition gives $\E\Log_{x_\star}(Y)=0$ and hence
$\E\xi=0$ and ${{\mathrm{S}}}_F=\cov\{\Log_{x_\star}(Y)\}$.  The term $q(x,Y)$ is centered by
construction.  From \eqref{eq:frechet-l2-envelope},
\[
 \E\norm{q(x,Y)}^2
 \le \E\norm{-\T_{x\to x_\star}\Log_x(Y)+\Log_{x_\star}(Y)}^2
 \le \E L(Y)^2\norm{{R_{x_\star}^{-1}(x)}}^2.
\]
Thus the score decomposition and the local second moment bound for the
state dependent score follow.  Differentiability of $G$ identifies its
derivative with ${{\mathrm{H}}}_F$ in the chosen tangent frame.  The remaining direct
remainder, solver, geometry, averaging, and martingale conditions are those
of \Cref{thm:main}.  The outer product of the leading score is unchanged by
its sign, and its conditional product covariance is \eqref{eq:frechet-Q}.  Applying
\Cref{cor:centered} proves \eqref{eq:frechet-clt}; the two solver reductions
follow from \Cref{cor:sandwich,thm:inflation}.
\end{proof}

\begin{proof}[Proof of \Cref{cor:sphere-frechet}]
For a fixed $y\in\mathbb S^q$ outside the antipode of $x_\star$, let
$r=d(x_\star,y)$ and $e=\Log_{x_\star}(y)/r$.  The Hessian of
$\frac12d^2(x,y)$ at $x_\star$ is
\[
 r\cot(r){{\mathrm{I}}}+\{1-r\cot(r)\}e\ot e.
\]
Indeed, the radial direction $e$ has eigenvalue $1$, while every direction
orthogonal to $e$ has eigenvalue $r\cot r$.  Taking expectations gives
\eqref{eq:sphere-H}; the target score is $-re$, which gives
\eqref{eq:sphere-S}.  On $0\le r\le r_0<\pi/2$,
$r\cot r\ge r_0\cot r_0>0$, so every sample Hessian is bounded below by
$r_0\cot(r_0){{\mathrm{I}}}$ and the same is true of ${{\mathrm{H}}}_{\mathbb S}$.  The support is
away from the cut locus and all derivatives used in
\eqref{eq:frechet-l2-envelope} are uniformly bounded.  The conclusion is
therefore \Cref{cor:frechet} with the displayed ${{\mathrm{H}}}_{\mathbb S}$ and
${{\mathrm{S}}}_{\mathbb S}$.
\end{proof}

\begin{proof}[Proof of \Cref{cor:spd-frechet}]
The affine-invariant positive definite manifold has the globally defined
maps \eqref{eq:spd-exp}--\eqref{eq:spd-log}.  On a bounded geodesic ball,
the logarithm and its derivative are bounded in any fixed smooth orthonormal
frame, which verifies \eqref{eq:frechet-l2-envelope}.  The first assertion
then follows directly from \Cref{cor:frechet}.  Jacobi's formula gives
$D\log\det({{\mathrm{X}}})[{{\mathrm{U}}}]=\tr({{\mathrm{X}}}^{-1}{{\mathrm{U}}})$, so
\eqref{eq:spd-logdet-clt} follows from the delta method in
\eqref{eq:delta-method-new}.
\end{proof}

\begin{proof}[Proof of \Cref{cor:grassmann-pca}]
Use the local graph representative
\begin{equation}\label{eq:pca-graph-chart-proof}
 {{\mathrm{U}}}({{\mathrm{K}}})=({{\mathrm{U}}}_1+{{\mathrm{U}}}_2{{\mathrm{K}}})({{\mathrm{I}}}+{{\mathrm{K}}}^*{{\mathrm{K}}})^{-1/2}.
\end{equation}
At ${{\mathrm{K}}}=0$, a tangent direction is ${{\mathrm{U}}}_2{{\mathrm{K}}}$.  The Riemannian gradient of
$-\frac12\norm{{{\mathrm{U}}}^*X}^2$ is
\[
 -({{\mathrm{I}}}-{{\mathrm{U}}}{{\mathrm{U}}}^*)XX^*{{\mathrm{U}}}.
\]
Its ${{\mathrm{U}}}_2$ coordinate at ${{\mathrm{U}}}_1$ is $-z_2z_1^*$.  Replacing $XX^*$ by
${{\mathrm{M}}}$ and differentiating at ${{\mathrm{K}}}=0$ gives
$\mathcal H[{{\mathrm{K}}}]={{\mathrm{K}}}{{\mathrm{\Lambda}}}_1-{{\mathrm{\Lambda}}}_2{{\mathrm{K}}}$.  For a matrix unit ${{\mathrm{E}}}_{ij}$,
\[
 \mathcal H[{{\mathrm{E}}}_{ij}]
 =(\lambda_j-\lambda_{r+i}){{\mathrm{E}}}_{ij},
\]
so the eigengap makes $\mathcal H$ self-adjoint positive definite.

On the fixed sufficiently small graph neighborhood, the derivative of the
coherently transported sample gradient is bounded by $C\norm X^2$.
Conditional on $\F_k$, the current coordinate ${{\mathrm{K}}}_k$ is fixed and the
observation is fresh.  Hence
\[
 \E(\norm{q_{k+1}}_{\mathrm F}^2\mid\F_k)
 \le C\E\norm X^4\norm{{{\mathrm{K}}}_k}_{\mathrm F}^2,
\]
and the same moment controls the second moment of $z_2z_1^*$.
A sufficient $(2+\tau)$ score moment instead follows from
$\E\norm{X}^{4+2\tau}<\infty$.  Smoothness of
the chart gives a quadratic population remainder.  The main theorem now
gives \eqref{eq:pca-clt}.

If $X$ is Gaussian in the eigenbasis, $z_1$ and $z_2$ are independent and
\[
 \cov\{(z_2z_1^*)_{ij},(z_2z_1^*)_{i'j'}\}
 =\ind(i=i')\ind(j=j')\lambda_{r+i}\lambda_j.
\]
Applying $\mathcal H^{-1}$ proves
\eqref{eq:pca-coordinate-variance}.  Finally,
$D{{\mathrm{U}}}(0)[{{\mathrm{K}}}]={{\mathrm{U}}}_2{{\mathrm{K}}}$ in \eqref{eq:pca-graph-chart-proof}; differentiating
${{\mathrm{U}}}({{\mathrm{K}}}){{\mathrm{U}}}({{\mathrm{K}}})^*$ yields \eqref{eq:pca-projector-derivative}.
\end{proof}

\end{document}